\documentclass[accepted]{uai2025} 
                        
\usepackage[american]{babel}

\usepackage{natbib} 
\usepackage{mathtools} 
\usepackage{booktabs} 
\usepackage{tikz} 
\usepackage{amsmath}
\usepackage{amssymb}
\usepackage{bm}
\usepackage{subcaption}
\usepackage{multirow}
\usepackage{tabularx}   
\usepackage{float}

\newtheorem{problem}{Problem}
\newtheorem{definition}{Definition}
\newtheorem{lemma}{Lemma}
\newtheorem{assumption}{Assumption}
\newtheorem{theorem}{Theorem}
\newtheorem{corollary}{Corollary}
\newtheorem{proposition}{Proposition}
\newtheorem{proof}{Proof}
\newtheorem{remark}{Remark}

\title{
Error Bounds for Physics-Informed Neural Networks in Fokker-Planck PDEs
}

\author[1]{\href{mailto:<chko1829@colorado.edu>?Subject=Your UAI 2025 paper}{Chun-Wei Kong}{}}
\author[2]{Luca Laurenti}
\author[1]{Jay McMahon}
\author[1]{Morteza Lahijanian}
\affil[1]{%
    Dept. of Aerospace Eng. Sciences\\
    University of Colorado Boulder
}
\affil[2]{%
    Center for Systems and Control \\
    Delft University of Technology  
}
  
\begin{document}
\maketitle

\begin{abstract}
Stochastic differential equations are commonly used to describe the evolution of stochastic processes. The state uncertainty of such processes is best represented by the probability density function (PDF), whose evolution is governed by the Fokker-Planck partial differential equation (FP-PDE). However, it is generally infeasible to solve the FP-PDE in closed form. In this work, we show that physics-informed neural networks (PINNs) can be trained to approximate the solution PDF. Our main contribution is the analysis of PINN approximation error: we develop a theoretical framework to construct tight error bounds using PINNs. In addition, we derive a practical error bound that can be efficiently constructed with standard training methods. We discuss that this error-bound framework generalizes to approximate solutions of other linear PDEs. Empirical results on nonlinear, high-dimensional, and chaotic systems validate the correctness of our error bounds while demonstrating the scalability of PINNs and their significant computational speedup in obtaining accurate PDF solutions compared to the Monte Carlo approach.
\end{abstract}

\section{Introduction}\label{sec:intro}


\emph{Stochastic differential equations} (SDEs) are widely used to model the evolution of stochastic processes across various fields, including sciences, engineering, and finance.
In many of these applications, particularly in \emph{safety-critical} domains, a key concern is understanding how the state uncertainty in SDEs propagates over space and time.
This state uncertainty can be represented by probability density function (PDF), which is governed by the Fokker-Planck partial differential equation (FP-PDE).
However, analytical solutions for general FP-PDEs are unavailable, and numerical methods---e.g., finite elements or finite difference  \citep{spencer1993numerical,drozdov1996solution,masud2005application,pichler2013numerical,qian2019conservative,urena2020non}---are typically employed, but these methods scale poorly as the dimensionality grows beyond three \citep{tabandeh2022numerical}.
Recent advancements in deep-learning suggest physics-informed learning frameworks, called \emph{physics-informed neural networks} (PINNs), can effectively learn PDE solutions, showing notable success in handling high-dimensional systems (up to 200 dimensions) and complex geometries \citep{sirignano2018dgm,lu2021deepxde}.
Despite their effectiveness, PINNs are still subject to approximation errors, 
a crucial concern in safety-critical systems.
In this work, we tackle this challenge by developing a novel framework to approximate FP-PDE solutions using PINNs and rigorously bounding the approximation error.

Recent works on using PINNs to approximate solutions to PDEs typically analyze approximation errors in terms of \emph{total} error, 
capturing cumulative approximation error across space and time
\citep{de2022generic,de2022error,mishra2023estimates,de2024error}.
While useful in some applications, this approach is less informative for SDEs and their PDF propagation. 
Moreover, total error bounds are often overly loose, sometimes exceeding the actual
errors by several orders of magnitude. Crucially, these bounds do not provide insight into the worst-case approximation error at specific time instances or within particular subsets of space, which
is essential in many stochastic systems. 
For example, in autonomous driving scenarios involving pedestrian crossings, accurate prediction and bounding the probability of collision requires precise reasoning over specific time instances and spatial regions. Loose over-approximations can lead to undesirable behaviors, such as sudden braking.

In this work, we show how PINNs can be used to approximate solutions to FP-PDE (i.e., PDF of an SDE's state) and, more importantly, introduce a framework for tightly bounding the worst-case approximation error 
over the subset of interest in state space as a function of time.
Our key insight is that the approximation error is related to the residual of the FP-PDE
and is governed by another PDE. 
Hence, a second PINN can be used to learn the error, with its own error also following a PDE. This results in a recursive formulation of error functions, each of which can be approximated using a PINN.  
We establish sufficient training conditions under which this series converges with a finite number of terms. Specifically, we prove that two PINNs are enough to obtain arbitrarily tight error bounds. Additionally, we derive a more practical bound requiring only one error PINN at the cost of losing arbitrary tightness, and provide a method to verify its sufficient condition. 
Furthermore, we propose a training scheme with regularization and discuss extensions to other linear PDEs. 
Finally, we illustrate and validate these error bounds through experiments on several SDEs, supporting our theoretical claims.

In short, the main contribution is five-fold:
\begin{itemize}
    \item a method for approximating the PDF of processes modeled by SDEs using PINNs,
    \item a novel approach to tightly bound the approximation error over time and space through a recursive series of error functions learned by PINNs,
    \item a proof that this recursive process converges with only two PINNs needed for arbitrarily tight bounds,
    \item the derivation of a more practical error bound requiring just one PINN, along with a method to verify its sufficiency, and
    \item validation of the proposed error bounds through experiments on several SDEs.
\end{itemize}

\paragraph{Related Work}
Research on using PINNs to approximate PDE solutions often focuses on total error, which represents the cumulative error across all time and space.
For instance, \citet{mishra2023estimates} derive an abstract total error bound. Nevertheless, their numerical experiments reveal that this total error bound is loose, exceeding the actual errors by nearly three orders of magnitude. 
A similar approach is extended to Navier-Stokes equations \citep{de2024error}, with comparable results.
\citet{de2022error} consider FP-PDEs derived from linear SDEs only. They propose an abstract approach to bound the total error, but no numerical experiments are presented.
\citet{de2022generic} also derive total error bounds for PINNs (and operators) assuming  a priori error estimate.
In contrast, our work emphasizes bounding the worst-case error at any time of interest for general SDEs, which is particularly valuable in practical applications of stochastic systems (e.g., systems subject to chance constraints \citep{oguri2021robust,paiola2024evaluation}).


To demonstrate the approximation capabilities of neural networks, error analysis is a key.
For example,
\citet{hornik1991approximation} proves that a standard multi-layer feed-forward neural network can approximate a target function arbitrarily well.
\citet{yarotsky2017error} considers the worst-case error and shows that deep ReLU neural networks are able to approximate universal functions in the Sobolev space.
Recently, deep operator nets (DeepONet) have been suggested to learn PDE operators, with
\citet{lanthaler2022error} proving that for every $\epsilon>0$, there exists DeepONets such that the total error is smaller than $\epsilon$.
While these studies show the capabilities of neural networks, they do not address the critical question: what are the quantified errors for a given neural network approximation? This is the central issue tackled by our work.

Error estimates have also been investigated when neural networks are trained as surrogate models for given target functions.
For instance, \citet{barron1994approximation} derives the error between the learned network and target function in terms of training configurations. To learn a latent function with quantified error, Gaussian process regression \citep{archambeau2007gaussian} is often employed, where observations of an underlying process are required to learn the mean and covariance.
Recently, \citet{yang2022guaranteed} estimate the worst-case error given target functions and neural network properties. 
Nevertheless, a fundamental difference between our work and these studies is that 
we do not assume knowledge of the true solutions (latent functions) or rely on data from the underlying processes.

Solving PDEs is an active research area with various established approaches. For the FP-PDE equation, numerical methods, such as the finite elements, finite differences, or Galerkin projection methods, have been employed \citep{spencer1993numerical,drozdov1996solution,masud2005application,chakravorty2006homotopic,pichler2013numerical,qian2019conservative,urena2020non}.
For PDF propagation, instead of solving the FP-PDE, some approaches perform a time-discretization of the SDE and use  
Gaussian mixture models \citep{terejanu2008uncertainty}. 
Recent works \citep{khoo2019solving, song2025finite, lin2024deep} employ numerical methods for approximating transition probability between two regions, which is also governed by the FP-PDE.
While these studies show accurate approximations from posterior evaluation, 
they can be computationally demanding and often lack rigorous error quantification and bounding.

\section{Problem Formulation} \label{sec:problem}
The aim of this work is state uncertainty propagation with quantified error bounds for continuous time
and space stochastic processes using deep neural networks. We specifically focus on 
(possibly nonlinear) Stochastic Differential Equations (SDEs) described by
\begin{equation}\label{eq:sde_general}
    d\bm{x}(t) = f(\bm{x}(t),t) dt + g(\bm{x}(t),t) d\bm{w}(t),
\end{equation}
where 
$t \in T \subseteq \mathbb{R}_{\geq 0}$ is time,
$\bm{x}(t) \in X \subseteq \mathbb{R}^n$ is the system state at time $t$, and $\bm{w}(t) \in \mathbb{R}^m$ is a standard Brownian motion. 
For $\Omega= X\times T$, function $f: \Omega \rightarrow \mathbb{R}^n$ represents the deterministic evolution of the system, and function $g: \Omega \rightarrow \mathbb{R}^{n \times m}$ is a term that defines the coupling of the noise.
We assume that $f(x,t)$ and $g(x,t)$ satisfy the usual regularity conditions (e.g., see \citet[Ch.~7.1]{evans2022partial}), and
denote the $i$-th dimension of $f$ and $(j,k)$-th element of $g$ by $f_i$ and $g_{jk}$, respectively.
The initial state $\bm{x}(0)$ is a random variable distributed according to a given probability density function (PDF) $p_0:X\rightarrow \mathbb{R}_{\geq 0}$, i.e., $\bm{x}(0) \sim p_0$. We assume that $p_0$ is bounded and smooth.

The solution to the SDE in Eq.~\eqref{eq:sde_general} is a stochastic process $\bm{x}$ 
with a corresponding PDF $p:\Omega \rightarrow \mathbb{R}_{\geq 0}$  
over space and time, i.e., $\bm{x}(t) \sim p(\cdot, t)$ \citep{oksendal2003stochastic}.
PDF $p$
is governed by the Fokker-Planck partial differential equation (FP-PDE):
\begin{multline}\label{eq:fp_pde}
    \frac{\partial p(x,t)}{\partial t} + \sum_{i=1}^n \frac{\partial}{\partial x_i} [f_i p(x,t)] - \\
    \frac{1}{2} \!\! \sum_{i=1,j=1}^n \!\! \frac{\partial ^2}{\partial x_i \partial x_j} \left[ \sum_{k=1}^m g_{ik}g_{jk}p(x,t) \right] = 0,
\end{multline}
and must satisfy the initial condition
\begin{equation}
    p(x,0) = p_0(x) \qquad \forall x \in X.
    \label{eq:fp_pde_init_cond}
\end{equation}
To simplify notation, we denote by $\mathcal{D}[\cdot]$ the differential operator associated with the FP-PDE, i.e.,
$$\mathcal{D}[\cdot]:= \frac{\partial}{\partial t}[\cdot] + \sum_{i=1}^n \frac{\partial}{\partial x_i} [f_i \cdot] - \frac{1}{2} \! \sum_{i,j=1}^n \frac{\partial ^2}{\partial x_i \partial x_j} \left[ \sum_{k=1}^m g_{ik}g_{jk} \cdot \right].$$ Then, Eqs.~\eqref{eq:fp_pde} and \eqref{eq:fp_pde_init_cond} can be rewritten in a compact form  as
\begin{equation}
    \mathcal{D}[p(x,t)] = 0 \quad \text{ subject to } \quad p(x,0) = p_0(x).
    \label{eq:fp_pde_compact}
\end{equation}
Note that, since $f$ and $g$ are assumed to be regular, the PDE in Eq.~\eqref{eq:fp_pde_compact} is well-posed, i.e., there exists a smooth and unique solution.

Obtaining solution $p(x,t)$ to Eq.~\eqref{eq:fp_pde_compact} in closed-form is generally not possible, and even numerical approaches are limited to simple SDEs \citep{spencer1993numerical,drozdov1996solution,masud2005application,chakravorty2006homotopic,pichler2013numerical,qian2019conservative,urena2020non}.
In this work, we focus on using PINNs to approximate $p$, and crucially, we aim to formally bound the resulting approximation error.
\begin{problem}\label{prob:1}
    Given FP-PDE in Eq.~\eqref{eq:fp_pde_compact}, a bounded subset $X' \subset X$, and a bounded time interval $T' \subset T$,
    train a neural network $\hat{p}(x,t)$ that approximates the solution $p(x,t)$, and for every $t \in T'$ construct an error bound $B:T' \rightarrow \mathbb{R}_{> 0}$ such that 
    \begin{equation}
        \sup_{x\in X'} |p(x,t)-\hat{p}(x,t)| \leq B(t).
    \end{equation}
\end{problem}


Note that no data is assumed on $p$.
Instead, our approach leverages the governing 
Eq.~\eqref{eq:fp_pde_compact} for both training $\hat{p}$ and quantifying its error. 
Specifically, we first 
show that existing PINN training methods for PDE solutions can be adapted to approximate $p$ effectively if the training loss is sufficiently small. 
Then, 
we show that the resulting approximation error can be written as a series of approximate error functions, each of which satisfying a PDE similar to Eq.~\eqref{eq:fp_pde_compact}. This implies that each error function itself can be approximated using a PINN. Then, we derive conditions, under which only a finite number of such PINNs is needed to obtain a guaranteed error bound $B(t)$.

\begin{remark}
    While we focus on $\hat{p}$ being a neural network, our method of deriving error bound $B(t)$ is not limited to neural networks and generalizes to any smooth function $\hat{p}$ that approximates the true solution $p$.
\end{remark}
\section{PDF Approximation and Error Characterization via PINNs
}\label{sec:learn}

Here, we first describe a method for training a neural network to approximate PDF $p(x,t)$ for Problem~\ref{prob:1}, and then derive a recursive error-learning approach to estimate the approximation error.

\paragraph{Learning PDF $p$}\label{sec:learn_pdf}
We approximate $p(x,t)$ by learning a neural network $\hat{p}(x,t;\theta)$, where $\theta$ represents the parameters. 
During training, 
spatio-temporal data points $\{(x_j,0)_j \}_{j=1}^{N_0}, \{(x_j,t_j)_j \}_{j=1}^{N_r} \subset \Omega$,
for some $N_0,N_r \in \mathbb{N}$, are sampled, and the loss function is derived from the governing physics in Eq.~\eqref{eq:fp_pde_compact} as 
\begin{align}\label{eq:pinn_loss}
    \mathcal{L} = w_{0}\mathcal{L}_{0} + w_{r}\mathcal{L}_{r},
\end{align}
where  $w_0,w_r \in \mathbb{R}^+$ are the weights, and
\begin{subequations}\label{eq:pinn_sub_loss}
    \begin{align} 
        \mathcal{L}_{0} &= \frac{1}{N_0}\sum_{j=1}^{N_0}\big( p_0(x_j) - \hat{p}(x_j,0;\theta) \big)^2, \\
        \mathcal{L}_{r} &= \frac{1}{N_r}\sum_{j=1}^{N_r}\big(\mathcal{D}[\hat{p}(x_j,t_j;\theta)]\big)^2.
    \end{align}
\end{subequations}
The loss function in Eq.~\eqref{eq:pinn_loss} quantifies the deviation of the true and approximate solutions in terms of the initial condition ($\mathcal{L}_0$) and the infinitesimal variation over space and time ($\mathcal{L}_r$).
The parameters of $\hat{p}(x,t;\theta)$ are learned by minimizing the loss function, i.e., $\theta^{*} = \arg\min \mathcal{L}.$ 

We note that overfitting can arise if the training samples in Eqs.~\eqref{eq:pinn_sub_loss} do not sufficiently cover the domain $\Omega$. 
To address this, we rely on common sampling practices in PINNs (e.g., uniformly sampled \citep{sirignano2018dgm}) along with an adaptive sampling method in \citet{lu2021deepxde} to improve sampling efficiency (see Appendix~\ref{appendix:train_and_results} for details).

\begin{assumption}\label{assumption:phat}
    $\hat{p}:\Omega \rightarrow \mathbb{R}$ is assumed to be
    smooth. 
\end{assumption}
Assumption~\ref{assumption:phat} is present because $\hat{p}$ is trained by the physics-informed loss in Eq.~\eqref{eq:pinn_loss}, in which the second term $\mathcal{L}_r$ requires the computation of the first and second derivatives with respect to time and space, respectively.
To satisfy Assumption~\ref{assumption:phat}, smooth activation functions (e.g., $\mathrm{Tanh}$ and $\mathrm{Softplus}$) can be used in the architecture of $\hat{p}(x,t;\theta)$.
While \(\hat{p}\) here is real-valued, one can further ensure \(\hat{p}\ge 0\) using non-negative activation functions (e.g., exponent or squared) for the last layer.

    
The weights \(w_0\) and \(w_r\) in Eq.~\ref{eq:pinn_loss} balance the initial-condition and PDE-residual terms. Although tuning these weights can affect the speed of convergence in practice, the training convergence does not rely on finding optimal weights \citep{shin2020convergence,mishra2023estimates}.


We emphasize that this method of training requires only the initial PDF $p_0(x)$ and differential operator $\mathcal{D}$, allowing loss evaluation on unlimited space-time samples. This key distinction sets PINNs apart from data-driven learning, which relies on (limited) data of (unavailable) true solution $p$.

\paragraph{Recursive Learning of Approximation Error}\label{sec:learn_issue}

Given trained $\hat{p}$, we show that its approximation error can be characterized as a series of approximate solutions to PDEs. Specifically, we define the error as
\begin{align}
    \label{eq:err_def}
    e_1(x,t) := p(x,t) - \hat{p}(x,t).
\end{align}
Note that FP-PDE operator $\mathcal{D}$ is a linear operator; hence, by applying it to $e_1$, we obtain:
\begin{equation*}
    \mathcal{D}[e_1] = \mathcal{D}[p-\hat{p}] = \mathcal{D}[p] - \mathcal{D}[\hat{p}].
    \label{eq:apply_fppde_on_error}
\end{equation*}
As $\mathcal{D}[{p}]=0$, we can see that the error is essentially related to the residue of $\mathcal{D}[\hat{p}]$.  Then, 
we can define the governing PDE of $e_1(x,t)$ as
\begin{equation}
    \mathcal{D}[e_1] + \mathcal{D}[\hat{p}] = 0
    \; \text{ s.t. } \;
    e_1(x,0) = p_0(x) - \hat{p}(x,0).
    \label{eq:error_1_e1_func_pde}
\end{equation}
Hence, using a similar approach as in Eqs.~\eqref{eq:pinn_loss} and ~\eqref{eq:pinn_sub_loss}, a PINN can be trained to approximate $e_1(x,t)$ using its governing physics in Eq.~\eqref{eq:error_1_e1_func_pde}.
Based on this, we can define the $i$-th error and its associated approximation in a recursive manner.
\begin{definition}[$i$-th error and approximation]\label{def:1}
    Let $e_0 := p$ and $\hat{e}_0 := \hat{p}$. For $i \geq 1$, we define the $i$-th error $$e_i(x,t):=e_{i-1}(x,t)-\hat{e}_{i-1}(x,t),$$ where each $\hat{e}_i$ is a smooth and bounded function constructed by PINN to approximate $e_i$. Each $e_i$ is the solution to the recursive PDE
    \begin{multline*}
        \mathcal{D}[e_i(x,t)] +  \sum_{j=1}^{i} \mathcal{D}[\hat{e}_{j-1}(x,t)] = 0 \quad {s.t.} \\
        e_i(x,0) = e_{i-1}(x,0) - \hat{e}_{i-1}(x,0).
    \end{multline*}
\end{definition}
By the construction in Definition~\ref{def:1}, the approximation error $e_1(x,t)$, for every choice of $n \geq 0$, is given by
\begin{align}
    e_1(x,t) &= p(x,t)-\hat{p}(x,t) \nonumber \\
    &= \sum_{i=1}^{n} \hat{e}_i(x,t) + e_{n+1}(x,t).
    \label{eq:error_series}
\end{align}
Although this recursive procedure estimates the unknown approximation error $e_1$, it does not directly provide a worst-case error bound. This is because, regardless of how many $\hat{e}_i$, $i=1,2,\dots,n$, are constructed, an unquantified error term $e_{n+1}$ always remains. To address this, we present our error bound theory in the next section.

\section{Arbitrary Tight Error Bound}\label{sec:theory}


Here, we derive upper bounds for the approximation error $e_1$, specifically, for the right-hand side of Eq.~\eqref{eq:error_series}. 
We show that, by training just two PINNs under certain sufficient conditions, the series can be bounded with arbitrary precision. 
All proofs for the lemmas and theorems are provided in Appendices~\ref{proof:lemma1}--\ref{proof:arbitrary_tight}.

First, we express how well $\hat{e}_i$ approximates the $i$-th error $e_i$ by defining the \textit{relative approximation} factor $\alpha_i(t)$ as 
\begin{equation}
    \alpha_i(t)  := \frac{\max_{x \in X'} | e_i(x,t) - \hat{e}_i(x,t)|} {\max_{x \in X'} |\hat{e}_i(x,t)|}.
    \label{eq:alpha_def}
\end{equation}
Recall from Definition~\ref{def:1} that $e_i-\hat{e}_i = e_{i+1}$.  Hence, Eq.~\eqref{eq:alpha_def} can be written in a recursive form as
\begin{equation}
    \max_{x \in X'}|e_{i+1}(x,t)| = \alpha_i(t) \max_{x \in X'} |\hat{e}_i(x,t)|,
    \label{eq:assume_alpha_relation}
\end{equation}
which relates the unknown $(i+1)$-th error to the $i$-th error approximation. 
Now, let $e_i^*(t)$ and $\hat{e}_i^*(t)$
denote the maximum of $e_i(x,t),\hat{e}_i(x,t)$ over $X'$, respectively, i.e., 
\begin{subequations}
    \label{eq:max_error}
    \begin{align}
        e_i^*(t) &:= \max_{x \in X'} |e_i(x,t)|,\\ 
        \hat{e}_i^*(t) &:= \max_{x \in X'} |\hat{e}_i(x,t)|.
    \end{align}
\end{subequations}
Recall that each $\hat{e}_i(x,t)$ can be represented using a PINN. 
Hence, it is safe to assume that the  absolute value of its upper-bound is strictly greater than zero.
\begin{assumption}\label{assumption:ehat_bigger_than_zero}
    Assume that, for all 
    $1 \leq i < n$,
    $\hat{e}^*_i(t) > 0$.
\end{assumption}
Then, the following lemma upper-bounds the approximation error $e_1(x,t)$ using $\hat{e}_i^*(t)$.
\begin{lemma}\label{lemma:bouding series of ratios}
    Consider the approximation error $e_1(x,t) = p(x,t)-\hat{p}(x,t)$ in Eq.~\eqref{eq:error_series} with $n \geq 2$, and the upper-bounds  $\hat{e}^*_i(t)$ for $1 \leq i < n$ in Eq.~\eqref{eq:max_error}.  Define ratio
    \begin{equation}
        \label{eq:approx ratio}
        \gamma_{\frac{i+1}{i}}(t) := \frac{\hat{e}_{i+1}^*(t)}{\hat{e}_{i}^*(t)}.
    \end{equation}
    Then, 
    under Assumption~\ref{assumption:ehat_bigger_than_zero}, 
    it holds that, $\forall x\in X'$,
    \begin{multline}\label{eq:error_bound_by_max_error_series_ratios}
    |e_1(x,t)| \le \hat{e}_1^*(t)\Bigl(1 + \sum_{m=2}^n \prod_{i=1}^{m-1}\gamma_{\tfrac{i+1}{i}}(t) \\ 
    + \frac{e_{n+1}^*}{\hat{e}_{n-1}^*}\prod_{i=1}^{n-2}\gamma_{\tfrac{i+1}{i}}(t)\Bigr)
    \end{multline}
\end{lemma}

Next, we derive an upper- and lower-bound for the ratio $\gamma_{\frac{i+1}{i}}(t)$ in Eq.~\eqref{eq:error_bound_by_max_error_series_ratios} using $\alpha_i(t)$.
\begin{lemma}\label{lemma:bounds_on_ratios}
    If the relative approximation factors $\alpha_i(t)<1$ for all $2 \leq i < n $, then 
    \begin{equation}
        \frac{\alpha_{i-1}(t)}{1+\alpha_{i}(t)} \leq \gamma_{\frac{i}{i-1}(t)} \leq \frac{\alpha_{i-1}(t)}{1-\alpha_{i}(t)}.
    \label{eq:bounds_on_ratios}
    \end{equation}
\end{lemma}

Lemma~\ref{lemma:bounds_on_ratios} establishes the relationship between ratio $\gamma_{\frac{i}{i-1}}$ and relative approximation factors $\alpha_i$ under condition $\alpha_i < 1$. 
Intuitively, this condition holds when $\hat{e}_i$ approximates $e_i$ reasonably well (see Eq.~\eqref{eq:alpha_def}). 
Lastly, we show that under certain conditions on $\alpha_1$ and $\alpha_2$, an ordering over $\gamma_{\frac{2}{1}},\gamma_{\frac{3}{2}},\dots,\gamma_{\frac{i}{i-1}}$ can be achieved.
\begin{lemma}\label{lemma:exist_decrease_ratio}
    If, for all $t \in T'$,
    \begin{subequations}
        \begin{align}
            \label{eq:alpha1_condition}
            & 0 < \alpha_1(t) < 1, \\
            \label{eq:alpha2_condition1}
            & 0 < \alpha_2(t) < 1-\alpha_1(t), \\
            \label{eq:alpha2_condition2}
            & \alpha_2(t)(1+\alpha_2(t)) < \alpha_1(t)^2,
        \end{align}
    \label{eq:alpha_conditions}
    \end{subequations}
    then there exist feasible $0 \leq \alpha_{i}(t) < 1$ for $2 < i < n$ such that
    \begin{equation}
        \gamma_{\frac{i}{i-1}}(t) < \gamma_{\frac{2}{1}}(t) < 1.
        \label{eq:decrease_ratio}
    \end{equation}
\end{lemma}

The intuition behind Lemma \ref{lemma:exist_decrease_ratio} is that if $\hat{e}_1$ and $\hat{e}_2$ are trained to certain accuracy (satisfying Conditions~\ref{eq:alpha_conditions}), 
then there exist feasible $\hat{e}_3,\hat{e}_4,\dots,\hat{e}_{n-1}$ such that the ratios $\gamma_{\frac{3}{2}},\gamma_{\frac{4}{3}},\dots,\gamma_{\frac{n-1}{n-2}}$ are upper bounded by $\gamma_{\frac{2}{1}} < 1$. 
Equipped with Lemmas~\ref{lemma:bouding series of ratios}-\ref{lemma:exist_decrease_ratio}, we can state our main result, which is an upper-bound on the approximation error of $\hat{p}$.
Specifically, the following theorem shows that the approximation error bound in Lemma~\ref{lemma:bouding series of ratios} becomes a geometric series as $n\to \infty$ under Conditions~\ref{eq:alpha_conditions}; hence, solving Problem~\ref{prob:1}.

\begin{theorem}[Second-order error bound]\label{theorem:temporal_error_bound}
    Consider Problem~\ref{prob:1} and two approximate error functions $\hat{e}_1(x,t), \hat{e}_2(x,t)$ constructed by Definition~\ref{def:1} 
    that satisfy Conditions~\ref{eq:alpha_conditions}. Then, 
    \begin{equation}
        |p(x,t)-\hat{p}(x,t)| \leq B_2(t) = \hat{e}_1^*(t)\Big( \frac{1}{1-\gamma_{\frac{2}{1}}(t)} \Big),
        \label{eq: temporal error bound}
    \end{equation}
    where $\hat{e}_1^*(t)$ is defined in Eq.~\eqref{eq:max_error}, and $\gamma_{\frac{2}{1}}(t) = \hat{e}_2^*(t)/ \hat{e}_1^*(t)$.
\end{theorem}

The above theorem shows that the second-order error bound $B_2(t)$ can be obtained by training only two PINNs that approximate the first two errors $e_1, e_2$ according to Definition~\ref{def:1} and that satisfy Conditions~\ref{eq:alpha_conditions}.
In fact, using these two PINNs, it is possible to construct an arbitrary tight $B_2$ as stated below.

\begin{theorem}[Arbitrary tightness]\label{theorem:error_bound_tightness}
    Given Problem~\ref{prob:1} and tolerance $\epsilon \in (0,\infty)$ on the error bound, an error bound $B_2(t)$ in Theorem~\ref{theorem:temporal_error_bound} can be obtained by training two approximate error functions $\hat{e}_1(x,t)$ and $\hat{e}_2(x,t)$ through physics-informed learning such that 
    \begin{equation}
        B_2(t) - \max_{x \in X'}|e_1(x,t)| < \epsilon.
        \label{eq:arbitrary_tight_def}
    \end{equation}
\end{theorem}

The proof of Theorem~\ref{theorem:error_bound_tightness} is based on the observation that $\gamma_{\frac{2}{1}} \to 0$ when (i) $\hat{e}_1(x,t) \to e_1(x,t)$ and (ii) $\hat{e}_2(x,t) \to e_2(x,t)$.  Then, according to Eq.~\eqref{eq: temporal error bound}, $B_2(t) \to \hat{e}^*_1(t)$, which itself $\hat{e}^*_1(t) \to e^*_1(t)$ under (i). 
By the theoretical convergence of PINNs \citep{shin2020convergence,mishra2023estimates}, $\hat{e}_1$ and $\hat{e}_2$ can be made arbitrary well; thus $B_2$ can be arbitrary tight. This result is important because it shows that arbitrary tightness can be achieved without the need for training infinite number of PINNs, i.e., $\hat{e}_i$, $i=1,2, \ldots$

\begin{remark}
The construction of $B_2(t)$ in Theorem~\ref{theorem:temporal_error_bound} only requires the values of $\hat{e}^*_1(t)$ and $\gamma_{\frac{2}{1}}(t)$ which are obtained from the known functions $\hat{e}_1(x,t),\hat{e}_2(x,t)$.
Checking for $\alpha_1$ and $\alpha_2$ conditions can be performed \textit{a posterior}.  
\end{remark}

\paragraph{$n$-th Order Space-time Error Bound ($n > 2$)}

Here, we derive a generalized error bound using approximation error PINNs $\hat{e}_i$, where $i = 1,\ldots, n$ for $n > 2$.
Note that an alternative way to express the error bound in Theorem~\ref{theorem:temporal_error_bound} is as an interval $e_1(x,t) \in \big[-B_2(t), B_2(t) \big],$ 
which is uniform over $x$ for any $t\in T$. 
Below, we show that, for $n > 2$, an error bound that depends on both space and time can be constructed.
\begin{corollary}[Space-time Error Bound]
    Consider PINNs $\hat{e}_i(x,t)$, $i = 1,\ldots, n$, for some $n > 2$ trained per Def.\ref{def:1}
    such that $\alpha_{n-1}$ and $\alpha_{n}$ satisfy Conditions~\ref{eq:alpha_conditions}, and define the $n$-th order temporal error bound to be
    $$B_n(t) = \hat{e}_{n-1}^*(t)( \frac{1}{1-\gamma_{\frac{n}{n-1}}(t)}),$$ where $\hat{e}_{n-1}^*(t)$ is defined in~\eqref{eq:max_error}, and $\gamma_{\frac{n}{n-1}}(t) = \hat{e}_{n}^*(t)/ \hat{e}_{n-1}^*(t)$.
    Then, 
    \begin{align}
        e_1(x,t) \in \Big[ \sum_{i=1}^{n-2} \hat{e}_i(x,t) - B_n(t), \; \sum_{i=1}^{n-2} \hat{e}_i(x,t) + B_n(t) \Big].
    \end{align}
    \label{corollary:spacetime error bound}
\end{corollary}
This corollary shows that, even though the $2$-nd order error approximation is sufficient to obtain a time-varying bound (Theorem~\ref{theorem:temporal_error_bound}), higher order approximations lead to more information, i.e., space in addition to time, on the error bound.

\paragraph{Feasibility Analysis} 
Now, we analyze the feasibility for $\hat{e}_1$ and $\hat{e}_2$ satisfying Conditions~\ref{eq:alpha_conditions} in Theorem~\ref{theorem:temporal_error_bound}.
Specifically, Condition~\ref{eq:alpha1_condition} on $\alpha_1$ indicates that $\hat{e}_1$ must be learned well enough so that the magnitude of its maximum approximation error is less than its own maximum magnitude (see Eq.~\eqref{eq:alpha_def}).  
By fixing $\alpha_1$, Conditions~\ref{eq:alpha2_condition1}-\ref{eq:alpha2_condition2} on $\alpha_2$ require $\hat{e}_2$ to approximate $e_2$ more accurately than the approximation of $e_1$ by $\hat{e}_1$.  
These conditions are feasible in principle by the same convergence argument above. However, there are some practical challenge as discussed below.

\paragraph{Practical Challenge}
The challenge to construct $B_2(t)$ stems from the condition that $\hat{e}_2$ needs to approximate $e_2$ far more accurately than $\hat{e}_1$ approximates $e_1$, making training a PINN for $\hat{e}_2$ extremely difficult. Additionally, since the explicit values of $\alpha_1$ and $\alpha_2$ are unknown, there is no clear criterion for determining when to stop training $\hat{e}_1$ and $\hat{e}_2$. 
To address this, we provide a method for verifying the condition on $\alpha_1$ and derive a bound that depends only on this condition below.

\section{First-order error bound}\label{sec:tight_error_bound}


Here, we introduce a first-order error bound using a single PINN, overcoming the challenge of obtaining $B_2$. We also provide an implicit formula to determine training termination. Additionally, we discuss the feasibility of this approach, outline our training scheme, and explore relevant extensions. Complete proofs can be found in Appendices~\ref{proof:coro1} and~\ref{proof:prop1}.

By considering Eq.~\eqref{eq:error_series} with $n=2$ and using Lemmas~\ref{lemma:bouding series of ratios} and~\ref{lemma:bounds_on_ratios}, we can derive the first-order error bound, as stated in the following corollary.
\begin{corollary}[First-order error bound]\label{corollary:special_error_bound}
    Consider Problem~\ref{prob:1}, and let $\hat{e}_1$ be trained such that $\alpha_1(t)<1$ for all $t \in T'$. Then
    \begin{equation}
        |p(x,t)-\hat{p}(x,t)| < B_1(t) = 2\hat{e}_1^*(t).
        \label{eq:tight_error_bound}
    \end{equation}
\end{corollary}
Note that the first-order error bound $B_1(t)$ can be at most twice as large as the arbitrary tight second-order bound $B_2(t)$ in Theorem~\ref{theorem:temporal_error_bound}, but it offers significant practical advantages. 
Firstly, the second-order bound $B_2(t)$ requires training a second PINN \(\hat{e}_2\) after training \(\hat{e}_1\). However, achieving the required accuracy for \(\hat{e}_2\) in practice is quite challenging. In contrast, the first-order bound in Corollary~\ref{corollary:special_error_bound} relies solely on the approximation provided by a single PINN, \(\hat{e}_1\). 
Secondly, the condition $\alpha_1(t) < 1$ can be verified during $\hat{e}_1$ training using properties of the FP-PDE, as detailed below.

\paragraph{Checking $\alpha_1 < 1$ Condition}
From the definition of $\alpha_1(t)$ in Eq.~\eqref{eq:alpha_def}, it suffices to bound the unknown term $|e_1(x,t)-\hat{e}_1(x,t)|$ for all $(x,t) \in \Omega$ to check for $\alpha_1(t)<1$.
We do this by using three constants: the first two constants are related to PDE stability and quadrature rules \citep{mishra2023estimates}, and the third constant comes from Sobolev embedding theorem 
\citep[Theorem 12.71]{hunter2001applied}\citep{mizuguchi2017estimation}.

The first constant $C_{pde}$ is related to the \emph{stability} of the first error PDE, which is defined as
\begin{equation*}
    \|e_1 - \hat{e}_1\|_Z \leq C_{pde} \| (\mathcal{D}[e_1] + \mathcal{D}[\hat{p}]) - (\mathcal{D}[\hat{e}_1]+\mathcal{D}[\hat{p}]) \|_Y,
\end{equation*}
where 
$Z=W^{k,q}$ norm , $Y = L^s$ norm, $1\leq s,q < \infty,$ and $k\geq 0$.
Note that since $e_1,\hat{e}_1$ and $(\mathcal{D}[e_1] + \mathcal{D}[\hat{p}]) - (\mathcal{D}[\hat{e}_1]+\mathcal{D}[\hat{p}]) = 0-(\mathcal{D}[\hat{e}_1]+\mathcal{D}[\hat{p}])$
are bounded
,
such constant $C_{pde}$ exists.

The second constant $C_{quad} > 0$ is related to the deviation between integral and its approximation with finite samples. Let $\mathcal{I} = \int_{\Omega} \Big(\mathcal{D}[\hat{e}_1(x,t)] + \mathcal{D}[\hat{p}(x,t)]\Big) dxdt$ be the integral of interest, and $\bar{\mathcal{I}} = \sum_{j=1}^N w_j \Big(\mathcal{D}[\hat{e}_1(x_j,t_j)] +  \mathcal{D}[\hat{p}(x_j,t_j)] \Big)$ be its associated approximation, where $\{(x_j,t_j)_j\}_{j=1}^N \in \Omega$ is a set of $N$ quadrature points, and $w_j \in \mathbb{R}_{>0}$ are weights according to the quadrature rules. Then $C_{quad}$ is defined such that, for some $\beta > 0$,
$
    \left|\mathcal{I} - \bar{\mathcal{I}} \right| \leq C_{quad}N^{-\beta}.
$

The third constant $C_{embed}$ from Sobolev embedding theorem is defined as
\begin{equation*}
    \|e_1(x,t)-\hat{e}_1(x,t)\|_{\infty} \leq C_{embed} \|e_1(x,t)-\hat{e}_1(x,t)\|_{W^{1,q}}.
\end{equation*}
Constant $C_{embed}$ exists because $e_1(x,t)$ and $\hat{e}_1(x,t)$ are bounded (per Definition~\ref{def:1}), and the first derivatives of $e_1(x,t)$ and $\hat{e}_1(x,t)$ are also bounded over the considered domain of Problem~\ref{prob:1}.
With these constants, we propose an implicit checking formula for $\alpha_1(t)<1$.
\begin{proposition}[Checking $\alpha_1(t)<1$]    \label{prop:checking_alpha1}
    Let $\{(x_j,t_j)_j\}_{j=1}^N \in \Omega$ be $N$ space-time samples based on quadrature rules, $\hat{e}_1(x,t)$ be the first error approximation, and $\mathcal{L}^{(1)}$ be the physics-informed loss of $\hat{e}_1(x,t)$ evaluated on the set $\{(x_j,t_j)_j\}_{j=1}^N$. Then for some $q\geq 2$ and $\beta > 0$, $\alpha_1(t) < 1$ for all $t \in T'$ if
    \begin{align}
        C_{embed}C_{pde}\Big( \mathcal{L}^{(1)} + C_{quad}^{\frac{1}{q}}N^{\frac{-\beta}{q}} \Big) < \min_t \hat{e}_1^*(t).
        \label{eq:checking_alpha1}
    \end{align}
\end{proposition}
By Proposition~\ref{prop:checking_alpha1}, it is clear that as the training loss decreases ($\mathcal{L}^{(1)} \rightarrow 0$) with sufficiently large number of samples ($N \rightarrow \infty$), the left-hand side of Eq.~\eqref{eq:checking_alpha1} approaches zero. Hence, condition $\alpha_1 < 1$ can be satisfied
as validated in our numerical evaluations.

Note that, for the constants in Proposition~\ref{prop:checking_alpha1}, it is sufficient to have upper bounds.
Specifically, \citet{mishra2023estimates} show a method of over-estimating $C_{pde}$. 
Constant $C_{embed}$ depends on the domain geometry \citep{mizuguchi2017estimation}.  
Also note that a sufficiently large $N$ can ensure $C_{quad}N^{-\beta} \ll 1$.

\begin{figure*}[t]
    \centering
    \begin{minipage}{0.59\textwidth}
        \centering
        \begin{subfigure}{0.40\textwidth}
            \centering
            \includegraphics[width=\linewidth]{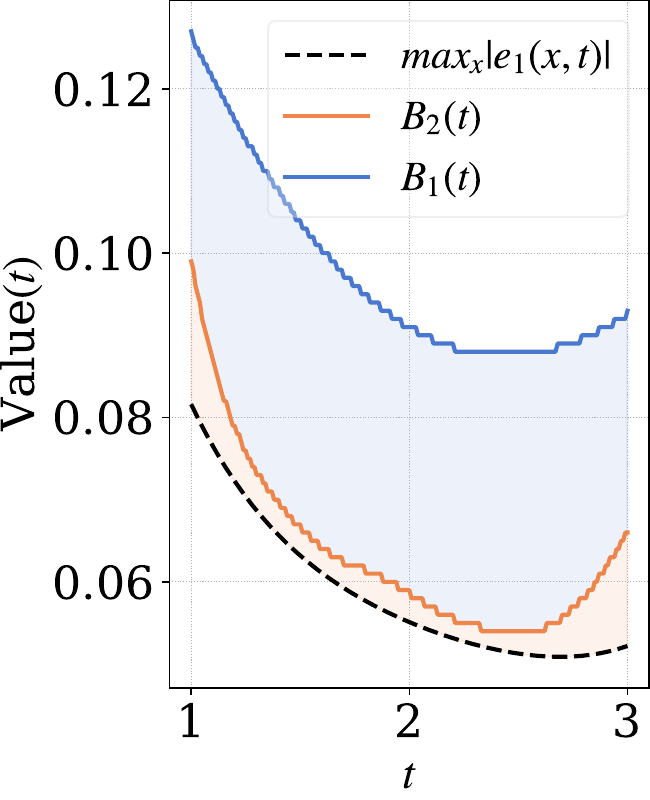}
            \caption{$e_1^*(t) < B_2(t) < B_1(t)$}
            \label{fig:sub1a}
        \end{subfigure}
        \hspace{0em}
        \begin{subfigure}{0.58\textwidth}
            \centering
            \includegraphics[width=\linewidth]{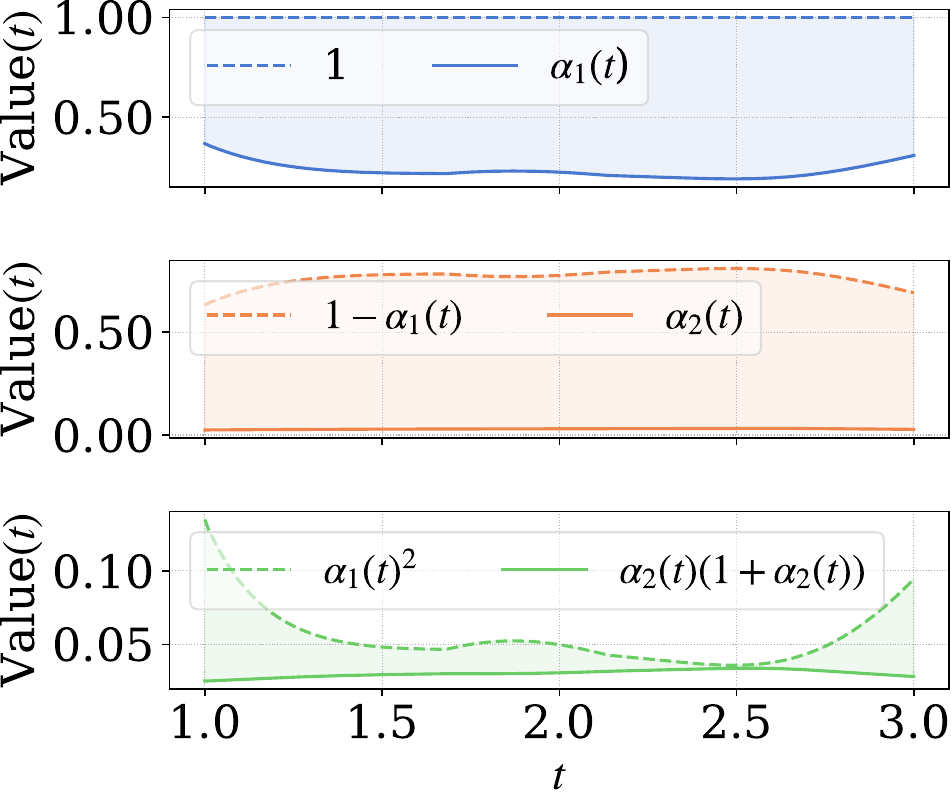}
            \caption{Satisfy Conditions~\ref{eq:alpha_conditions}.}
           \label{fig:sub1b}
        \end{subfigure}
        \caption{second-order error bound $B_2(t)$ on 1D Linear system}
        \label{fig:1}
    \end{minipage}
    \hspace{0em}
    \begin{minipage}{0.39\textwidth}
        \centering
        \includegraphics[width=1.0\linewidth]{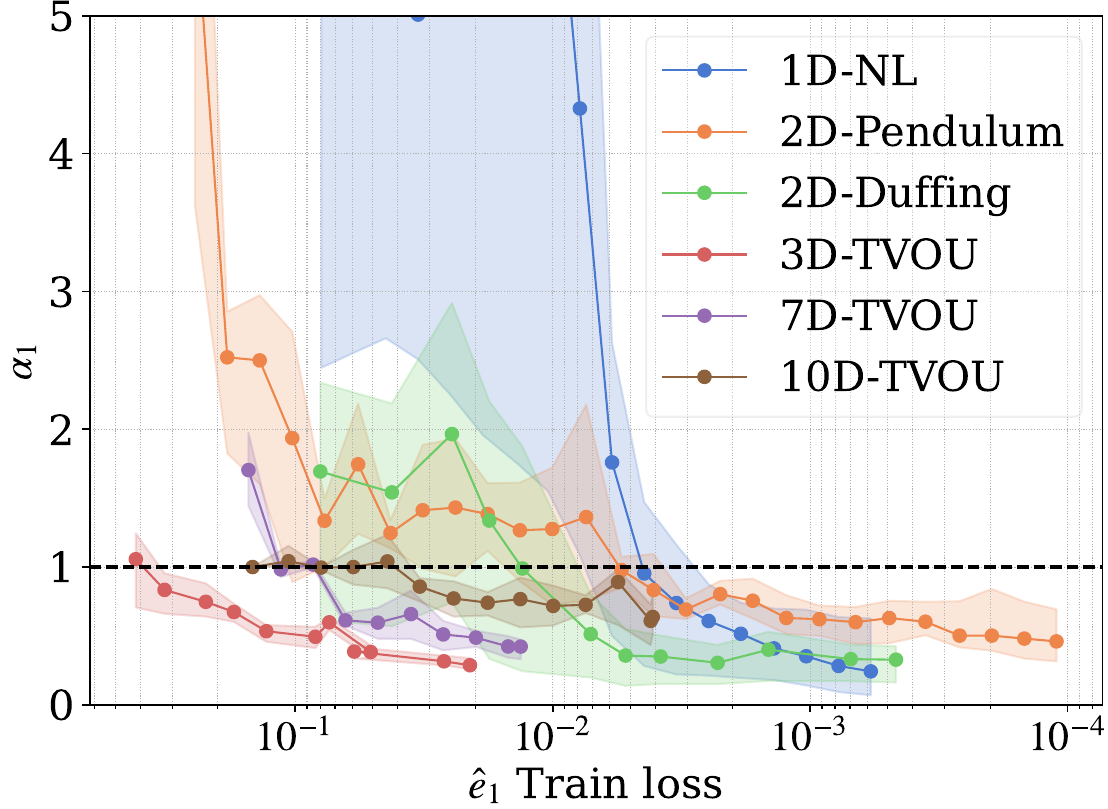}
        \caption{$\alpha_1(t),\;t \in T'$ vs train loss of $\hat{e}_1$, for all first-order error bound experiments.}
        \label{fig:meta}
    \end{minipage}
\end{figure*}

\paragraph{Training Scheme for First-Order Error Bound}\label{sec:train_scheme}
Guided by Proposition~\ref{prop:checking_alpha1}, our goal is to train $\hat{e}_1$ to achieve sufficiently small loss.
By the PINN loss in Eq.~\eqref{eq:pinn_loss} and the PDE of the first error in Eqs.~\eqref{eq:err_def} and~\eqref{eq:error_1_e1_func_pde}, 
the training loss of $\hat{e}_1$
is:
\begin{subequations}\label{eq:pinn_loss_e1}
    \begin{align} 
    \mathcal{L}^{(1)} & = w_{0}\mathcal{L}_{0}^{(1)} + w_{r}\mathcal{L}_{r}^{(1)}, \\ 
    \label{eq:pinn_loss_0_e1}
    \mathcal{L}_{0}^{(1)} 
    & = \frac{1}{N_0}\sum_{k=1}^{N_0}\big( p_0(x_k)-\hat{p}(x_k,0) - \hat{e}_1(x_k,0) \big)^2, \\
    \label{eq:pinn_loss_r_e1}
    \mathcal{L}_{r}^{(1)} 
    & = \frac{1}{N_r}\sum_{k=1}^{N_r}\big(\mathcal{D}[\hat{e}_1(x_k,t_k)] + \mathcal{D}[\hat{p}(x_k,t_k)]\big)^2.
    \end{align}
\end{subequations}
By Eq.~\eqref{eq:pinn_loss_0_e1}--\eqref{eq:pinn_loss_r_e1}, we see that training $\hat{e}_1$ requires inputs from neural network $\hat{p}$ and its derivatives $\mathcal{D}[\hat{p}]$.
This could lead to difficult training for $\hat{e}_1$ if the input $\mathcal{D}[\hat{p}]$ is highly oscillating even when $\hat{p}$ is smooth by construction~\citep{zhao2023pinnsformer}.
To address this issue, we implement a regularization loss to prevent rapid changes in $\mathcal{D}[\cdot]$ of the first PINN $\hat{p}$. Specifically, we train $\hat{p}$ by adding the following regularization loss to Eq.~\eqref{eq:pinn_loss}: 
\begin{equation}\label{eq:grad_reg}
\mathcal{L}_{\nabla}=\frac{1}{N_r}\sum_{k=1}^{N_r}\| \nabla \Big( \mathcal{D}[\hat{p}(x_k,t_k)] \Big) \|^2_2,
\end{equation}
where $\nabla$ is the gradient operator and $\|\cdot\|_2$ is the L2 norm.
We note that this regularization loss does not violate the paradigm of physics-informed learning, because as the residual of $\mathcal{D}[\cdot] \rightarrow 0$ for all $(x,t) \in \Omega$, the gradient of the differential residual $\nabla(\mathcal{D}[\cdot])$ also converges to zero.
In fact, this regularization is investigated in \citet{yu2022gradient} to improve training stability of PINNs. 
Note that such gradient regularization loss in Eq.~\eqref{eq:grad_reg} is not applied to the training of $\hat{e}_1$ in Eqs.~\eqref{eq:pinn_loss_e1} because $\mathcal{D}[\hat{e}_1]$ is not used to train subsequent error functions.
A detailed description of our training scheme is provided in Appendix~\ref{appendix:train_and_results}.


\begin{remark}\label{remark:3}
    Finally, we note that, while the presented approach focuses on FP-PDE and training an approximate PDF $\hat{p}$ and bounding its error, the only essential requirement is that the FP-PDE operator $\mathcal{D}[\cdot]$ is linear. Therefore, this approach naturally extends to other linear PDEs subject to initial and boundary conditions (e.g., Dirichlet and Neumann conditions). We illustrate this in a case study in Appendix~\ref{proof:1D_Heat},~\ref{appendix:1D-HEAT-sys}, and~\ref{appendix:1D-HEAT-exp}.
\end{remark}
\section{Experiment}\label{sec:experiment}


We demonstrate our proposed error bound approach via several numerical experiments. First, we illustrate a synthetic second-order error bound on a simple linear system. Then, we present our main applications of the first-order error bound for non-trivial systems including nonlinear, chaotic, and high dimensional SDEs. The details of these systems are provided in Appendix~\ref{appendix:system}.
For all the experiments, fully connected neural networks are used for both $\hat{p}(x,t)$ and $\hat{e}_1(x,t)$. 
Throughout all experiments, we employ a fixed weighting scheme for training (see Appendix \ref{appendix:train_and_results}). 
More sophisticated weight-tuning strategies (e.g., \citet{basir2023adaptive} and references therein) have been shown to enhance PINN accuracy and could further improve the results reported here.
The code implemented in Python and Pytorch is available on GitHub \citep{PINN_Error:github}. All the experiments are conducted on a MacBook Pro with Apple M2 processor and 24GB RAM.

\begin{table*}[t]
    \centering
    \caption{First-order error bound results categorized into three groups: (1) Linear, where $p$ is obtained analytically, (2) Nonlinear, where `true' $p$ is obtained by Monte-Carlo (M.C.) since no analytical solutions exist, and (3) High Dimensional, where `true' $p$ is obtained by semi-analytical numerical integration. Here, time $p$ reports the computation time \textbf{in seconds} to obtain the `true' $p$ via M.C., time $\hat{p}$ and $\hat{e}_1$ are the training time for PINNs \textbf{in seconds}, $\alpha_1^{\max}:=\max_t \alpha_1(t)$, $\alpha_1$ reports the mean and standard deviation of $\alpha_1(t)$ over $t$.
     $\text{Gap}^{\min}:=\min_t((B_1(t)-e_1^*(t))/\max_x p(x,t))$ is the minimum gap (over time) between the error bound and maximum error normalized by the true solution, and
     $B_1^{\mathcal{N}}$ reports the average and standard deviation (over time) of the normalized error bound $B_1(t)/\max_x p(x,t)$.}
    \renewcommand{\arraystretch}{0.9} 
    \setlength{\tabcolsep}{4.0pt}
    \begin{tabular}{p{1.8cm}|lcccccccc}  
      \toprule
      \bfseries Category & \bfseries Experiment & \bfseries time $p$ & \bfseries time $\hat{p}$ & \bfseries time $\hat{e}_1$ & \bfseries $\alpha_1^{\max}$ & \bfseries $\alpha_1$ & \bfseries $\text{Gap}^{\min}$ & \bfseries $B_1^{\mathcal{N}}$ \\
      \midrule
      \multirow{3}{2cm}{\textbf{Nonlinear}}  
      & 1D Nonlinear & 37634 (M.C.) & 1031 & 1840 & 0.63 & 0.24 $\pm$ 0.16 & 2e-2 & 0.12 $\pm$ 6e-2 \\
      & 2D Inverted Pendulum & 45409 (M.C.). & 1054 & 4484 & 0.69 & 0.46 $\pm$ 0.14 & 2e-2 & 0.14 $\pm$ 7e-2 \\
      \midrule
      \multirow{3}{2cm}{\textbf{High Dimensional}} 
      & 3D Time-varying OU & Semi-Analytical & 267 & 943 & 0.34 & 0.29 $\pm$ 0.03 & 2e-2 & 0.05 $\pm$ 1e-2 \\
      & 7D Time-varying OU & Semi-Analytical & 478 & 1475 & 0.49 & 0.43 $\pm$ 0.05 & 6e-2 & 0.18 $\pm$ 6e-2 \\
      & 10D Time-varying OU & Semi-Analytical & 803 & 5954 & 0.83 & 0.64 $\pm$ 0.16 & 5e-2 & 0.19 $\pm$ 8e-2 \\
      \bottomrule
    \end{tabular}
    \label{tab:main_result}
\end{table*}

\textbf{Second-order error bound illustration \quad}\label{sec:arb_tight_error_bound_example}
We consider a FP-PDE for a 1D linear SDE. This simple system has analytical PDF $p(x,t)$, which allows us to synthesize $\hat{e}_2(x,t)$ and validate second-order error bound $B_2(t)$.
Specifically, we first train two PINNs in sequence: one $\hat{p}(x,t)$ and the other for $\hat{e}_1(x,t)$. Due to the practical challenge discussed in Section~\ref{sec:theory}, we synthesize $\hat{e}_2(x,t)$ from the analytical solution and the trained PINNs, i.e., $\hat{e}_2(x,t)=p(x,t)-\hat{p}(x,t)-\hat{e}_1(x,t)+\delta(x,t)$, where $\delta(x,t)$ is a chosen sinusoidal perturbation such that Conditions~\ref{eq:alpha_conditions} are satisfied.
    With the learned $\hat{e}_1(x,t)$ and synthesized $\hat{e}_2(x,t)$, we construct  second- and first-order error bound $B_2(t)$  and $B_1(t)$ from Eqs.~\eqref{eq: temporal error bound} and \eqref{eq:tight_error_bound}, respectively. Fig.~\ref{fig:sub1a} validates that both $B_2(t)$ and $B_1(t)$ upper-bound the worst-case error for all time. Furthermore, it shows that $B_2(t)$ is tighter than $B_1(t)$, and the relative tightness $B_1(t)/B_2(t)$ is at most $1.63<2$ as predicted by Corollary~\ref{corollary:special_error_bound}. Fig.~\ref{fig:sub1b} visualizes the satisfaction of the sufficient conditions on $\alpha_1(t)$ and $\alpha_2(t)$ in Eqs.~\eqref{eq:alpha_conditions} for all $t \in T'$. These results validate the soundness of our 
second-order error bound under the proposed conditions, and its tightness relative to the first-order bound.

\textbf{First-order error bound application \quad}\label{sec:exp_tight_error_bound}
We apply the approach in Section~\ref{sec:tight_error_bound} to construct error bound $B_1(t)$ for several FP-PDEs that do not have closed-form solutions. First, we consider FP-PDEs associated with nonlinear SDEs (see rows labeled as Nonlinear in Table~\ref{tab:main_result}). We note that 2D Duffing Oscillator is from \citet{anderson2024fisher} and exhibits chaotic behaviors. Due to complexity of these systems, highly detailed Monte Carlo (M.C.) simulations with extremely small time steps and very large numbers of samples were necessary to generate high-fidelity `ground-truth' PDF distributions at specific discrete time instances (see Appendix~\ref{appendix:system}). 
The final set of experiments consider high-dimensional (up to 10-D) time-varying Ornstein-Uhlenbeck (OU) processes (see rows labeled as High Dimensional in Table~\ref{tab:main_result}). 
We note that running M.C. to obtain `ground-truth' PDF for fully nonlinear and high-dimensional systems is not tractable. Hence, we choose these time-varying OU because they are non-trivial (i.e., no closed-form solutions) but allow us to efficiently estimate accurate $p(x,t)$ via semi-analytical method that numerically integrates the Gaussian distributions by exploiting the time-varying linear dynamics \citep{sarkka2019applied}. 
Thus, unlike our PINN-based method or standard M.C. simulations, this semi-analytical approach cannot handle general nonlinear dynamics.

\begin{figure*}[t!] 
    \centering
    \begin{subfigure}{0.32\textwidth} 
        \includegraphics[width=\textwidth]{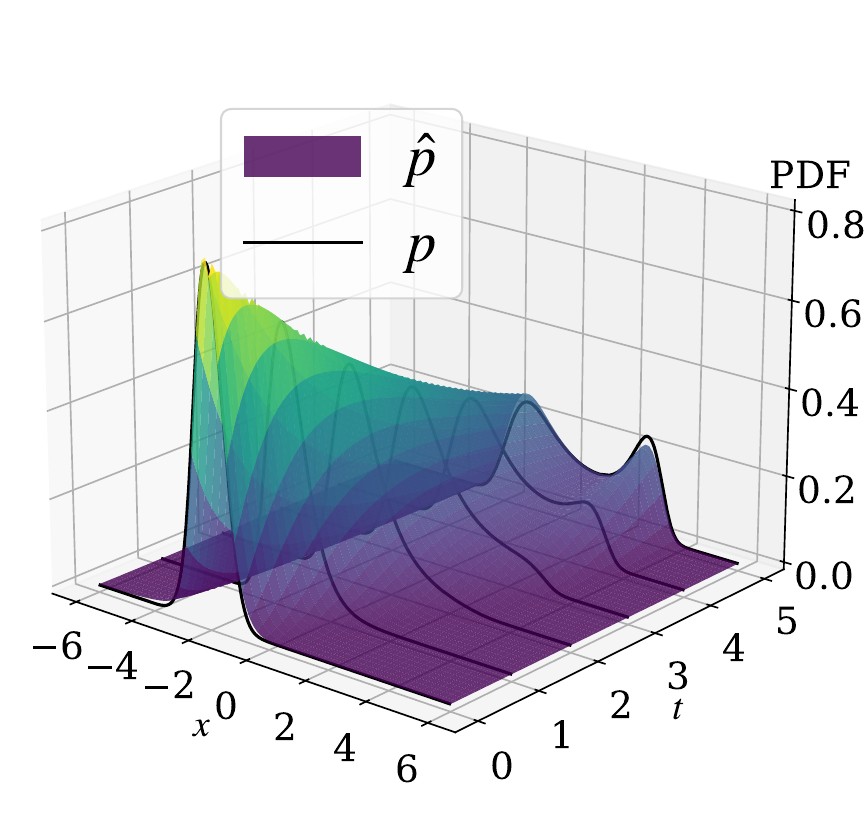}
        \caption{}
        \label{fig:1dnl_paper_p}
    \end{subfigure}
    \hfill
    \begin{subfigure}{0.32\textwidth} 
        \includegraphics[width=\textwidth]{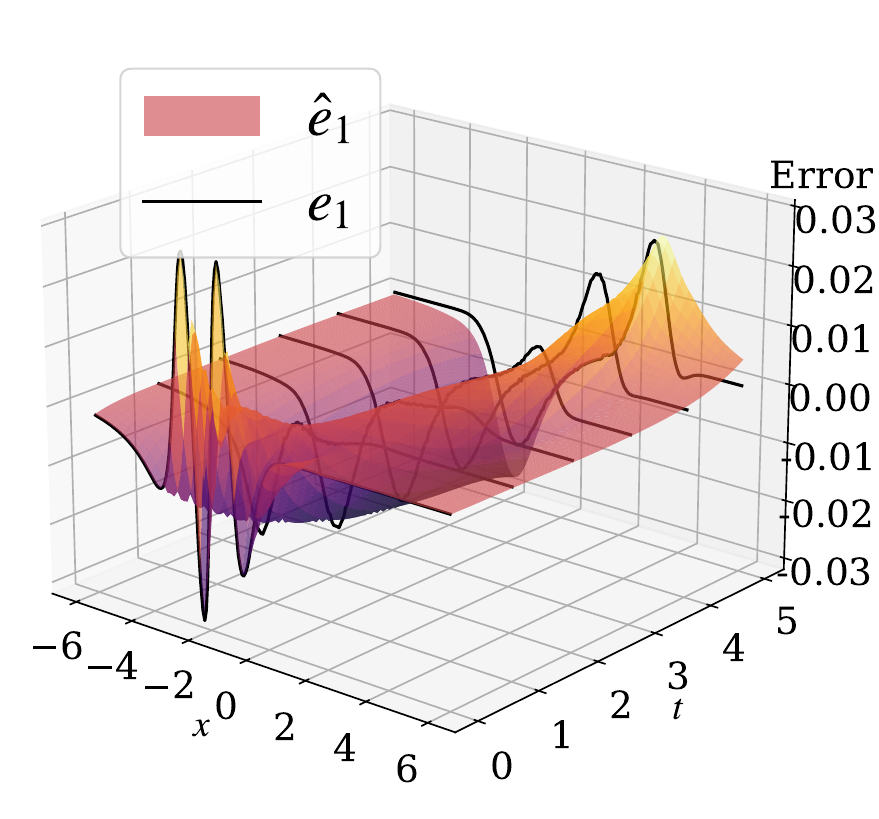}
        \caption{}
        \label{fig:1dnl_paper_e}
    \end{subfigure}
    \hfill
    \begin{subfigure}{0.32\textwidth} 
        \includegraphics[width=\textwidth]{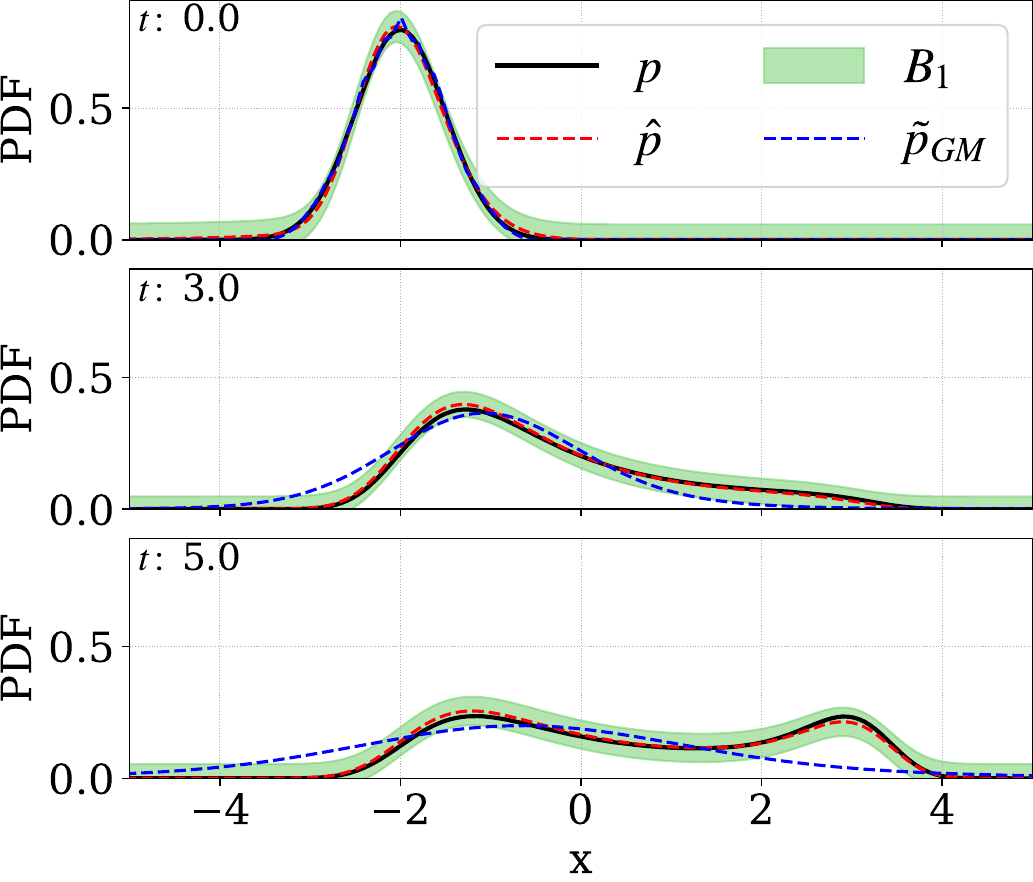} 
        \caption{}
        \label{fig:1dnl_paper}
    \end{subfigure}
    
    \begin{subfigure}{\textwidth}
        \centering
        \includegraphics[width=\textwidth]{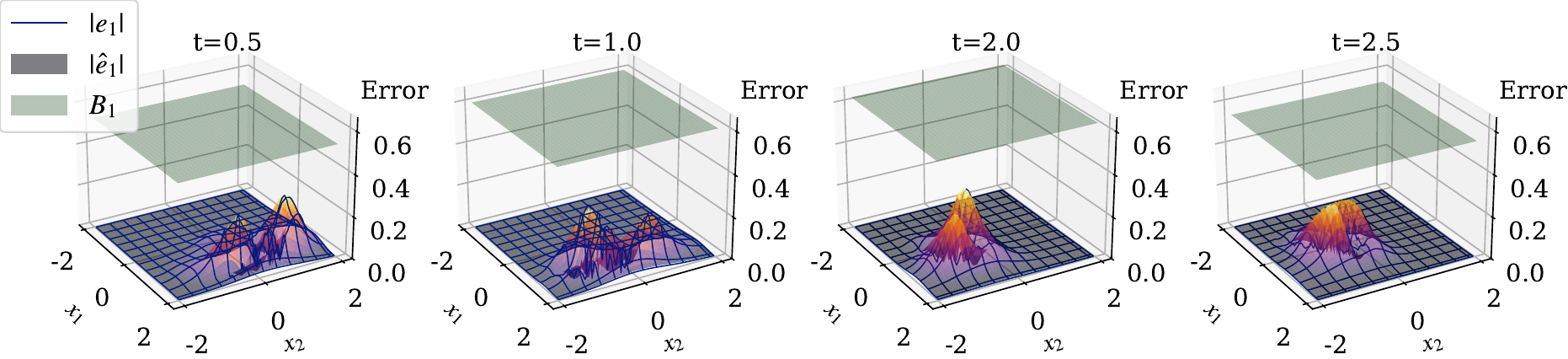}
        \caption{}
        \label{fig:2dduff_paper}
    \end{subfigure}
    
    \begin{subfigure}{0.24\textwidth}
        \includegraphics[width=\textwidth]{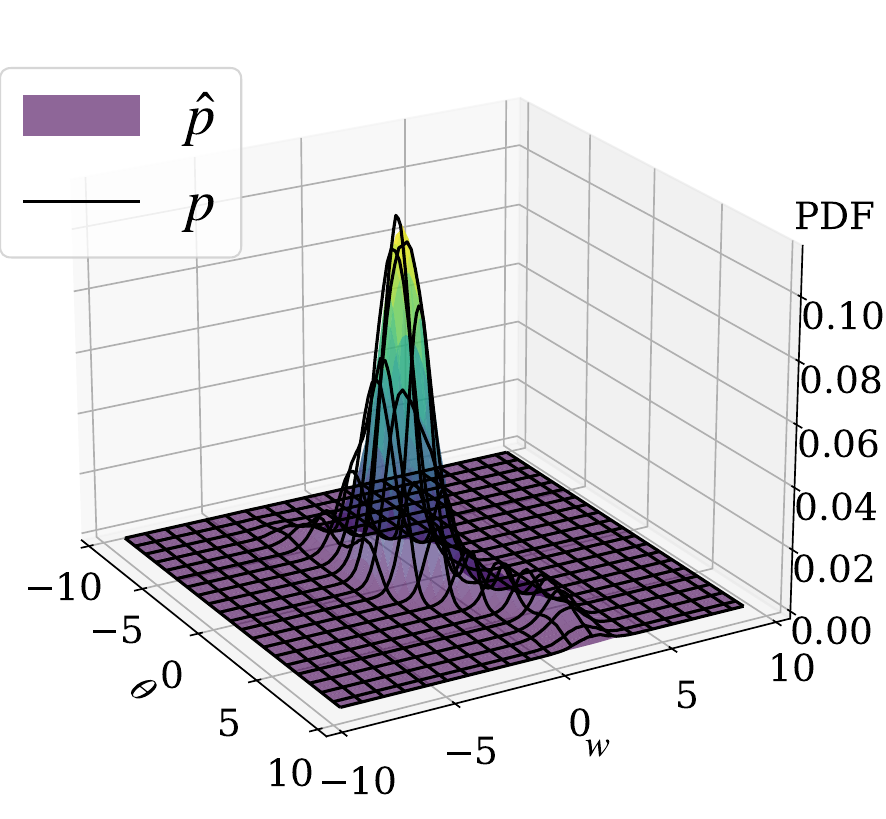} 
        \caption{}
        \label{fig:2dip_paper_1}
    \end{subfigure}
    \hfill
    \begin{subfigure}{0.24\textwidth} 
        \includegraphics[width=\textwidth]{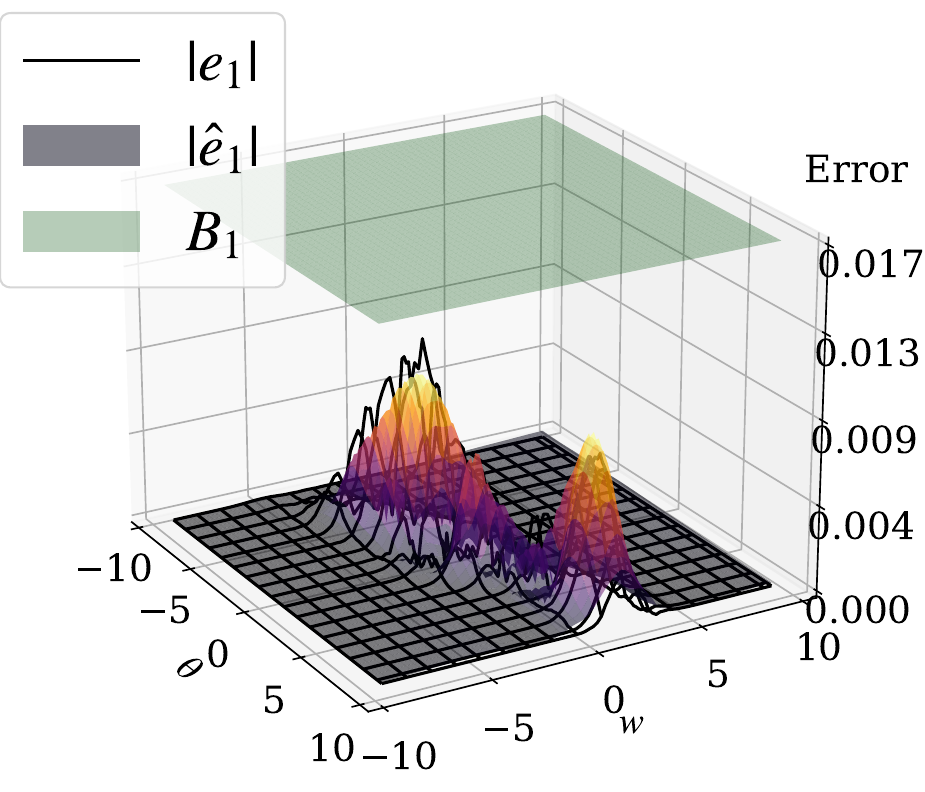} 
        \caption{}
        \label{fig:2dip_paper_2}
    \end{subfigure}
    \hfill
        \begin{subfigure}{0.24\textwidth}
        \includegraphics[width=\textwidth]{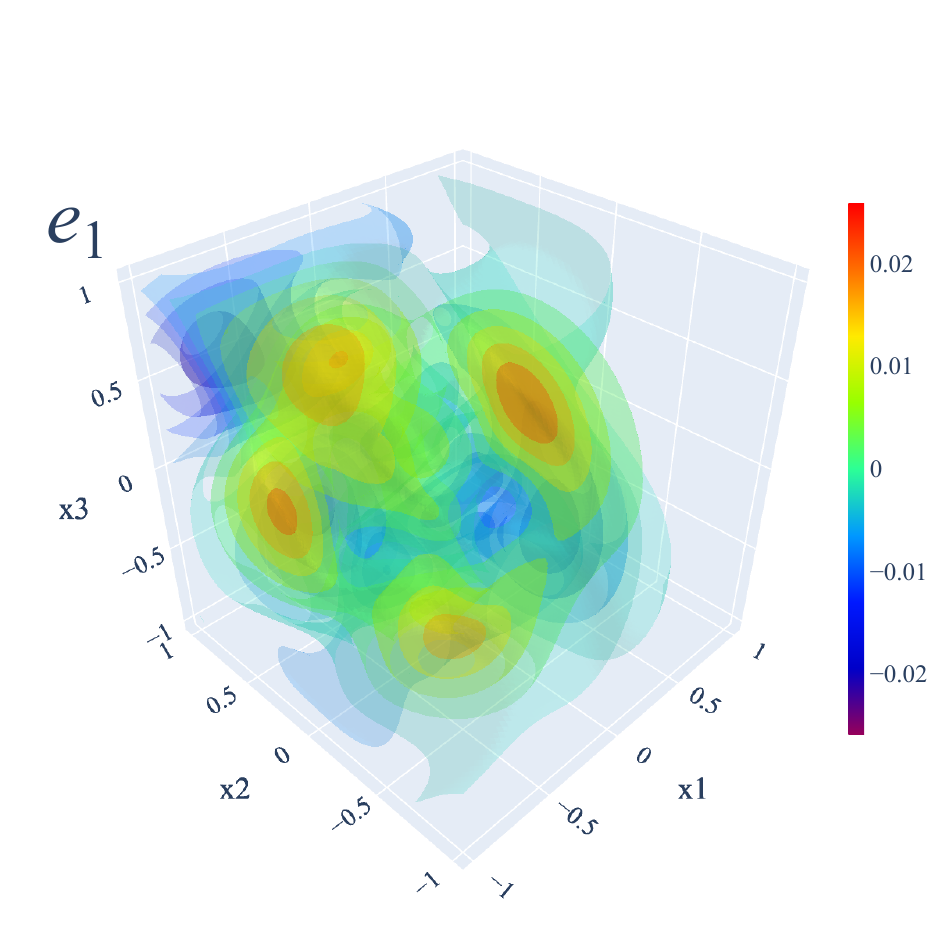}
        \caption{}
        \label{fig:3d_e1t1.0}
    \end{subfigure}
    \hfill
    \begin{subfigure}{0.24\textwidth}
        \includegraphics[width=\textwidth]{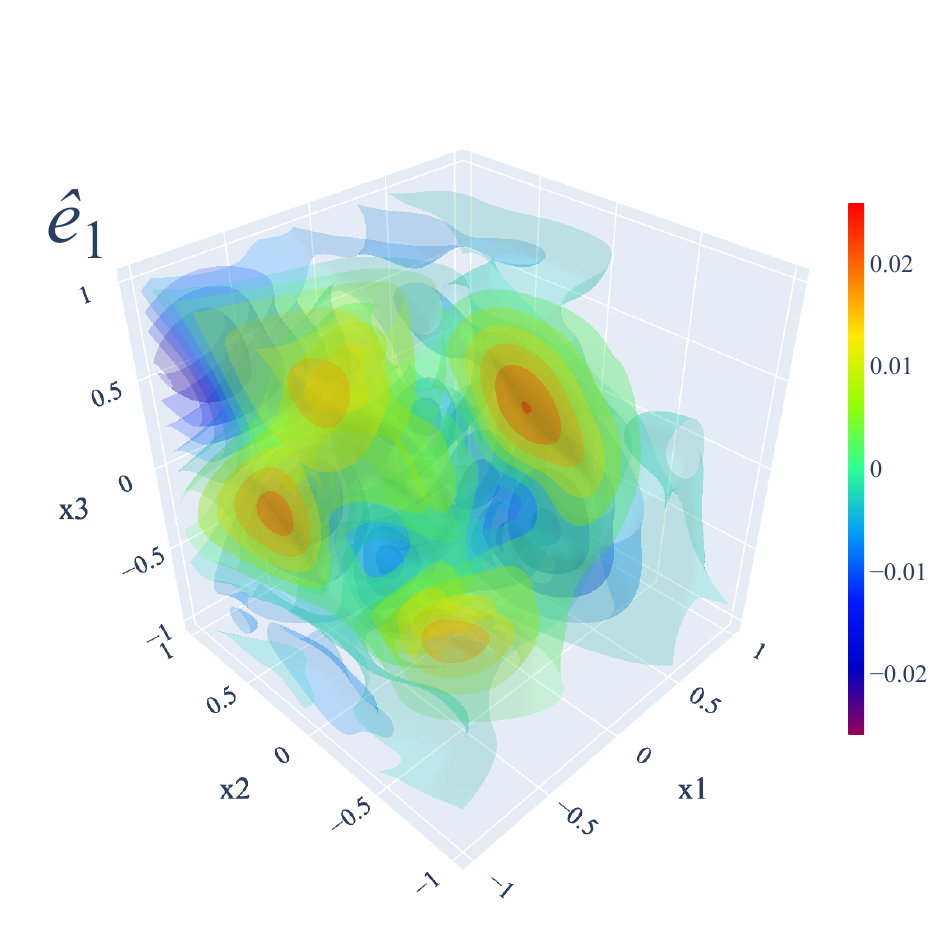}
        \caption{}
        \label{fig:3d_e1hatt1.0}
    \end{subfigure}
    \caption{First-order error bound results. (a)-(c) 1D Nonlinear PDF $p$ vs PINN $\hat{p}$, error $e_1$ vs error PINN $\hat{e}_1$, and error bound $B_1$ compared to the classical Gaussian mixture method $\hat{p}_{GM}$, illustrated at three time points. 
    (d) 2D Duffing Oscillator true error $|e_1|$, error PINN $|\hat{e}_1|$, and error bound $B_1 \geq |e_1|$ over time.
    (e)-(f) 2D Inverted Pendulum PDF $p$, PINN $\hat{p}$, true error $|e_1|$, error PINN $|\hat{e}_1|$, and error bound $B_1 \geq |e_1|$ at $t=3$. 
    (g)-(h) 3D Time-varying OU error $e_1$ and error PINN $\hat{e}_1$ at $t=1$.
    }
    \label{fig:representative_results}
\end{figure*}

\begin{figure}[htbp!]
    \centering
    \begin{subfigure}[b]{0.23\textwidth} 
        \includegraphics[width=\textwidth]{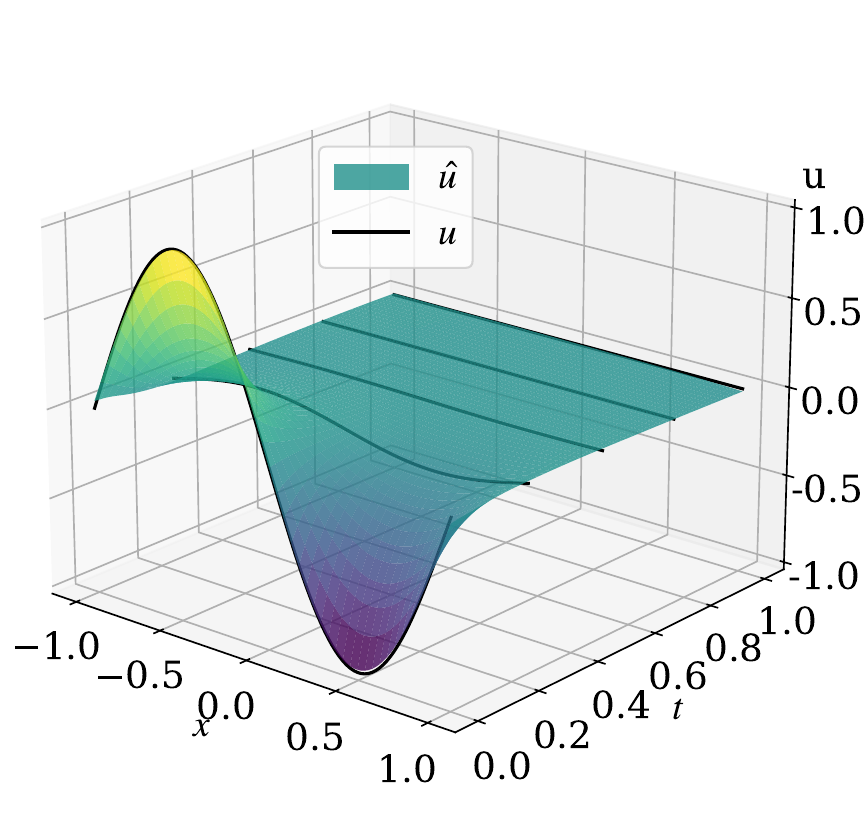}
        \caption{$\hat{u}$ vs $u$}
    \end{subfigure}
    \hfill
    \begin{subfigure}[b]{0.23\textwidth} 
        \includegraphics[width=\textwidth]{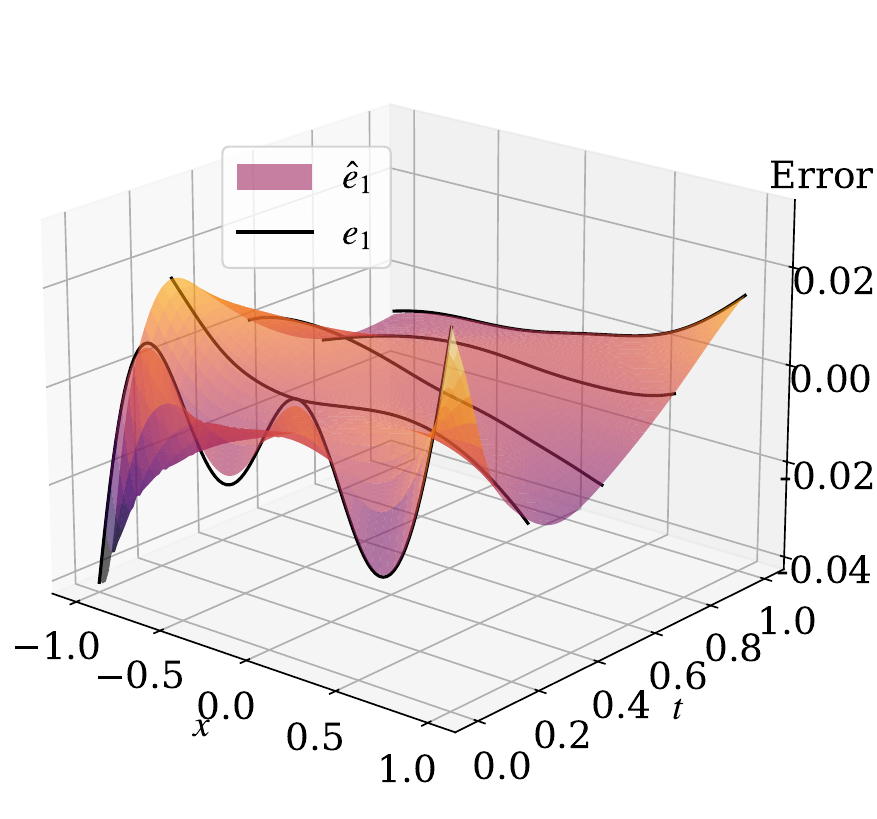}
        \caption{$\hat{e}_1$ vs $e_1$}
    \end{subfigure}
    \caption{1D Heat PINN solution $\hat{u}(x,t)$ and error $\hat{e}_1(x,t)$ v.s. true solution $u(x,t)$ and error $e_1(x,t)$.}
    \label{fig:1dheat_surfaces_main}
\end{figure}

Table~\ref{tab:main_result} summarizes the results on all systems. 
The metrics include computation time for $p$ and training times for $\hat{p}(x,t)$ and $\hat{e}_1$ PINNs in seconds. 
Also, we show training feasibility of $\hat{e}_1(x,t)$ by reporting $\alpha_1(t)$ and its maximum value over time ($\alpha^{\max}$).  To show soundness of error bound $B_1$, we calculate the minimum gap $\text{Gap}^{\min}$ between $B_1$ and the true error, i.e., positive $\text{Gap}^{\min}$ indicate correctness of $B_1$. To assess the tightness of $B_1(t)$, we report the statistics of its magnitude normalized by $\max_x p(x,t)$, denoted by $B_1^\mathcal{N}$.
Overall, the results show: 
(i) \textit{soundness \& feasibility:} $B_1$ correctly bounds the approximation error ($\text{Gap}^{\min} > 0$) and the learning condition is satisfied ($\alpha_1^{\max}<1$) though it becomes increasingly challenging to meet as dimensionality grows,
(ii) \textit{efficiency:} the time comparisons between $p$ vs $\hat{p}$ illustrate significant speedup (two orders of magnitude: $38\times$ to $65\times$) in obtaining accurate PDF with PINNs for systems that do not have analytical solutions (nonlinear systems),
and
(iii) \textit{scalability:} our PINN method is able to scale to 10-dimensional systems with fairly tight error bounds across all cases (on average 5\%-21\% with respect to the true solution). 

We plot $\alpha_1(t)$ distributions (over time) vs training loss of $\hat{e}_1$ in Fig.~\ref{fig:meta} to show that the condition in Corollary~\ref{corollary:special_error_bound} is satisfied for all systems as training loss decreases.
We visualize the error bounds for some representative cases in Fig.~\ref{fig:representative_results}. 
Specifically, Fig.~\ref{fig:1dnl_paper_p} plots the PDF $p(x,t)$ and its PINN approximation $\hat{p}(x,t)$ of the 1D Nonlinear experiment. Note that $\hat{p}(x,t)$ is a continuous surface over time and space, while $p(x,t)$ is illustrated by curves at discrete time instances according to those used in Monte-Carlo simulation.
Fig.~\ref{fig:1dnl_paper_e} shows the `true' error $e_1(x,t)$ at discrete time instances and its PINN approximation $\hat{e}_1(x,t)$ as a continuous surface of the 1D Nonlinear experiment.
Observe that both $\hat{p}$ and $\hat{e}$ closely approximate $p$ and $e$, respectively, over all space and time, respectively.
In addition, Fig.~\ref{fig:1dnl_paper} shows the PINN‐learned density $\hat p$ alongside the GMM approximation $\tilde p_{\rm GM}$ (125 Gaussian mixtures integrated over $\Delta t=0.001$ time step). 
We include the GMM---a common tool for uncertainty propagation in nonlinear dynamics \citep{archambeau2007gaussian,terejanu2008uncertainty,vittaldev2016spacecraft,10886416}---as a classical alternative.
Compared to the PINN $\hat{p}$, observe that $\tilde{p}_{GM}$ gradually deviates from $p$ as time increases.
More importantly, the true PDF $p$ lies within the first-order error bound $B_1$ (illustrated by green regions) of approximate PDF $\hat{p}$, while $\tilde{p}_{GM}$ does not provide rigorous error bound.
For the 2D Duffing Oscillator experiment, Fig.~\ref{fig:2dduff_paper} plots the time evolution of the true error, PINN error approximation, and the first-order error bound. Observe that the magnitude of the error bound does not necessarily grow over time, suggesting that the error bound may remain tight even over extended horizons.
Figs.~\ref{fig:2dip_paper_1} and~\ref{fig:2dip_paper_2} visualize PINNs results and the constructed $B_1$ of the 2D Inverted Pendulum experiment at a given time ($t=3$). 
Observe that both $\hat{p}$ and $\hat{e}_1$ closely approximate the unknown complicated distributions of $p$ and $e_1$, respectively. 
For the 3D Time-varying OU, Figs.~\ref{fig:3d_e1t1.0} and~\ref{fig:3d_e1hatt1.0} visualize the true error $e_1$ and PINN error $\hat{e}_1$ at $t=1$, showing good approximation (i.e., $\alpha_1 < 1)$ for constructing the first-order error bound $B_1$.
Lastly, PINN predictions $\hat u$ and $\hat e_1$ for the 1D Heat PDE (Fig.~\ref{fig:1dheat_surfaces_main}) closely match the true solution $u$ and error $e_1$ across space and time, respectively, demonstrating straightforward extension to other linear PDEs (Remark~\ref{remark:3}).
See Appendix~\ref{appendix:train_and_results} for complete training details and additional results of the conducted experiments.

\section{Conclusion}
We proposed a physics-informed learning method to approximate the PDF evolution governed by FP-PDE and bound its error using recursively learned PINN-based error functions. We proved that only two error terms are needed for arbitrarily tight bounds and introduced a more efficient first-order bound requiring just one error function, reducing computation while providing clear termination criteria. 
Our results validate the bounds' correctness and demonstrate significant computational speedups over Monte Carlo methods. 
We trained the solution and error PINNs separately, but joint training may work better, which is left for future work.



\begin{acknowledgements} 
    We acknowledge the anonymous reviewers for their constructive feedback. 
    This material is based upon work supported by the Air Force Research Laboratory/RVSW under Contract No. FA945324CX026. Any opinions, findings and conclusions or recommendations expressed in this material are those of the authors and do not necessarily reflect the views of the Air Force Research Laboratory/RVSW.
\end{acknowledgements}

\bibliography{uai2025-template}

\newpage

\onecolumn

\title{Error Bounds for Physics-Informed Neural Networks in Fokker-Planck PDEs\\(Supplementary Material)}
\maketitle


\appendix
\section{Proofs}

\subsection{Derivation of Definition~\ref{def:1}}
\label{sec:proof_def1}
Let $e_1(x,t) = p(x,t) - \hat{p}(x,t)$ and initialize $e_0(x,t):= p(x,t)$ and $\hat{e}_0(x,t)=\hat{p}(x,t)$. Then, Eq.~\eqref{eq:error_1_e1_func_pde} becomes Definition 1 for $i=1$: $$ \mathcal{D}[e_1(x,t)]+\mathcal{D}[\hat{e}_0(x,t)]=0, \quad \text{subject to } e_1(x,0)=e_0(x,0)-\hat{e}_0(x,0).$$ 
For $i=2$, we define $e_2(x,t):=e_1(x,t)-\hat{e}_1(x,t)$ and obtain $\mathcal{D}[e_2(x,t)]=\mathcal{D}[e_1(x,t)]-\mathcal{D}[\hat{e}_1(x,t)]$ (because $\mathcal{D}[\cdot]$ is a linear operator). Since $\hat{e}_1\neq e_1$, a residual will remain such that $$\mathcal{D}[\hat{e}_1]+\mathcal{D}[\hat{e}_0]:=r_1\neq 0.$$ Hence, we have the recursive PDE for $i=2$ (omitting $x$ and $t$ for simplicity of presentation): $$\mathcal{D}[e_2]=\mathcal{D}[e_1]-\mathcal{D}[\hat{e}_1] =(-\mathcal{D}[\hat{e}_0])-(-\mathcal{D}[\hat{e}_0]+r_1)=-r_1\implies \mathcal{D}[e_2]+r_1:=\mathcal{D}[e_2]+\sum_{j=1}^2 \mathcal{D}[\hat{e}_{j-1}]=0.$$ The derivation recursively follows for $i>2$.
In addition, following PINNs training in Eq.~\eqref{eq:pinn_loss}, the training loss of each $\hat{e}_i$ is:
\begin{subequations}
\begin{align} 
& \mathcal{L}^{(i)} = w_{0}\mathcal{L}_{0}^{(i)} + w_{r}\mathcal{L}_{r}^{(i)}, \;w_0,w_r \in \mathbb{R}_{>0}, \\ 
& \mathcal{L}_{0}^{(i)} = \frac{1}{N_0}\sum_{k=1}^{N_0}\Big( e_i(x_k,0) - \hat{e}_i(x_k,0) \Big)^2, \\
& \mathcal{L}_{r}^{(i)} = \frac{1}{N_r}\sum_{k=1}^{N_r}\Big(\mathcal{D}[\hat{e}_i(x_k,t_k)] + \sum_{j=1}^i \mathcal{D}[\hat{e}_{j-1}(x_k,t_k)]\Big)^2.
\end{align}
\label{eq:pinn_loss_general}
\end{subequations}
    
\subsection{Proof of Lemma~\ref{lemma:bouding series of ratios}}\label{proof:lemma1}
\begin{proof}
    From Definition~\ref{def:1}, we have that, for all $x \in X'$,
    \begin{align*}
        |p(x,t)-\hat{p}(x,t)| &= \left| \sum_{i=1}^{n} \hat{e}_i(x,t) + e_{n+1}(x,t) \right| 
        \leq \sum_{i=1}^{n} |\hat{e}_i(x,t)| + |e_{n+1}(x,t)| \notag \\
        & \leq \sum_{i=1}^{n} \max_x|\hat{e}_i(x,t)| + \max_x|e_{n+1}(x,t)| := \sum_{i=1}^{n} \hat{e}_i^*(t) + e_{n+1}^*(t).
        \label{eq:error_bound_by_max_error_series}
    \end{align*}
    From the definition of $\gamma_{\frac{i+1}{i}}$ in Eq.~\eqref{eq:approx ratio}, we obtain (omitting $t$ for simplicity of presentation)
    \begin{align*}
        & |p(x,\cdot)-\hat{p}(x,\cdot)| \leq \hat{e}_1^*\Big( 1 + \frac{\hat{e}_{2}^*}{\hat{e}_{1}^*} + \frac{\hat{e}_{3}^*}{\hat{e}_{1}^*} + \dots + \frac{\hat{e}_{n}^*}{\hat{e}_{1}^*} + \frac{e_{n+1}^*}{\hat{e}_1^*}
        \Big) \notag \\
        & = \hat{e}_1^* \Big[ 1 + \gamma_{\frac{2}{1}} + \gamma_{\frac{2}{1}}\gamma_{\frac{3}{2}} + \dots + 
        (\gamma_{\frac{2}{1}}\gamma_{\frac{3}{2}}\dots \gamma_{\frac{n}{n-1}}) + (\gamma_{\frac{2}{1}}\gamma_{\frac{3}{2}} \dots \gamma_{\frac{n-1}{n-2}} \gamma_{\frac{n}{n-1}} \frac{e_{n+1}^*}{\hat{e}_n^*}) \Big] \notag \\
        & = \hat{e}_1^* \Big[ 1 + \gamma_{\frac{2}{1}} + \gamma_{\frac{2}{1}}\gamma_{\frac{3}{2}} + \dots + 
        (\gamma_{\frac{2}{1}}\gamma_{\frac{3}{2}}\dots \gamma_{\frac{n}{n-1}}) + (\gamma_{\frac{2}{1}}\gamma_{\frac{3}{2}} \dots \gamma_{\frac{n-1}{n-2}}  \frac{\hat{e}_{n}^*}{\hat{e}_{n-1}^*} \frac{e_{n+1}^*}{\hat{e}_n^*}) \Big] \notag \\
        & = \hat{e}_1^* \Big[ 1 + \gamma_{\frac{2}{1}} + \gamma_{\frac{2}{1}}\gamma_{\frac{3}{2}} + \dots + 
        (\gamma_{\frac{2}{1}}\gamma_{\frac{3}{2}}\dots \gamma_{\frac{n}{n-1}}) + (\gamma_{\frac{2}{1}}\gamma_{\frac{3}{2}} \dots \gamma_{\frac{n-1}{n-2}}  \frac{e_{n+1}^*}{\hat{e}_{n-1}^*}) \Big].
    \end{align*}
\end{proof}

\subsection{Proof of Lemma~\ref{lemma:bounds_on_ratios}}
\begin{proof}
From Definition~\ref{def:1}, we have, for $i \geq 0$,
\begin{equation}
    e_{i}(x,t) = \hat{e}_{i}(x,t) + e_{i+1}(x,t).
    \label{eq:error_recursive_relation}
\end{equation}
By taking the maximum on the absolute value of Eq.~\eqref{eq:error_recursive_relation}, 
we get
\begin{equation}
    \max_{x} |e_{i}(x,t)| \leq \max_{x} |\hat{e}_{i}(x,t)| + \max_{x} |e_{i+1}(x,t)|. \label{eq:basic_inequalities_1}
\end{equation}
Similarly, from Eq.~\eqref{eq:error_recursive_relation}, we obtain
\begin{align}
    &\hat{e}_{i}(x,t) = e_{i}(x,t) - e_{i+1}(x,t) \implies \notag \\
    & \max_{x} |\hat{e}_{i}(x,t)| \leq \max_{x} |e_{i}(x,t)| + \max_{x} |e_{i+1}(x,t)|.
    \label{eq:basic_inequalities}
\end{align}
Now take $2 \leq i < n$, and suppose the corresponding $\alpha_i(t)<1$. Then, we can write the two inequalities in Eqs.~\eqref{eq:basic_inequalities_1} and \eqref{eq:basic_inequalities} with the definition of $\hat{e}_i^*(t)$ in Eq.~\eqref{eq:max_error} and the expression in Eq.~\eqref{eq:assume_alpha_relation} as
\begin{align}
\left\{
    \begin{aligned}
        & \alpha_{i-1}(t) \hat{e}_{i-1}^*(t) \leq \hat{e}_i^*(t) + \alpha_i(t) \hat{e}_i^*(t) \\
        & \hat{e}_i^*(t) \leq \alpha_{i-1}(t) \hat{e}_{i-1}^*(t) + \alpha_i(t) \hat{e}_i^*(t).
    \end{aligned}
\right.
\label{eq:e1e2_inequality}
\end{align}
By rearranging Eq.~\eqref{eq:e1e2_inequality}, we obtain the lower and upper bounds of $\gamma_{\frac{i}{i-1}}(t)$:
\begin{equation}
    \frac{\alpha_{i-1}(t)}{1+\alpha_i(t)} \leq \frac{\hat{e}_i^*(t)}{\hat{e}_{i-1}^*(t)} = \gamma_{\frac{i}{i-1}(t)} \leq \frac{\alpha_{i-1}(t)}{1-\alpha_i(t)},\; 2 \leq i < n,
    \label{eq:ratio_and_alphas_general}
\end{equation}
which is well defined because the denominator $\hat{e}_{i-1}^* > 0$ by Assumption~\ref{assumption:ehat_bigger_than_zero}, and the (RHS) of Eq.~\eqref{eq:ratio_and_alphas_general} is always $\geq$ the (LHS) of Eq.~\eqref{eq:ratio_and_alphas_general} if $0\leq \alpha_i(t) < 1$ for all $2 \leq i < n$.
\end{proof}

\subsection{Proof of Lemma~\ref{lemma:exist_decrease_ratio}}
\begin{proof}
    For simplicity of presentation, we omit writing the dependent variable $t$. 
    Assume the conditions in Eq.~\eqref{eq:alpha_conditions} are satisfied; then it is true that $0 < \alpha_2 < 1$.
    Since both $\alpha_1,\alpha_2 < 1$, by Lemma~\ref{lemma:bounds_on_ratios} and Condition~\eqref{eq:alpha2_condition1}, we obtain 
    $$\gamma_{\frac{2}{1}} \leq \frac{\alpha_1}{1-\alpha_2} < 1, $$
    proving the RHS of Eq.~\eqref{eq:decrease_ratio}.

    For the LHS of Eq.~\eqref{eq:decrease_ratio}, let $\alpha_{i} \leq \alpha_2$ for all $2 < i < n$.
    Since $\alpha_2 < 1$, then by Lemma~\ref{lemma:bounds_on_ratios}, we have
    \begin{align}
        \gamma_{\frac{i}{i-1}} \leq \frac{\alpha_{i-1}}{1-\alpha_{i}} 
        \leq \frac{\alpha_{i-1}}{1-\alpha_2} 
        \leq \frac{\alpha_2}{1-\alpha_2}. \label{eq: mid step}
    \end{align}
    What remains is to show that RHS of Eq.~\eqref{eq: mid step} is $<\gamma_{\frac{2}{1}}$.  
    From Condition~\eqref{eq:alpha2_condition2}, we have
    \begin{align}
        \alpha_2 (1+\alpha_2) &< \alpha_1^2 \\
        & < \alpha_1 (1-\alpha_2), \label{eq: mid step 2}
    \end{align}
    where Eq.~\eqref{eq: mid step 2} holds by Condition~\eqref{eq:alpha2_condition1}.  From Eq.~\eqref{eq: mid step 2}, we obtain
    \begin{align}
        \frac{\alpha_2}{1-\alpha_2} < \frac{\alpha_1}{1+\alpha_2}. \label{eq: mid step 3}
    \end{align}
    By combining Eqs. \eqref{eq: mid step} and \eqref{eq: mid step 3}, we have
    \begin{equation*}
        \gamma_{\frac{i}{i-1}} < \frac{\alpha_1}{1+\alpha_2} < \gamma_{\frac{2}{1}}, \; 2 < i < n.
    \end{equation*}
\end{proof}

\subsection{Proof of Theorem~\ref{theorem:temporal_error_bound}}
\begin{proof} 
    Take $n\rightarrow \infty$ for Lemma~\ref{lemma:bouding series of ratios}, and
    train $\hat{e}_1$ and $\hat{e}_2$ such that the sufficient conditions of Eq.~\eqref{eq:alpha_conditions} are met, therefore, $\gamma_{\frac{3}{2}},\gamma_{\frac{4}{3}},\dots, \gamma_{\frac{n-1}{n-2}} < \gamma_{\frac{2}{1}}<1$ by Lemma~\ref{lemma:exist_decrease_ratio}. Then we have
    \begin{align}
        & |p(x,t)-\hat{p}(x,t)| \notag \\
        &\leq \hat{e}_1^* \lim_{n\rightarrow \infty} \Big( 1 + \gamma_{\frac{2}{1}} + \gamma_{\frac{2}{1}}\gamma_{\frac{3}{2}} + \dots + \Big[ \gamma_{\frac{2}{1}}\gamma_{\frac{3}{2}}\dots \gamma_{\frac{n-1}{n-2}} \gamma_{\frac{n}{n-1}} \Big] + \Big[ \gamma_{\frac{2}{1}}\gamma_{\frac{3}{2}}\dots \gamma_{\frac{n-1}{n-2}}\frac{e_{n+1}^*}{\hat{e}_{n-1}^*} \Big] \Big) \notag \\
        &= \hat{e}_1^* \lim_{n\rightarrow \infty} \Big( 1 + \gamma_{\frac{2}{1}} + \gamma_{\frac{2}{1}}\gamma_{\frac{3}{2}} + \dots + \Big[ \gamma_{\frac{2}{1}}\gamma_{\frac{3}{2}}\dots \gamma_{\frac{n-1}{n-2}} \frac{\hat{e}_n^*}{\hat{e}_{n-1}^*} \Big] + \Big[ \gamma_{\frac{2}{1}}\gamma_{\frac{3}{2}}\dots \gamma_{\frac{n-1}{n-2}}\frac{e_{n+1}^*}{\hat{e}_{n-1}^*} \Big] \Big) \notag \\
        &\leq \hat{e}_1^* \lim_{n\rightarrow \infty} \Big( 1 + \gamma_{\frac{2}{1}} + \gamma_{\frac{2}{1}}^2 + \dots + \gamma_{\frac{2}{1}}^{n-2} + \Big[ \gamma_{\frac{2}{1}}^{n-2}\frac{\hat{e}_n^*}{\hat{e}_{n-1}^*} \Big] + \Big[ \gamma_{\frac{2}{1}}^{n-2} \frac{e_{n+1}^*}{\hat{e}_{n-1}^*} \Big] \Big) \notag \\
        &= \Big[ \hat{e}_1^* \lim_{n\rightarrow \infty} \Big( 1 + \gamma_{\frac{2}{1}} + \gamma_{\frac{2}{1}}^2 + \dots + \gamma_{\frac{2}{1}}^{n-2} \Big) \Big] + \Big[ \hat{e}_1^* \lim_{n\rightarrow \infty} \Big( \gamma_{\frac{2}{1}}^{n-2}\frac{(\hat{e}_n^* + e_{n+1}^*)}{\hat{e}_{n-1}^*} \Big) \Big]. 
        \label{eq:gamma_series_bound}
    \end{align}
    The first term in Eq.~\eqref{eq:gamma_series_bound} forms a geometric series, and the second term in Eq.~\eqref{eq:gamma_series_bound} is zero as $n$ goes to infinity, because $\hat{e}_1^*, \hat{e}_{n-1}^*, \hat{e}_n^*, e_{n+1}^*$ are bounded by construction and $\hat{e}_{n-1}^* > 0$ by Assumption~\ref{assumption:ehat_bigger_than_zero}. Hence,
    \begin{align}
        |p(x,t)-\hat{p}(x,t)| 
        \leq 
        \hat{e}_1^* \Big( \frac{1}{1-\gamma_{\frac{2}{1}}}(t) \Big) := B_2(t).
        \label{eq:geo_series}
    \end{align}
\end{proof}

\subsection{Proof of Theorem~\ref{theorem:error_bound_tightness}}
\label{proof:arbitrary_tight}
\begin{proof}
We omit the time variable $t$ in this proof for readability. By Definition~\ref{def:1}, the maximum approximation error $\max_x|e_1(x,\cdot)| := e_1^*$.
Using the relations of $\hat{e}_1 = e_1 - e_2, \hat{e}_1^* \leq e_1^* + e_2^*$, the error bound in Theorem~\ref{theorem:temporal_error_bound} can be upper-bounded by
\begin{align}
    B_2 = \hat{e}_1^* \Big(\frac{1}{1-\hat{e}_2^*/\hat{e}_1^*} \Big) \leq (e_1^* + e_2^*) \Big(\frac{1}{1-\hat{e}_2^*/\hat{e}_1^*} \Big).
\end{align}
Hence, the gap between $B_2$ and the maximum approximation error $e_1^*$ is upper-bounded by
\begin{align}
    B_2 - e_1^* \leq e_1^* \Big(\frac{1}{1-\hat{e}_2^*/\hat{e}_1^*} - 1 \Big) + e_2^* \Big(\frac{1}{1-\hat{e}_2^*/\hat{e}_1^*} \Big).
    \label{eq:derive_arbitrary_tight_1}
\end{align}
Now suppose $\hat{e}_1$ approximates $e_1$ sufficiently well such that $e_2(x,t)=e_1(x,t)-\hat{e}_1(x,t):=\delta(x,t)$, where $\delta(x,t)$ denotes a sufficiently small function for all $(x,t) \in \Omega$. Furthermore, suppose $\hat{e}_2$ approximates $e_2$ sufficiently well such that $\hat{e}_2(x,t) \rightarrow e_2(x,t) = \delta(x,t)$ for all $(x,t) \in \Omega$. 
Define $\delta^* := \max_x|\delta(x,\cdot)|$, then $\hat{e}_2^* \rightarrow \delta^*$, and $\delta^* \rightarrow 0$ as $\delta(x,t) \rightarrow 0$ for all $(x,t) \in \Omega$. Consequently, the RHS of Eq.~\eqref{eq:derive_arbitrary_tight_1}, at the limit, becomes
\begin{align}
    & \lim_{\hat{e}_2^* \rightarrow \delta^*, \delta^* \rightarrow 0} \Big[ e_1^* \Big(\frac{1}{1-\hat{e}_2^*/\hat{e}_1^*} - 1 \Big) + e_2^* \Big(\frac{1}{1-\hat{e}_2^*/\hat{e}_1^*} \Big) \Big] \notag \\
    & =  \lim_{\delta^* \rightarrow 0} \Big[ e_1^* \Big(\frac{1}{1-\delta^*/\hat{e}_1^*} - 1 \Big) + \delta^* \Big(\frac{1}{1-\delta^*/\hat{e}_1^*} \Big) \Big] = \delta^*
\end{align}
Lastly, for every $\epsilon \in (0,\infty)$, take $\delta^*$ to be smaller than $\epsilon$, then the proof is completed.
\end{proof}

\subsection{Proof of Corollary~\ref{corollary:special_error_bound}}\label{proof:coro1}
\begin{proof}
    For all $t\in T'$, let $0 < \alpha_1(t) <1$. 
    First, $e_1 = \hat{e}_1 + e_2$ by Def.~\ref{def:1}, which implies
    \begin{equation}\label{eq:40}
        |e_1(x,t)| \leq \max_x|\hat{e}_1(x,t)| + \max_x|e_2(x,t)|
    \end{equation}
    for all $x\in X'$.
    Then by $0< \alpha_1 < 1$ and its definition in Eq.~\eqref{eq:assume_alpha_relation}, we have
    \[
    \alpha_1 \max_x|\hat{e}_1(x,t)| := \max_x|e_1(x,t)-\hat{e}_1(x,t)| = \max_x|e_2(x,t)|.
    \]
    Hence, Eq.~\eqref{eq:40} becomes
    \begin{align*}
        |e_1(x,t)| & \leq \max_x|\hat{e}_1(x,t)| + \alpha_1 (t)\max_x|\hat{e}_1(x,t)| \\
        &= \max_x|\hat{e}_1(x,t)|(1+\alpha_1(t)) \\
        & < 2 \hat{e}_1^*(t) := B_1(t)
    \end{align*}
    It is clear that $B_1(t)$ is not arbitrary tight because of the constant 2.
\end{proof}

\subsection{Proof of Proposition~\ref{prop:checking_alpha1}}\label{proof:prop1}
\begin{proof}
    Let $x \in \mathbb{R}^n$. By \cite[theorem 2.6]{mishra2023estimates}, we know
    \begin{align}
        & \varepsilon_{G} := \|e_1-\hat{e}_1\|_{W^{1,q}} \leq C_{pde}\mathcal{L}^{(1)} + C_{pde}C_{quad}^{\frac{1}{q}}N^{\frac{-\beta}{q}},
        \label{eq:total_error_bound_norms}
    \end{align}
    where $\mathcal{L}^{(1)}$ is the training loss of $\hat{e}_1$, $C_{pde}>0$ is the stability estimate of the first error PDE associated with the $W^{1,q}$ norm ($q \geq 2$), and $C_{quad},\beta >0$ are the constants according to the quadrature sampling points. By Definition~\ref{def:1}, $e_2 = e_1 - \hat{e}_1$, and since $e_1(x,t),\hat{e}_1(x,t)$ and their first derivatives are bounded over the considered domain of Problem~\ref{prob:1},
    we know there exists a universal embedding constant $C_{embed}$ \citep{mizuguchi2017estimation} such that 
    \begin{equation}
        |e_2(x,t)| \leq C_{embed} \|e_2(x,t)\|_{W^{1,q}}.
    \end{equation}
    Hence, we have
    \begin{equation}
        |e_2(x,t)| \leq C_{embed} \Big( C_{pde}\mathcal{L}^{(1)} + C_{pde}C_{quad}^{\frac{1}{q}}N^{\frac{-\beta}{q}} \Big).
    \end{equation}
    Using the definition of $\alpha_1(t) := \frac{\max_x|e_2(x,t)|}{\hat{e}_1^*(t)}$, we obtain
    \begin{align}
        \alpha_1(t) & \leq \frac{\max_x|e_2(x,t)|}{\min_t \hat{e}_1^*(t)} \notag \\
        & \leq \frac{1}{\min_t \hat{e}_1^*(t)} \Big[  C_{embed} \Big( C_{pde}\mathcal{L}^{(1)} + C_{pde}C_{quad}^{\frac{1}{q}}N^{\frac{-\beta}{q}} \Big) \Big].
    \end{align}
\end{proof}

\subsection{Derivation of Extension to Heat PDE with Dirichlet Boundary Condition}\label{proof:1D_Heat}
Here, we take heat equation for example. The governing partial differential equation of solution $u:\Omega = (\mathbb{R}^n \times [0,t_f]) \rightarrow \mathbb{R}$ is
\begin{align*}
    \frac{\partial u(x,t)}{\partial t} = \Delta [u(x,t)],
\end{align*}
subject to initial and Dirichlet boundary conditions
\begin{align*}
& u(x,0) = u_{ic}(x), \\
& u(x,t) = u_{bc}(x,t),\; (x,t) \in \partial \Omega,
\end{align*}
where $\partial \Omega$ is the boundary, and $\Delta[\cdot]:=\sum_i^n \frac{\partial ^2}{\partial x_i^2}[\cdot]$. Define the heat differential operator $\mathcal{D}_h[\cdot] := \frac{\partial}{\partial t}[\cdot] - \Delta[\cdot]$. 
By adding the boundary constraints into the training loss in Eq.~\eqref{eq:pinn_loss}, which is common in standard PINNs \citep{sirignano2018dgm}, we can train $\hat{u}$ that approximates the solution $u$.
Define the approximation error $e_1 = u - \hat{u}$, then a trained $\hat{u}(x,t)$ yields
\begin{align*}
 & \mathcal{D}_h[\hat{u}] = r_1(x,t), \\
 & \hat{u}(x,0)=u_{ic}(x)-e_{1,ic}(x,0), \\
 & \hat{u}(x,t) = u_{bc}(x,t) - e_{1,bc}(x,t),\; (x,t) \in \partial \Omega.
\end{align*}
Apply the heat differential operator on the first error, we obtain
\begin{align}
& \mathcal{D}_h[e_1] + r_1 = 0, \notag \\
& e_1(x,0) = u_{ic}(x)-e_{1,ic}(x,0), \notag \\
& e_1(x,t) = e_{1,bc}(x,t),\;(x,t)\in \partial \Omega.
\label{eq:heat_derive}
\end{align}
Compared Eq.~\eqref{eq:heat_derive} to Eq.~\eqref{eq:pinn_loss_e1}, the only difference is the boundary condition on $\partial \Omega$. Thus, if an additional loss term regarding boundary condition $\mathcal{L}_{bc}$ is added into Eq.~\eqref{eq:pinn_loss_general} to construct $\hat{e}_1$ (as well as other $\hat{e}_i$), and $u$ is smooth and bounded, then the derivation of theorem \ref{theorem:temporal_error_bound} can be followed.  
\section{System Configurations}\label{appendix:system}
Here, we report system configurations of all the conducted numerical experiments.

\subsection{1D Linear}\label{appendix:1D-L-sys}
We consider an 1D linear system (Ornstein-Uhlenbech process) in \citet[Eq. 4.19]{pavliotis2014stochastic}
$$dx = -\beta x dt + \sqrt{D} dw.$$
From the analytical solution in \citet[Eq. 4.22]{pavliotis2014stochastic}:
$$p(x,t|x_0)=\sqrt{\frac{\beta}{2 \pi D(1-e^{-2 \beta t})}} \exp \Big(-\frac{\beta(x-x_0e^{-\beta t})^2}{2D(1-e^{-2\beta t})} \Big),$$ 
we set the initial distribution as $p_0(x) = p(x,t=1|x_0=1)$, and the system parameters are $\beta=D=0.2$. The computation domain is $t\in[1,3]$ and $x\in[-6,6]$.

\subsection{1D Nonlinear}\label{appendix:1D-NL-sys}
We consider an 1D nonlinear system
$$ dx = (ax^3+bx^2+cx+d)dt + edw.$$
The initial distribution is Gaussian
$$p_0(x)\sim \mathcal{N}(\mu,\sigma^2).$$
The computation domain is $t \in [0,5]$ and $x\in[-6,6]$, and the system parameters are $a=-0.1,b=0.1,c=0.5,d=0.5,e=0.8,\mu=-2$ and $\sigma=0.5$.
Since there is no analytical solution, the "true" PDF $p(x,t)$ is obtained by extensive Monte-Carlo simulation using Euler Scheme with small integration time step $\Delta t = 0.0005$ and $10^8$ samples.

\subsection{2D Inverted Pendulum}\label{appendix:2D-PEND-sys}
We consider 2D nonlinear system of inverted pendulum
\begin{align*}
    d x_1 &= x_2 dt + B_{11} dw, \\
    d x_2 &= -\frac{g}{l}\sin(x_1) dt + B_{22} dw,
\end{align*}
where $x_1$ is the angle, and $x_2$ is the angular rate. The initial distribution is multivariate Gaussian
$$
p_0(x)\sim \mathcal{N}(\mu, \Sigma).
$$
The computation domain is $t \in [0,5]$ and $x_1,x_2 \in [-3 \pi, 3\pi]$, and the system parameters are $g=9.8,l=9.8,B_{11}=B_{22}=0.5,\mu=[0.5\pi,0]^T$ and $diag([0.5,0.5]$.
Similarly, the "true" PDF $p(x,t)$ is obtained by Monte-Carlo simulation using Euler Scheme with integration time step $\Delta t=0.01$ and $2 \times 10^7$ samples.

\subsection{2D Duffing Oscillator}
We consider chaotic dynamics of a 2D duffing oscillator in \citet{anderson2024fisher}
\begin{align*}
    dx_1 &= x_2 dt \\
    dx_2 &= (1.0x_1 -0.2 x_2 -1.0 x_1^3)dt + \frac{1}{\sqrt{20}} dw.
\end{align*}
The initial distribution is multivariate Gaussian
$$
p_0(x)\sim \mathcal{N}(\mu, \Sigma).
$$
The computation domain is $t \in [0,2.5]$ and $x_1,x_2\in[-2,2]$, and the system parameters are $\mu=[1,1]^T, \Sigma=diag([0.05,0.05])$.
Again, extensive Monte-Carlo simulations using Euler Scheme are carried out to estimate the "true" PDF $p(x,t)$, with integration time step $\Delta t=0.005$ and $2 \times 10^7$ samples.

\subsection{High-Dimensional Time-Varying OU}\label{appendx:HighD-sys}
The generic dynamics of high-dimensional time-varying Ornstein-Uhlenbech is
$$
dx=\Big( A_n+\Delta A_n(t)x \Big)dt+B_n dw,
$$
where $x,w \in R^n$. The initial distribution is multivariate Gaussian
$$
p(x,0)\sim \mathcal{N}(\mu_n, \Sigma_n).
$$
Since there is no closed-form solution as discussed in Sec. XXX, we use Euler forward numerical integration with $\Delta t=0.0001$ time step to obtain the "true" PDF. The computation domain is $t\in[0,1]$ and $x\in[-1,1]^n$.
Below, we summarize the system parameters of 3D, 7D, and 10D experiments. For $n=3$, we set
\begin{align*}
    & A_3 = diag([0.3, 0.3, 0.3]) \\
    & \Delta A_3(t) = \sin(t)\begin{bmatrix}
        0 & 0 & -0.1 \\
        0 & 0 & 0 \\
        0 & 0 & 0 
    \end{bmatrix} \\
    & B_3 = diag([0.05,0.05,0.05]) \\
    & \mu_3 = [-0.2, 0.2, 0.0]^T \\
    & \Sigma_3 = diag([0.1, 0.1, 0.1]).
\end{align*}
For $n=7$, we set
\begin{align*}
    & A_7 = \begin{bmatrix}
        0.3 & 0 & 0 & 0 & 0 & 0 & 0 \\
        0 & 0.3 & 0 & 0 & 0 & 0 & 0 \\
        0 & 0 & 0.15 & 0 & 0 & 0 & 0 \\
        0 & 0 & 0 & 0.3 & 0 & 0 & 0 \\
        0 & 0 & 0 & 0 & 0.3 & 0 & 0 \\
        0 & 0 & 0 & 0 & 0 & -0.3 & 0 \\
        -0.01 & 0 & 0 & 0 & 0 & 0 & 0.3 \\
    \end{bmatrix} \\
    & \Delta A_7(t) = \cos(t)\begin{bmatrix}
        0 & 0.1 & 0 & 0 & 0 & 0 & 0 \\
        0 & 0 & 0.1 & 0.2 & 0 & 0 & 0 \\
        0 & 0 & 0 & 0 & 0 & 0 & 0 \\
        0 & 0 & 0 & 0 & 0 & 0 & 0 \\
        0 & 0 & 0 & 0 & 0 & 0 & 0 \\
        0 & 0 & 0 & 0 & 0 & 0 & 0 \\
        0 & -0.1 & 0 & 0 & 0 & 0 & 0 \\
    \end{bmatrix} \\
    & B_7 = diag([0.05,0.05,0.05,0.05,0.05,0.05]) \\
    & \mu_7 = [0,0,0,0,0,0,0]^T \\
    & \Sigma_7 = diag([0.12,0.12,0.12,0.12,0.12,0.12,0.12]).
\end{align*}
For $n=10$, we set
\begin{align*}
    & A_{10} = \begin{bmatrix}
        0.3 & 0 & 0 & 0 & 0 & 0 & 0 & 0 & 0 & 0 \\
        0 & 0.3 & 0 & 0 & 0 & 0.03 & 0 & 0 & 0 & 0 \\
        0 & 0 & -0.3 & 0 & 0 & 0 & 0 & 0 & 0 & 0 \\
        0 & 0 & 0 & 0.3 & 0 & 0 & 0 & 0 & 0 & 0 \\
        0 & 0 & 0 & 0 & 0.06 & 0 & 0 & 0 & 0 & 0 \\
        0 & 0 & 0 & 0 & 0 & 0.3 & 0 & 0 & 0 & 0 \\
        0 & 0 & 0 & 0 & 0 & 0 & 0.3 & 0 & 0 & 0 \\
        0 & 0 & 0 & 0 & 0 & 0 & 0 & 0.21 & 0 & 0 \\
        0 & 0 & 0 & 0 & 0 & 0 & 0 & 0 & 0.3 & 0 \\
        0 & 0 & 0 & 0 & 0 & 0 & 0 & -0.02 & 0 & 0.3 \\
    \end{bmatrix} \\
    & \Delta A_{10}(t) = \sin(t)\begin{bmatrix}
        0 & 0.1 & 0 & 0 & 0 & 0 & 0 & 0 & 0 & 0 \\
        0 & 0 & 0.05 & 0 & 0 & 0 & 0 & 0 & 0 & 0 \\
        0 & 0 & 0 & 0 & 0 & 0 & 0 & 0 & 0 & 0 \\
        0 & 0 & 0 & 0 & 0 & 0 & 0 & 0 & 0 & 0 \\
        0 & 0 & 0 & 0 & 0 & 0 & 0 & 0 & 0 & 0 \\
        0 & 0 & 0 & 0 & 0 & 0 & 0 & 0 & 0 & 0 \\
        0 & 0 & 0 & 0 & 0 & 0 & 0 & 0 & 0 & 0 \\
        0 & 0 & 0 & 0 & 0 & 0 & 0 & 0 & 0 & 0 \\
        0 & 0 & 0 & 0 & 0 & 0 & 0 & 0 & 0 & 0 \\
        0 & -0.1 & 0 & 0 & 0 & 0 & 0 & 0 & 0 & 0 \\
    \end{bmatrix} \\
    & B_{10} = diag([0.05,0.05,0.05,0.05,0.05,0.05,0.05,0.05]) \\
    & \mu_{10} = [0,0,0,0,0,0,0,0,0,0]^T \\
    & \Sigma_{10} = diag([0.12,0.12,0.12,0.12,0.12,0.12,0.12,0.12,0.12,0.12]).
\end{align*}

\subsection{1D Heat}\label{appendix:1D-HEAT-sys}
We consider a one-dimensional heat equation as in Appendix~\ref{proof:1D_Heat}:
\begin{align*}
    u_t-u_{xx}=0,
\end{align*}
subject to initial and Dirichlet boundary conditions
\begin{align*}
& u_0(x)=-\sin(\pi x), \\
& u(\pm1,t)=0,\forall t.
\end{align*}
The computation domain is $t \in [0,1]$ and $x \in [-1,1]$.
For this particular setting, analytical solution exists
$$ u(x,t)=-\sin(\pi x)\exp^{-\pi^2t}.$$
\section{Training Configurations and Additional Results}\label{appendix:train_and_results}
Here we first discuss the general training setting across all experiments. Then we discuss each experiment in details and present additional results and visualizations.
In terms of training scheme, we use Adam optimizer for all experiments. For the systems in Section~\ref{sec:exp_tight_error_bound}, we use the regularization technique discussed in Section~\ref{sec:train_scheme} to train the PDF PINN $\hat{p}$. We also employ adaptive sampling to make training of both $\hat{p}$ and the error PINN $\hat{e}_1$ more efficient (see \citet{lu2021deepxde} for detailed explanation). As for the 1D Linear system in Section~\ref{sec:arb_tight_error_bound_example}, we simply sample random space-time points at every training iteration. For the architecture of the neural network, we use simple fully-connected feed-forward neural networks for both the solution PINN $\hat{p}$ and the error PINN $\hat{e}_1$. Information of the number of hidden layers, neurons, and activation functions for each experiment is provided below.

\subsection{1D Linear Experiment}\label{appendix:1D-L-exp}
\begin{figure}[htbp!]
    \centering
    \includegraphics[width=0.7\linewidth]{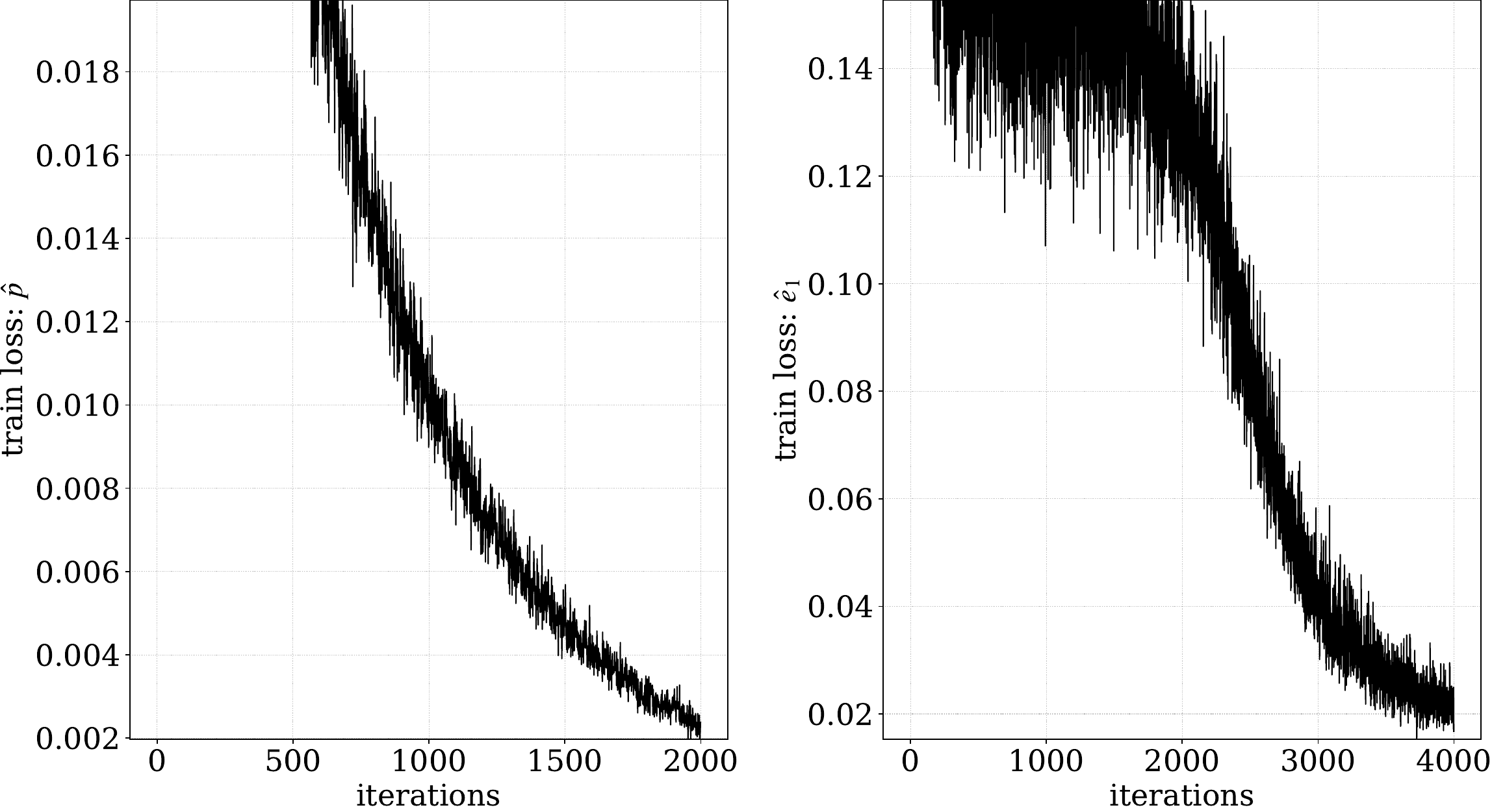}
    \caption{Training losses of $\hat{p}$ and $\hat{e}_1$}
    \label{fig:1dl_trainloss}
\end{figure}

\begin{figure}[htbp!] 
    \centering
    \begin{subfigure}[b]{0.47\textwidth} 
        \includegraphics[width=\textwidth]{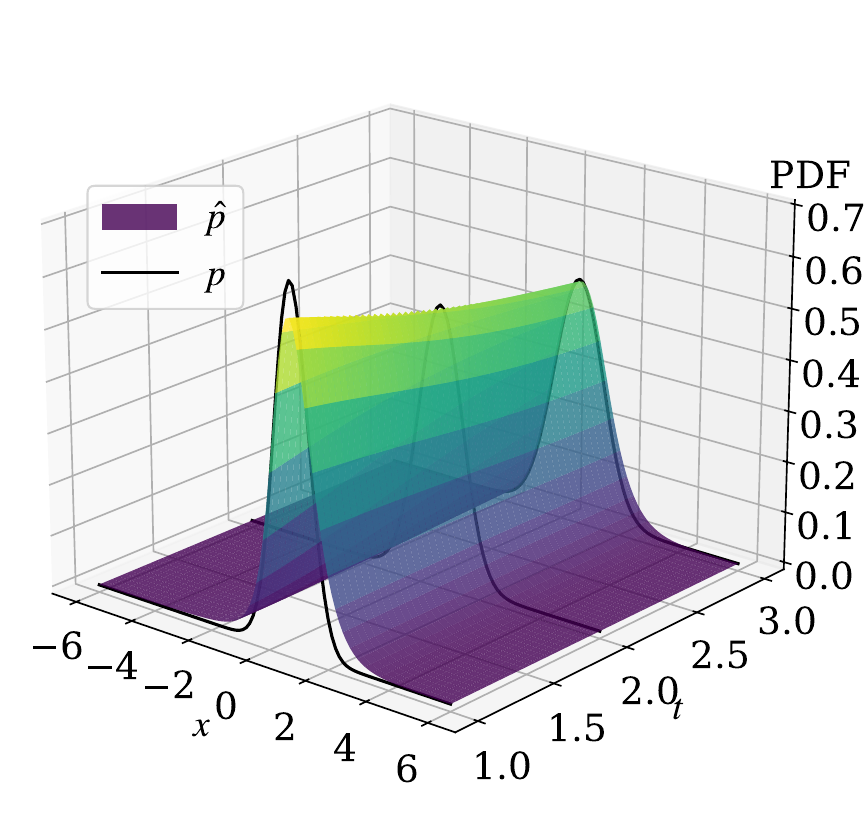}
        \caption{$\hat{p}$ vs $p$}
    \end{subfigure}
    \hfill
    \begin{subfigure}[b]{0.47\textwidth} 
        \includegraphics[width=\textwidth]{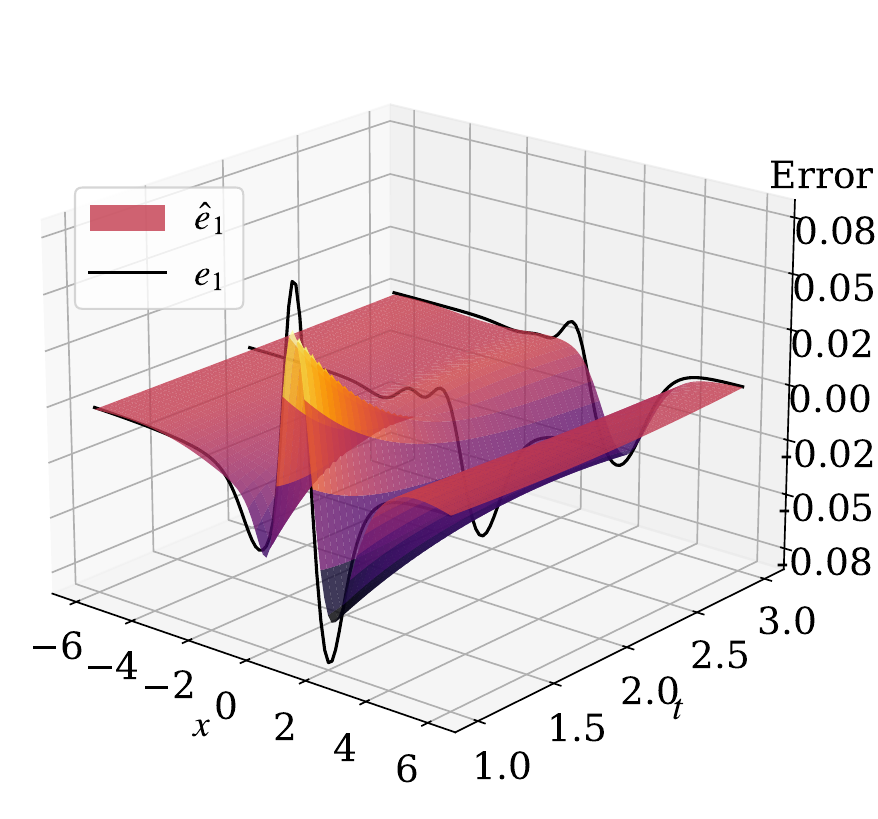}
        \caption{$\hat{e}_1$ vs $e_1$}
    \end{subfigure}
    \caption{Trained PINNs $\hat{p}(x,t)$ and $\hat{e}_1(x,t)$ v.s. true PDF $p(x,t)$ and error $e_1(x,t)$ for all $x$ and $t$.}
    \label{fig:1dl_surfaces}
\end{figure}

\begin{figure}[htbp!]
    \centering
    \includegraphics[width=0.7\linewidth]{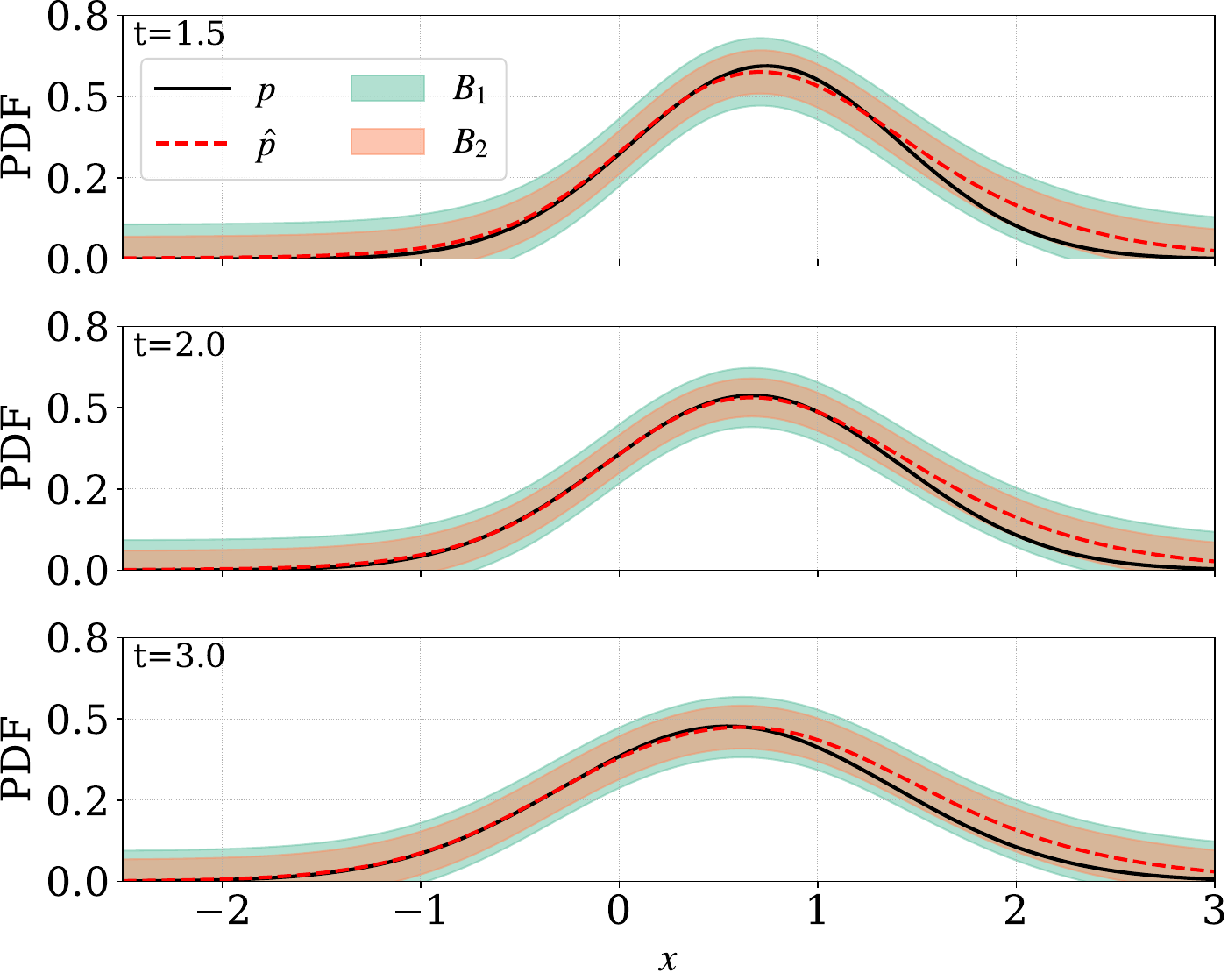}
    \caption{$p$, $\hat{p}$, $B_1$, and $B_2$ at some $t$.}
    \label{fig:1dl_errorboundsresult}
\end{figure}

The neural networks of $\hat{p}$ and $\hat{e}_1$ are summarized in Table~\ref{tab:nnphat_1D-L} and~\ref{tab:nne1hat_1D-L}. For each training iteration, $N_0=N_r=500$ space-time points are uniformly sampled as in Eq.~\ref{eq:pinn_loss_general}, with weights $w_0=1$ and $w_r=|T|=2$. 
The training losses of $\hat{p}$ and $\hat{e}_1$ are illustrated in Fig.~\ref{fig:1dl_trainloss}. The training results of $\hat{p}$ vs $p$ and $\hat{e}_1$ vs $e_t$ are shown in Fig.~\ref{fig:1dl_surfaces}, and the constructed (and synthesized) error bounds at some time instances are visualized in Fig.~\ref{fig:1dl_errorboundsresult}.

\begin{table}[htbp]
    \centering
    \begin{tabular}{lccc}
        \toprule
        \textbf{Layer Connection} & \textbf{Type} & \textbf{\# Neurons (Output)} & \textbf{Activation Function} \\
        \midrule
        Input Layer $\rightarrow$ Hidden Layer 1  & Fully Connected & 32  & Softplus \\
        Hidden Layer 1 $\rightarrow$ Hidden Layer 2  & Fully Connected & 32  & Softplus \\
        Hidden Layer 2 $\rightarrow$ Output Layer  & Fully Connected & 1  & Softplus \\
        \bottomrule
    \end{tabular}
    \caption{Neural Network Architecture and Hyperparameters of $\hat{p}$}
    \label{tab:nnphat_1D-L}
\end{table}

\begin{table}[htbp]
    \centering
    \begin{tabular}{lccc}
        \toprule
        \textbf{Layer Connection} & \textbf{Type} & \textbf{\# Neurons (Output)} & \textbf{Activation Function} \\
        \midrule
        Input Layer $\rightarrow$ Hidden Layer 1  & Fully Connected & 32  & Softplus \\
        Hidden Layer 1 $\rightarrow$ Hidden Layer 2  & Fully Connected & 32  & Softplus \\
        Hidden Layer 2 $\rightarrow$ Output Layer  & Fully Connected & 1  & N/A \\
        \bottomrule
    \end{tabular}
    \caption{Neural network architecture and hyper-parameters of $\hat{e}_1$}
    \label{tab:nne1hat_1D-L}
\end{table}

\clearpage
\subsection{1D Nonlinear Experiment}\label{appendix:1D-NL-exp}
\begin{figure}[htbp!]
    \centering
    \includegraphics[width=0.7\linewidth]{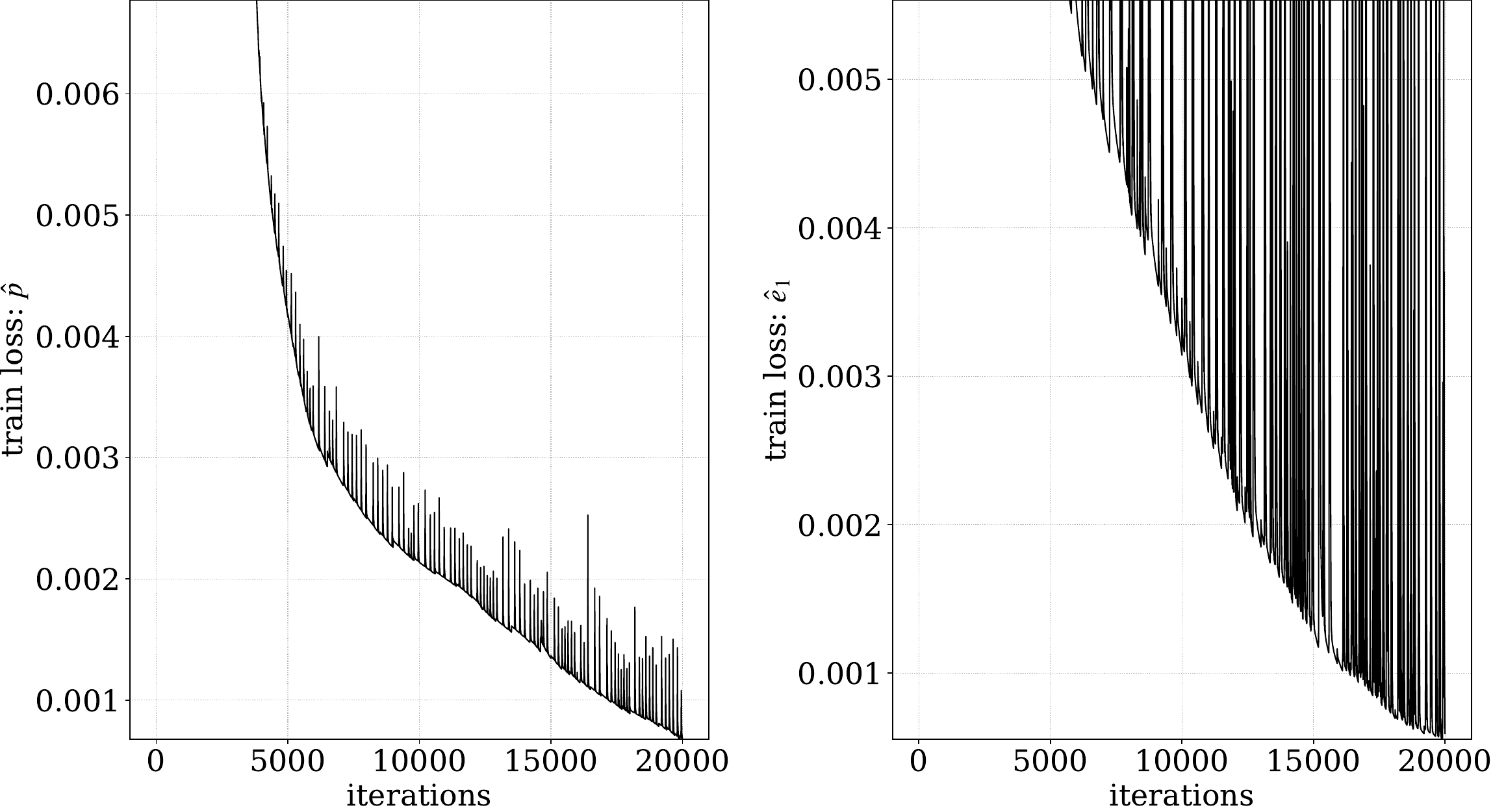}
    \caption{Training losses of $\hat{p}$ and $\hat{e}_1$}
    \label{fig:1dnl_trainloss}
\end{figure}

\begin{figure}[htbp!] 
    \centering
    \begin{subfigure}[b]{0.46\textwidth} 
        \includegraphics[width=\textwidth]{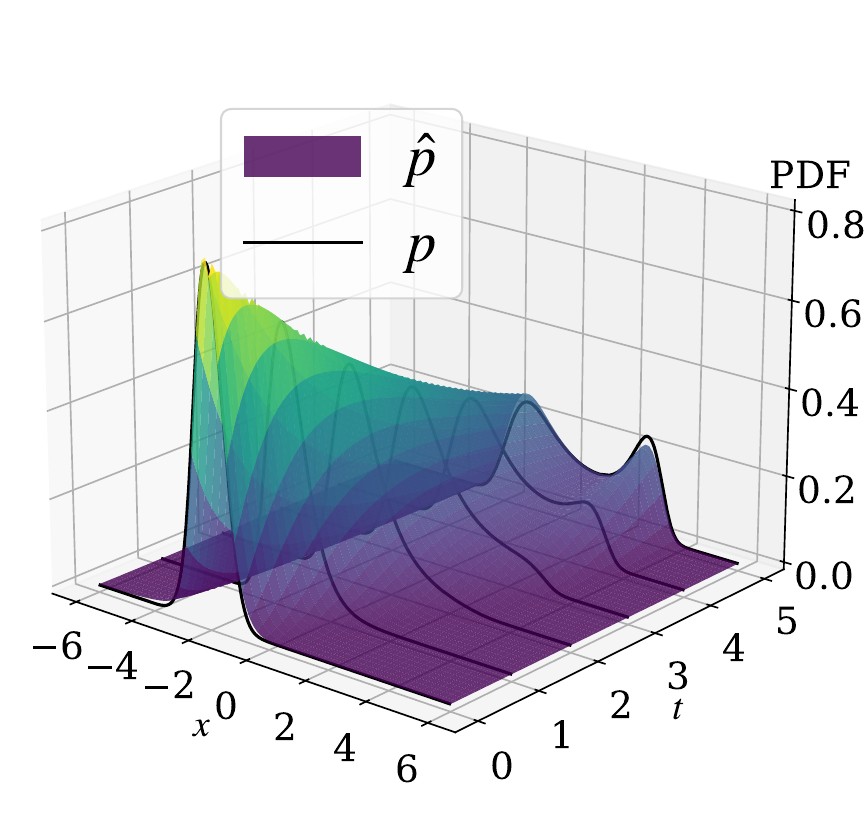}
        \caption{$\hat{p}$ vs $p$}
    \end{subfigure}
    \hfill
    \begin{subfigure}[b]{0.46\textwidth} 
        \includegraphics[width=\textwidth]{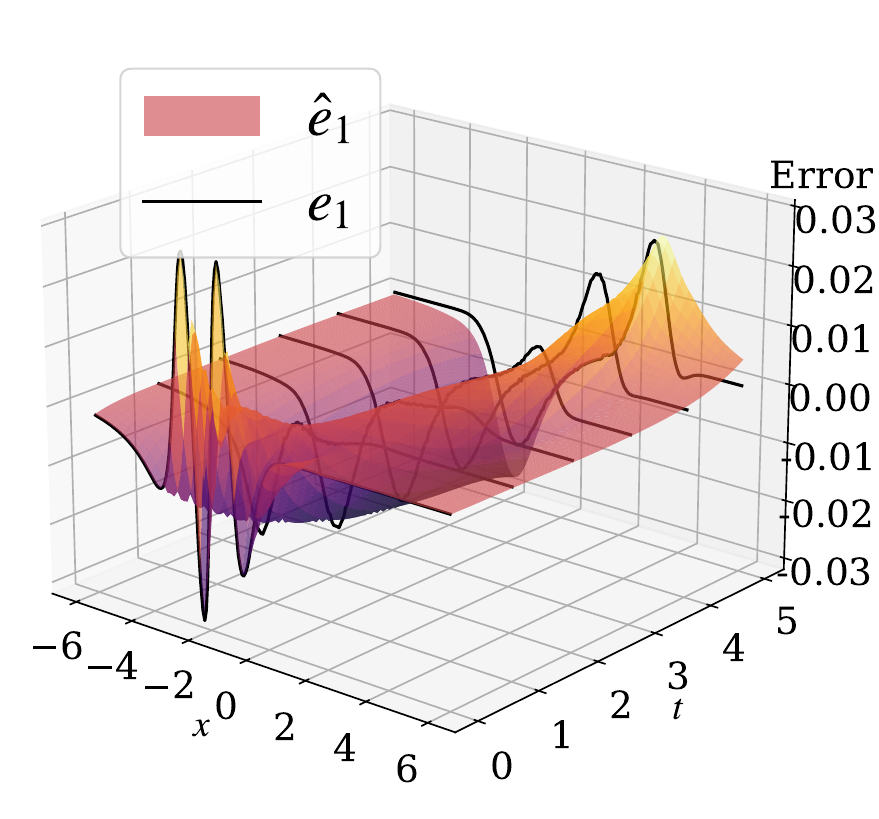}
        \caption{$\hat{e}_1$ vs $e_1$}
    \end{subfigure}
    \caption{Trained PINNs $\hat{p}(x,t)$ and $\hat{e}_1(x,t)$ v.s. true PDF $p(x,t)$ and error $e_1(x,t)$ for all $x$ and $t$.}
    \label{fig:1dnl_surfaces}
\end{figure}

The neural networks of $\hat{p}$ and $\hat{e}_1$ are summarized in Table~\ref{tab:nnphat_1D-NL} and~\ref{tab:nne1hat_1D-NL}. 
The training starts from $N_0=N_r=1000$ uniformly distributed samples for both $\hat{p}$ and $\hat{e}_1$.
We regularize the training of $\hat{p}$ by setting the weights $w_0=1$ and $w_r=w_{\nabla}=|T|=5$. For $\hat{e}_1$, the weights are $w_0=1,w_r=|T|$, and $w_{\nabla}=0$.
The training losses of $\hat{p}$ and $\hat{e}_1$ are illustrated in Fig.~\ref{fig:1dnl_trainloss}. Note that the periodic spikes are not due to unstable training. Instead they are due to the adaptive sampling scheme that periodically adds space-time points at which the residual values are large.
The training results of $\hat{p}$ vs $p$ and $\hat{e}_1$ vs $e_t$ are shown in Fig.~\ref{fig:1dnl_surfaces}; the constructed error bounds at some time instances can be seen in Fig.~\ref{fig:representative_results}a.

\begin{table}[htbp]
    \centering
    \begin{tabular}{lccc}
        \toprule
        \textbf{Layer Connection} & \textbf{Type} & \textbf{\# Neurons (Output)} & \textbf{Activation Function} \\
        \midrule
        Input Layer $\rightarrow$ Hidden Layer 1  & Fully Connected & 50  & Softplus \\
        Hidden Layer 1 $\rightarrow$ Hidden Layer 2  & Fully Connected & 50  & Softplus \\
        Hidden Layer 2 $\rightarrow$ Hidden Layer 3  & Fully Connected & 50  & Softplus \\
        Hidden Layer 3 $\rightarrow$ Output Layer  & Fully Connected & 1  & Softplus \\
        \bottomrule
    \end{tabular}
    \caption{Neural Network Architecture and Hyperparameters of $\hat{p}$}
    \label{tab:nnphat_1D-NL}
\end{table}

\begin{table}[htbp]
    \centering
    \begin{tabular}{lccc}
        \toprule
        \textbf{Layer Connection} & \textbf{Type} & \textbf{\# Neurons (Output)} & \textbf{Activation Function} \\
        \midrule
        Input Layer $\rightarrow$ Hidden Layer 1  & Fully Connected & 50  & GeLU \\
        Hidden Layer 1 $\rightarrow$ Hidden Layer 2  & Fully Connected & 50  & GeLU \\
        Hidden Layer 2 $\rightarrow$ Hidden Layer 3  & Fully Connected & 50  & GeLU \\
        Hidden Layer 3 $\rightarrow$ Hidden Layer 4  & Fully Connected & 50  & GeLU \\
        Hidden Layer 4 $\rightarrow$ Hidden Layer 5  & Fully Connected & 50  & GeLU \\
        Hidden Layer 5 $\rightarrow$ Hidden Layer 6  & Fully Connected & 50  & GeLU \\
        Hidden Layer 6 $\rightarrow$ Output Layer  & Fully Connected & 1  & N/A \\
        \bottomrule
    \end{tabular}
    \caption{Neural network architecture and hyper-parameters of $\hat{e}_1$}
    \label{tab:nne1hat_1D-NL}
\end{table}

\clearpage
\subsection{2D Inverted Pendulum Experiment}\label{appendix:2D-PEND-exp}
\begin{figure}[htbp!]
    \centering
    \includegraphics[width=0.7\linewidth]{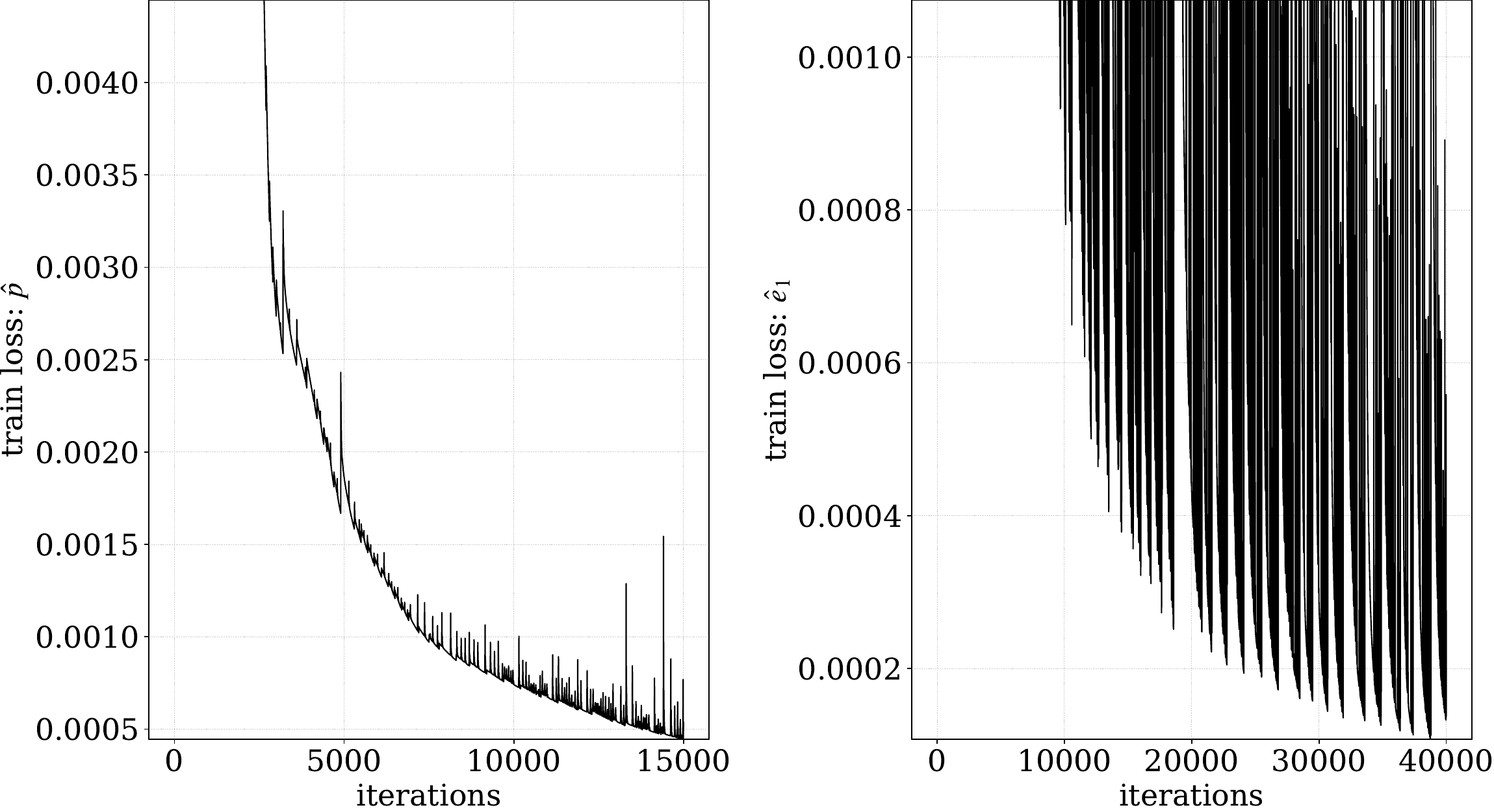}
    \caption{Training losses of $\hat{p}$ and $\hat{e}_1$}
    \label{fig:2dpend_trainloss}
\end{figure}

\begin{figure}[htbp!] 
    \centering
    \begin{subfigure}[b]{0.98\textwidth} 
        \includegraphics[width=\textwidth]{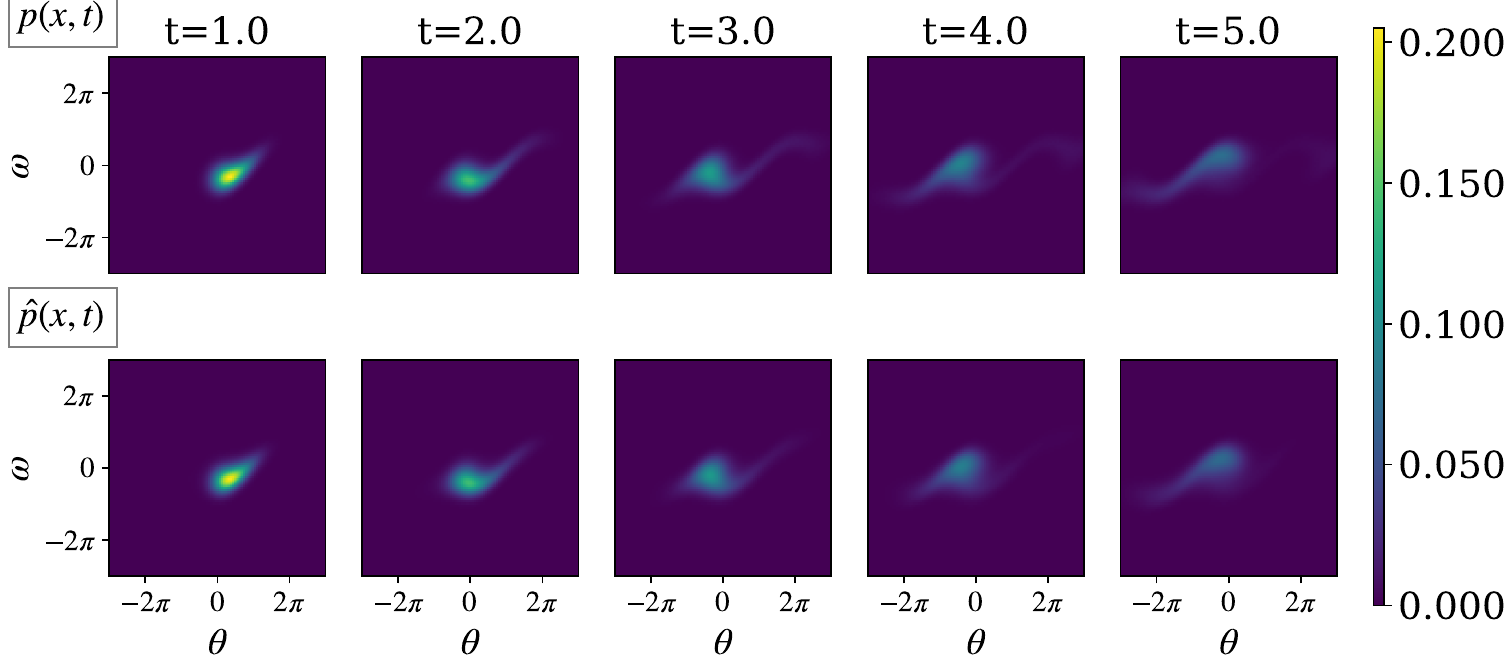}
        \caption{$\hat{p}$ vs $p$}
    \end{subfigure}
    \vfill
    \begin{subfigure}[b]{0.98\textwidth} 
        \includegraphics[width=\textwidth]{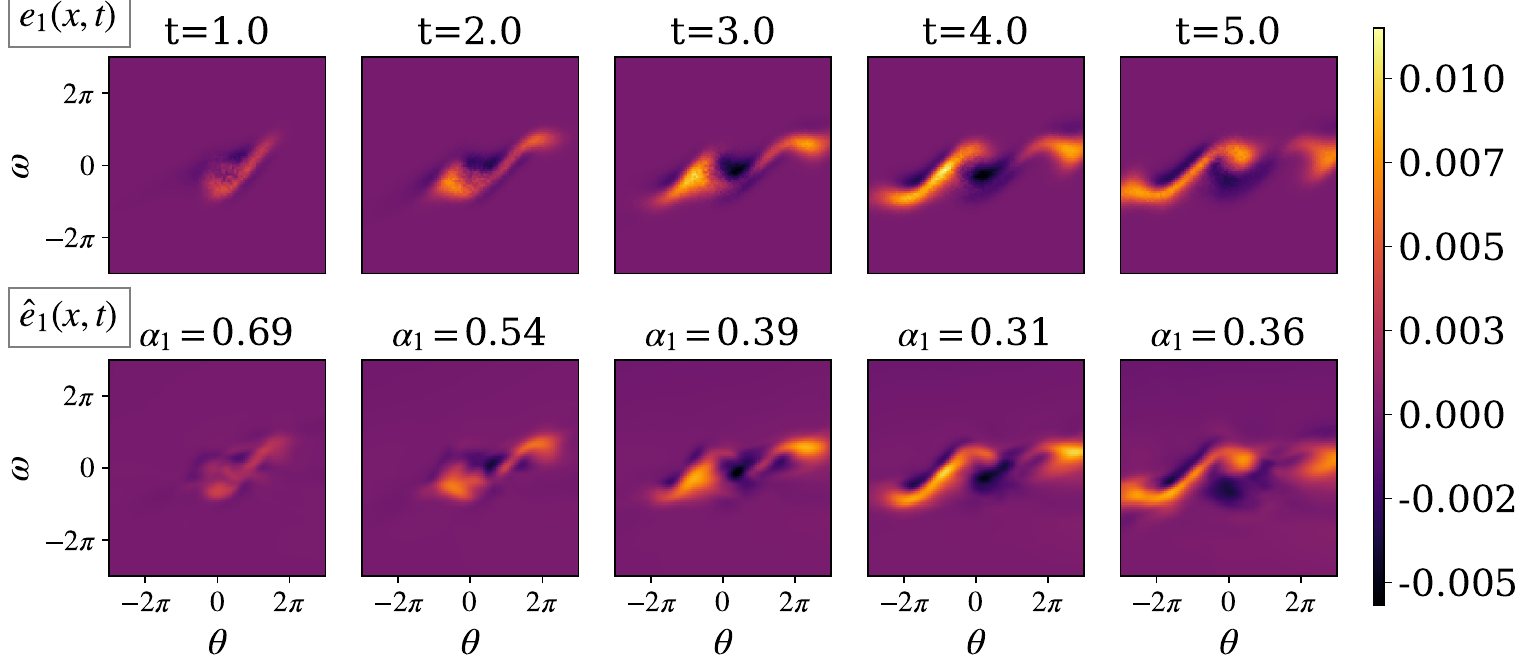}
        \caption{$\hat{e}_1$ vs $e_1$}
    \end{subfigure}
    \caption{Trained PINNs $\hat{p}(\theta,\omega,t)$ and $\hat{e}_1(\theta,\omega,t)$ v.s. true PDF $p(\theta,\omega,t)$ and error $e_1(\theta,\omega,t)$ for all $\theta,\omega$ at some $t$.}
    \label{fig:2dpend_results}
\end{figure}

\begin{figure}[htbp!]
    \centering
    \includegraphics[width=1.0\linewidth]{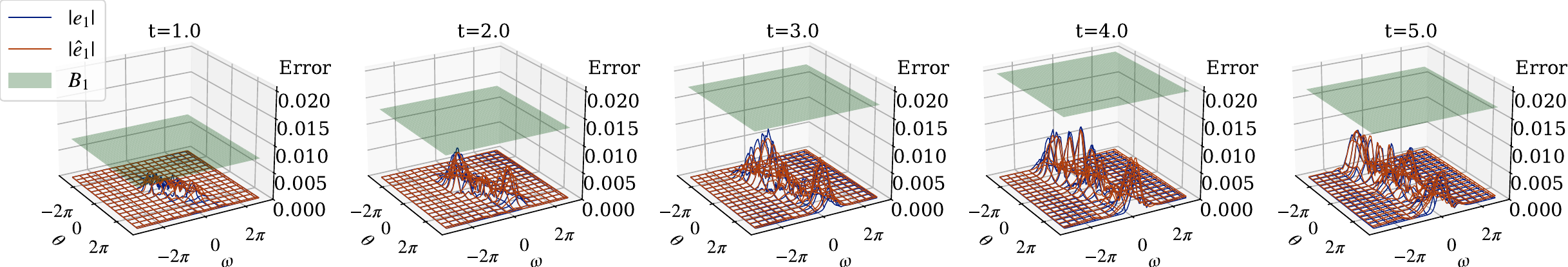}
    \caption{$|e_1|$, $|\hat{e}_1|$, and $B_1$at some $t$.}
    \label{fig:2dpend_errorbound}
\end{figure}

The neural networks of $\hat{p}$ and $\hat{e}_1$ are summarized in Table~\ref{tab:nnphat_2D-PEND} and~\ref{tab:nne1hat_2D-PEND}. 
As in Appendix~\ref{appendix:1D-NL-exp}, the training starts from $N_0=N_r=1000$ uniformly distributed samples for both $\hat{p}$ and $\hat{e}_1$.
We regularize the training of $\hat{p}$ by setting the weights $w_0=1$ and $w_r=w_{\nabla}=|T|=5$. For $\hat{e}_1$, the weights are $w_0=1,w_r=|T|$, and $w_{\nabla}=0$.
The training losses of $\hat{p}$ and $\hat{e}_1$ are illustrated in Fig.~\ref{fig:2dpend_trainloss}. 
Again, the periodic spikes in training loss are due to the adaptive sampling scheme that periodically adds space-time points, which becomes more effective as the system dimension grows.
The training results of $\hat{p}$ vs $p$ and $\hat{e}_1$ vs $e_t$ are shown in Fig.~\ref{fig:2dpend_results}. 
The constructed tight error bounds at some time instances are visualized in Fig.~\ref{fig:2dpend_errorbound}.

\begin{table}[htbp]
    \centering
    \begin{tabular}{lccc}
        \toprule
        \textbf{Layer Connection} & \textbf{Type} & \textbf{\# Neurons (Output)} & \textbf{Activation Function} \\
        \midrule
        Input Layer $\rightarrow$ Hidden Layer 1  & Fully Connected & 50  & Softplus \\
        Hidden Layer 1 $\rightarrow$ Hidden Layer 2  & Fully Connected & 50  & Softplus \\
        Hidden Layer 2 $\rightarrow$ Hidden Layer 3  & Fully Connected & 50  & Softplus \\
        Hidden Layer 3 $\rightarrow$ Output Layer  & Fully Connected & 1  & Softplus \\
        \bottomrule
    \end{tabular}
    \caption{Neural Network Architecture and Hyperparameters of $\hat{p}$}
    \label{tab:nnphat_2D-PEND}
\end{table}

\begin{table}[htbp]
    \centering
    \begin{tabular}{lccc}
        \toprule
        \textbf{Layer Connection} & \textbf{Type} & \textbf{\# Neurons (Output)} & \textbf{Activation Function} \\
        \midrule
        Input Layer $\rightarrow$ Hidden Layer 1  & Fully Connected & 50  & Softplus \\
        Hidden Layer 1 $\rightarrow$ Hidden Layer 2  & Fully Connected & 50  & Softplus \\
        Hidden Layer 2 $\rightarrow$ Hidden Layer 3  & Fully Connected & 50  & Softplus \\
        Hidden Layer 3 $\rightarrow$ Hidden Layer 4  & Fully Connected & 50  & Softplus \\
        Hidden Layer 4 $\rightarrow$ Hidden Layer 5  & Fully Connected & 50  & Softplus \\
        Hidden Layer 5 $\rightarrow$ Hidden Layer 6  & Fully Connected & 50  & Softplus \\
        Hidden Layer 6 $\rightarrow$ Output Layer  & Fully Connected & 1  & N/A \\
        \bottomrule
    \end{tabular}
    \caption{Neural network architecture and hyper-parameters of $\hat{e}_1$}
    \label{tab:nne1hat_2D-PEND}
\end{table}

\clearpage
\subsection{2D Duffing Oscillator Experiment}\label{appendix:2D-DUFF-exp}
\begin{figure}[htbp!]
    \centering
    \includegraphics[width=0.7\linewidth]{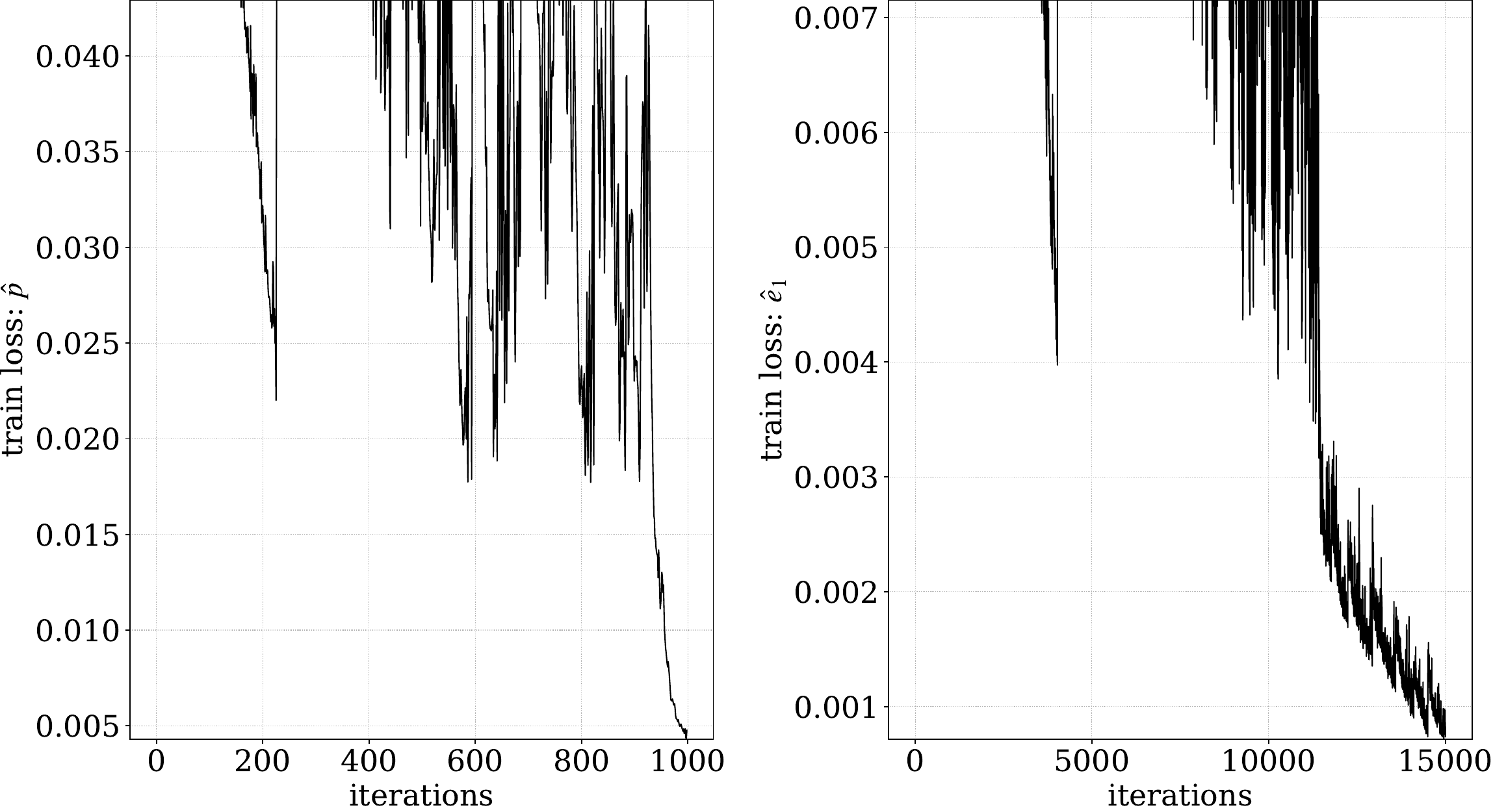}
    \caption{Training losses of $\hat{p}$ and $\hat{e}_1$}
    \label{fig:2dduff_trainloss}
\end{figure}

\begin{figure}[htbp!] 
    \centering
    \begin{subfigure}[b]{0.98\textwidth} 
        \includegraphics[width=\textwidth]{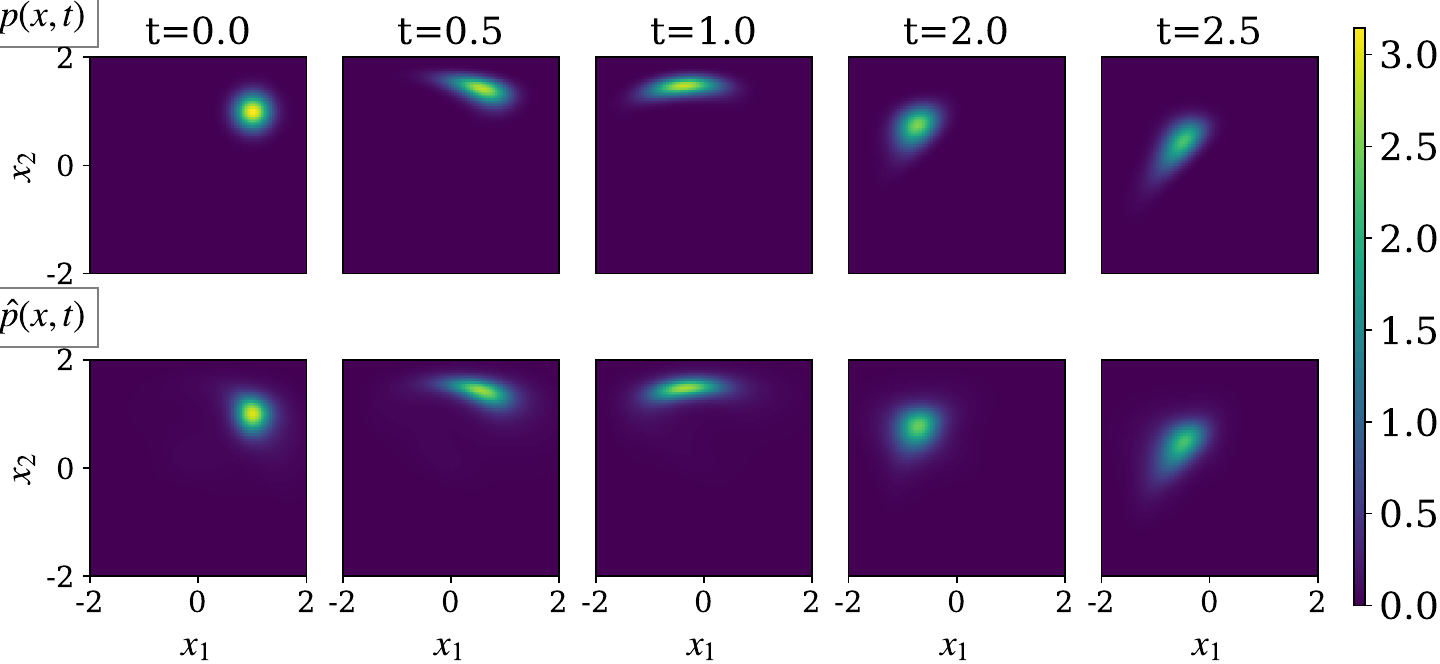}
        \caption{$\hat{p}$ vs $p$}
    \end{subfigure}
    \vfill
    \begin{subfigure}[b]{0.98\textwidth} 
        \includegraphics[width=\textwidth]{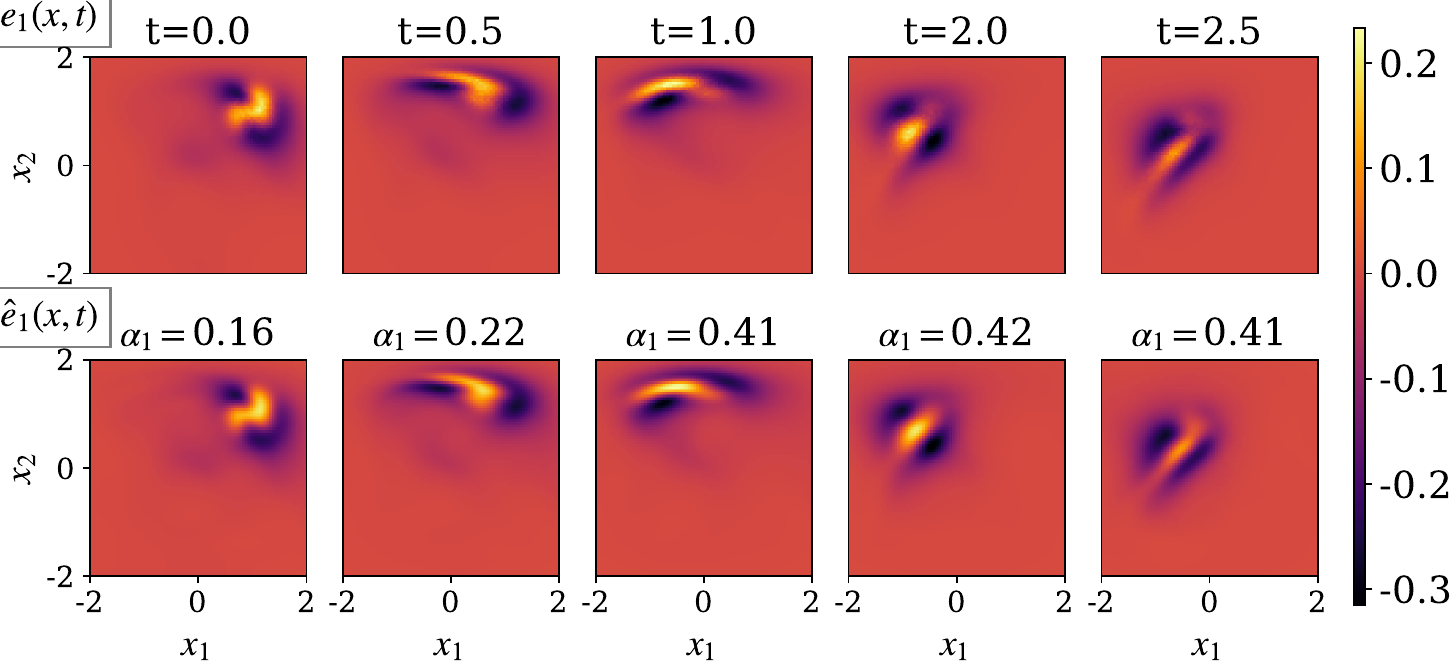}
        \caption{$\hat{e}_1$ vs $e_1$}
    \end{subfigure}
    \caption{Trained PINNs $\hat{p}(x_1,x_2,t)$ and $\hat{e}_1(x_1,x_2,t)$ v.s. true PDF $p(x_1,x_2,t)$ and error $e_1(x_1,x_2,t)$ for all $x_1,x_2$ at some $t$.}
    \label{fig:2dduff_results}
\end{figure}

\begin{figure}[htbp!]
    \centering
    \includegraphics[width=1.0\linewidth]{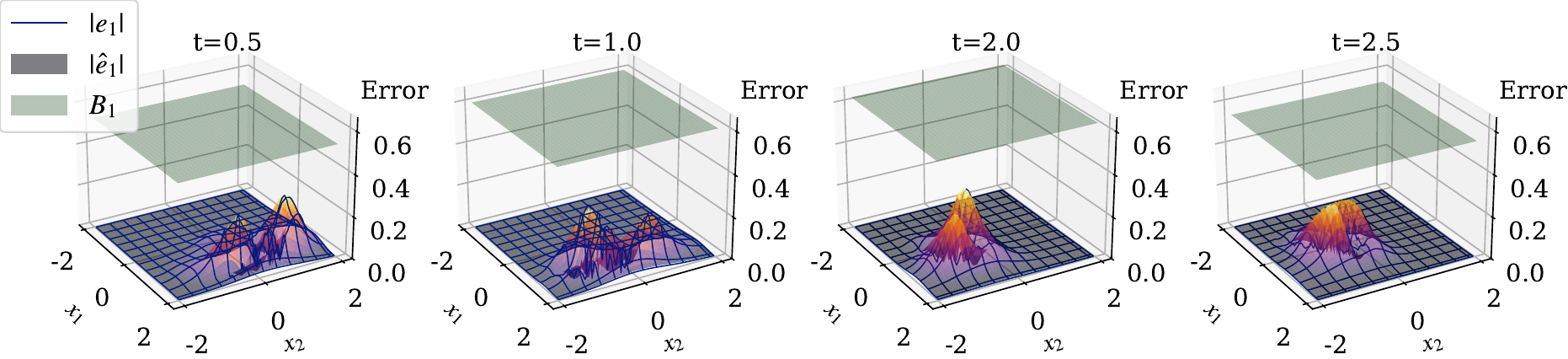}
    \caption{$|e_1|$, $|\hat{e}_1|$, and $B_1$at some $t$.}
    \label{fig:2dduff_errorbound}
\end{figure}

The neural networks of $\hat{p}$ and $\hat{e}_1$ are summarized in Table~\ref{tab:nnphat_2D-DUFF} and~\ref{tab:nne1hat_2D-DUFF}. 
The training starts from $N_0=N_r=1000$ samples for both $\hat{p}$ and $\hat{e}_1$. Half of these samples are drawn uniformly, and the other half follow the normal distribution specified by the initial condition.
We regularize the training of $\hat{p}$ by setting the weights $w_0=1$ and $w_r=w_{\nabla}=|T|=5$. For $\hat{e}_1$, the weights are $w_0=1,w_r=|T|$, and $w_{\nabla}=0$.
The training losses of $\hat{p}$ and $\hat{e}_1$ are illustrated in Fig.~\ref{fig:2dduff_trainloss}. 
Again, the periodic spikes in training loss are due to the adaptive sampling scheme that periodically adds space-time points, which becomes more effective as the system dimension grows.
The training results of $\hat{p}$ vs $p$ and $\hat{e}_1$ vs $e_t$ are shown in Fig.~\ref{fig:2dduff_results}. 
The constructed tight error bounds at some time instances are visualized in Fig.~\ref{fig:2dduff_errorbound}.

\begin{table}[htbp]
    \centering
    \begin{tabular}{lccc}
        \toprule
        \textbf{Layer Connection} & \textbf{Type} & \textbf{\# Neurons (Output)} & \textbf{Activation Function} \\
        \midrule
        Input Layer $\rightarrow$ Hidden Layer 1  & Fully Connected & 60  & GeLU \\
        Hidden Layer 1 $\rightarrow$ Hidden Layer 2  & Fully Connected & 60  & GeLU \\
        Hidden Layer 2 $\rightarrow$ Hidden Layer 3  & Fully Connected & 60  & GeLU \\
        Hidden Layer 3 $\rightarrow$ Hidden Layer 4  & Fully Connected & 60  & GeLU \\
        Hidden Layer 4 $\rightarrow$ Hidden Layer 5  & Fully Connected & 60  & GeLU \\
        Hidden Layer 5 $\rightarrow$ Output Layer  & Fully Connected & 1  & Softplus \\
        \bottomrule
    \end{tabular}
    \caption{Neural Network Architecture and Hyperparameters of $\hat{p}$}
    \label{tab:nnphat_2D-DUFF}
\end{table}

\begin{table}[htbp]
    \centering
    \begin{tabular}{lccc}
        \toprule
        \textbf{Layer Connection} & \textbf{Type} & \textbf{\# Neurons (Output)} & \textbf{Activation Function} \\
        \midrule
        Input Layer $\rightarrow$ Hidden Layer 1  & Fully Connected & 100  & GeLu \\
        Hidden Layer 1 $\rightarrow$ Hidden Layer 2  & Fully Connected & 100  & GeLu \\
        Hidden Layer 2 $\rightarrow$ Hidden Layer 3  & Fully Connected & 100  & GeLu \\
        Hidden Layer 3 $\rightarrow$ Hidden Layer 4  & Fully Connected & 100  & GeLu \\
        Hidden Layer 4 $\rightarrow$ Hidden Layer 5  & Fully Connected & 100  & GeLu \\
        Hidden Layer 5 $\rightarrow$ Output Layer  & Fully Connected & 1  & N/A \\
        \bottomrule
    \end{tabular}
    \caption{Neural network architecture and hyper-parameters of $\hat{e}_1$}
    \label{tab:nne1hat_2D-DUFF}
\end{table}

\clearpage
\subsection{3D Time-Varying OU Experiment}\label{appendix:3D-TVOU-exp}
\begin{figure}[htbp!]
    \centering
    \includegraphics[width=0.7\linewidth]{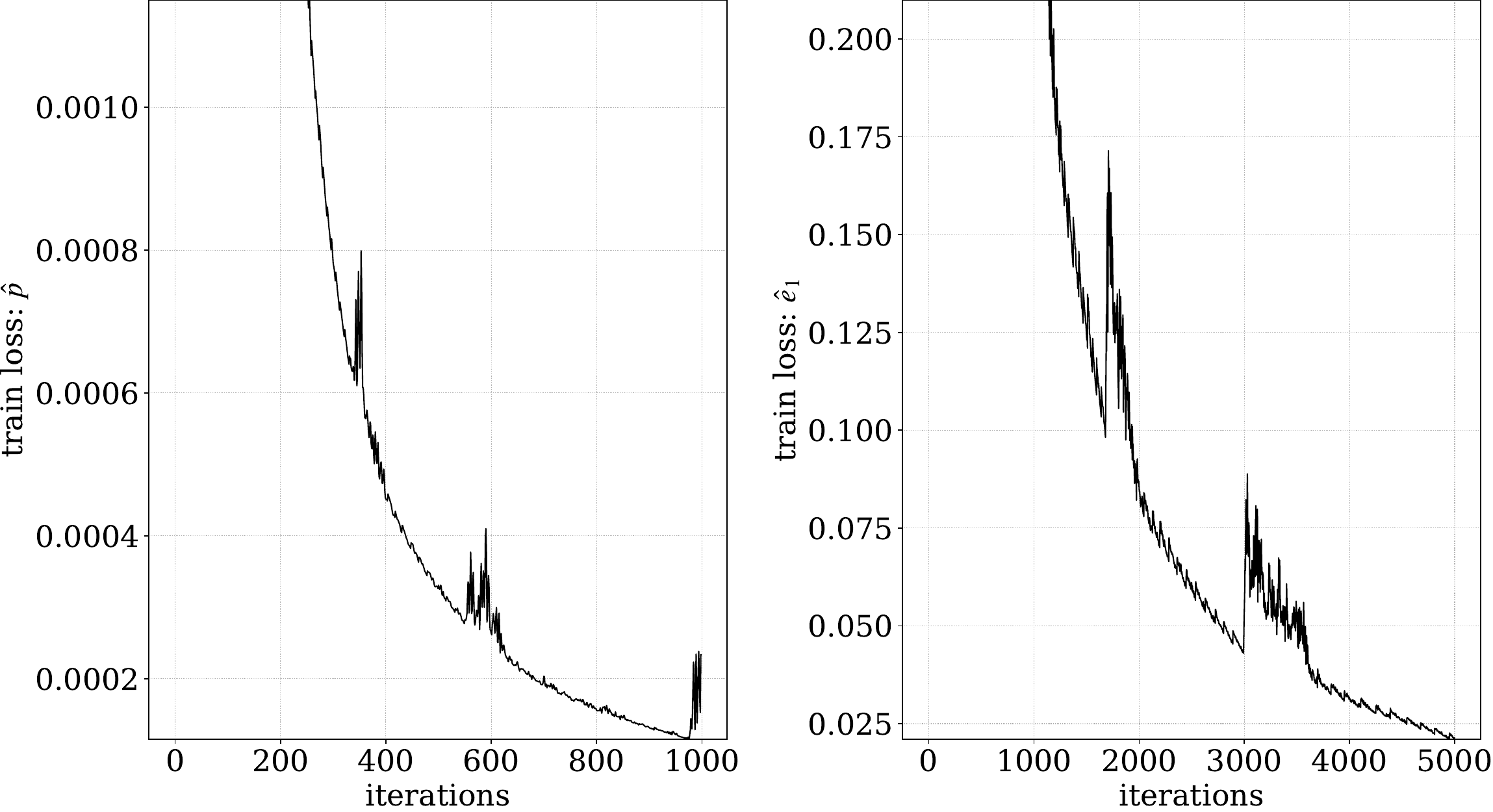}
    \caption{Training losses of $\hat{p}$ and $\hat{e}_1$}
    \label{fig:3dtvou_trainloss}
\end{figure}

\begin{figure}[htbp!] 
    \centering
    \begin{subfigure}[b]{0.40\textwidth} %
        \includegraphics[width=\textwidth]{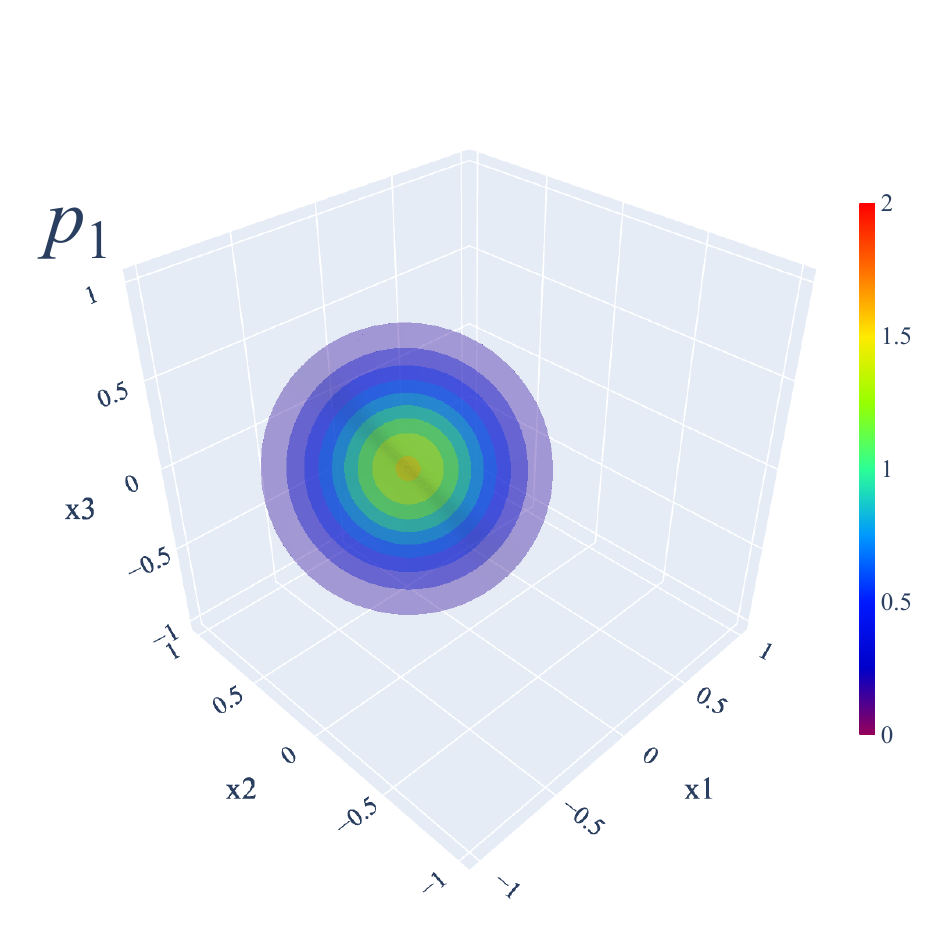}
        \caption{$p(x,t=0.2)$}
    \end{subfigure}
    \begin{subfigure}[b]{0.40\textwidth} %
        \includegraphics[width=\textwidth]{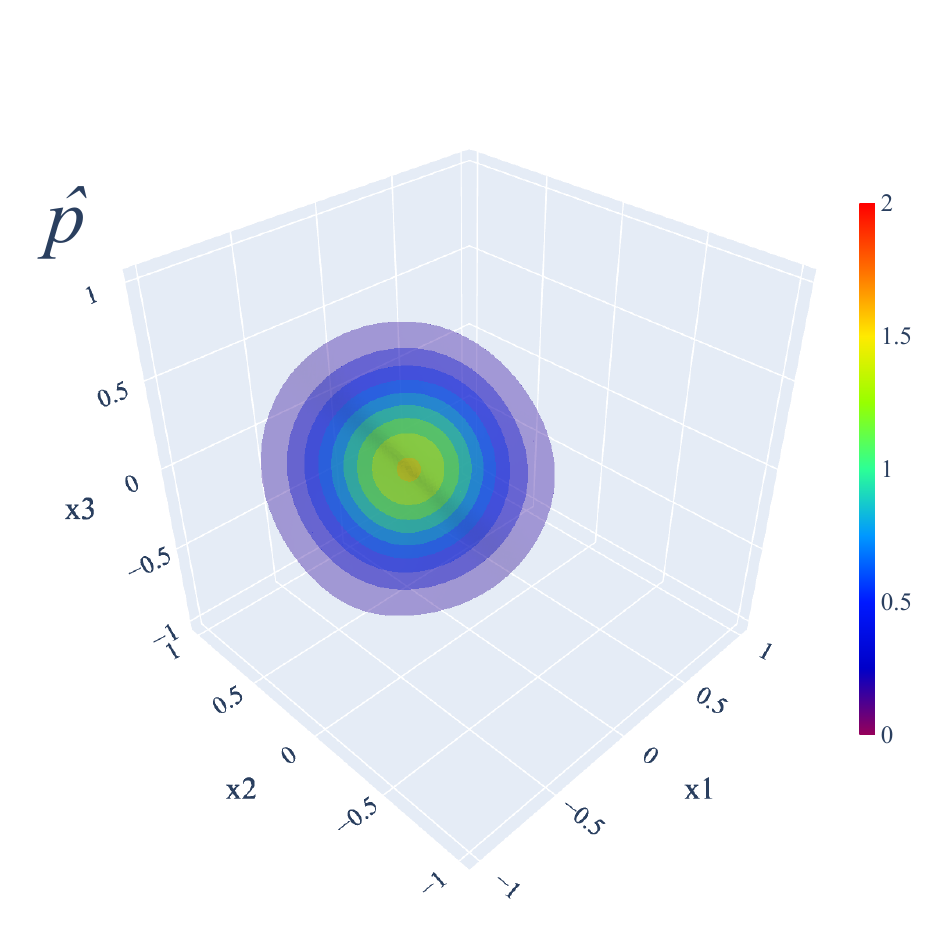}
        \caption{$\hat{p}(x,t=0.2)$}
    \end{subfigure}
    \begin{subfigure}[b]{0.40\textwidth} %
        \includegraphics[width=\textwidth]{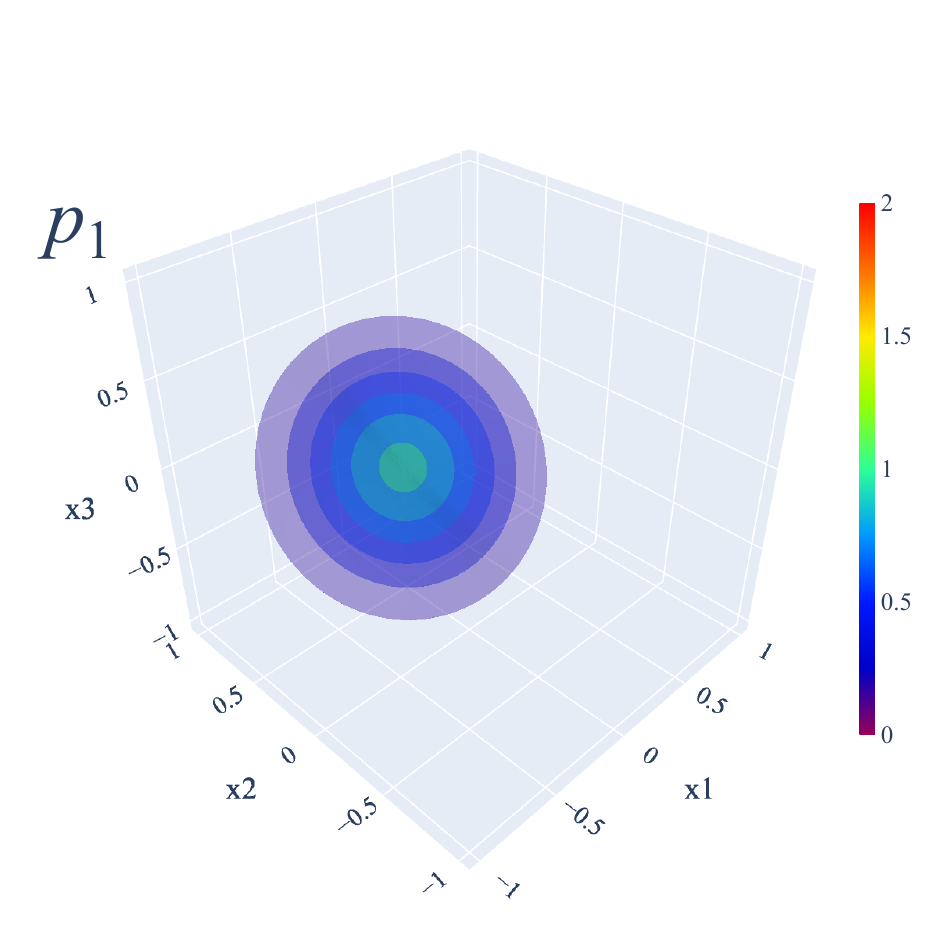}
        \caption{$p(x,t=0.6)$}
    \end{subfigure}
    \begin{subfigure}[b]{0.40\textwidth} %
        \includegraphics[width=\textwidth]{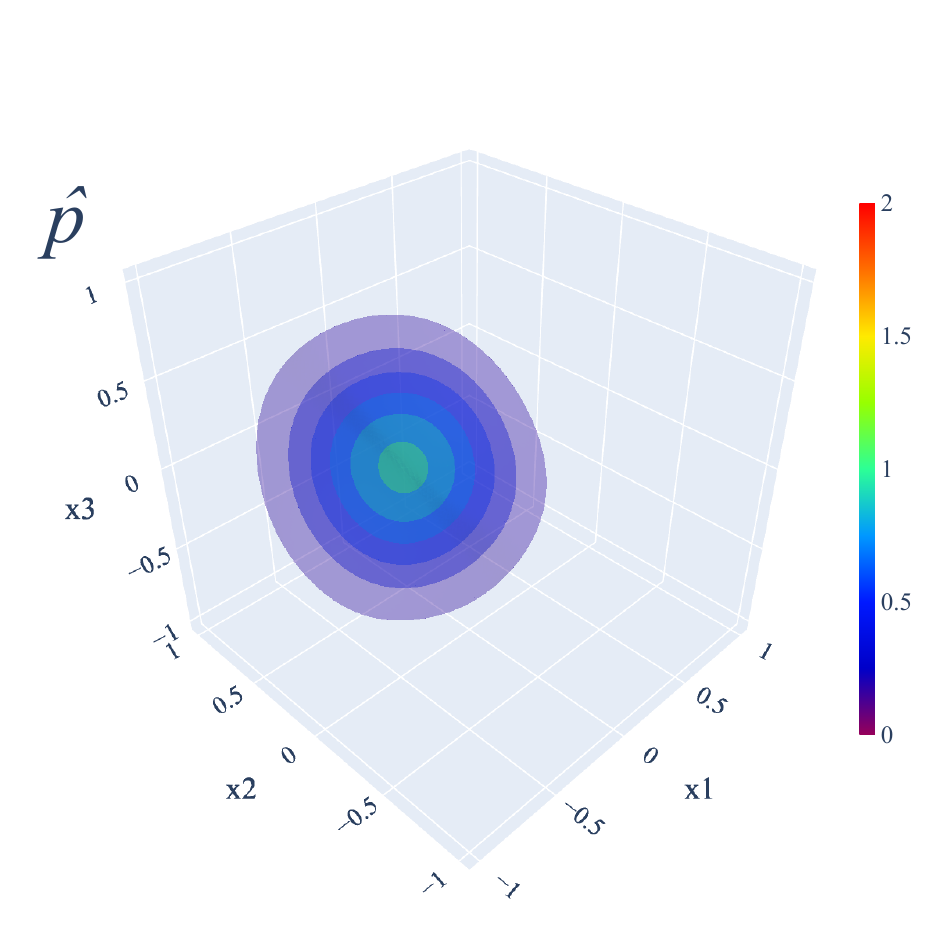}
        \caption{$\hat{p}(x,t=0.6)$}
    \end{subfigure}
    \begin{subfigure}[b]{0.40\textwidth} %
        \includegraphics[width=\textwidth]{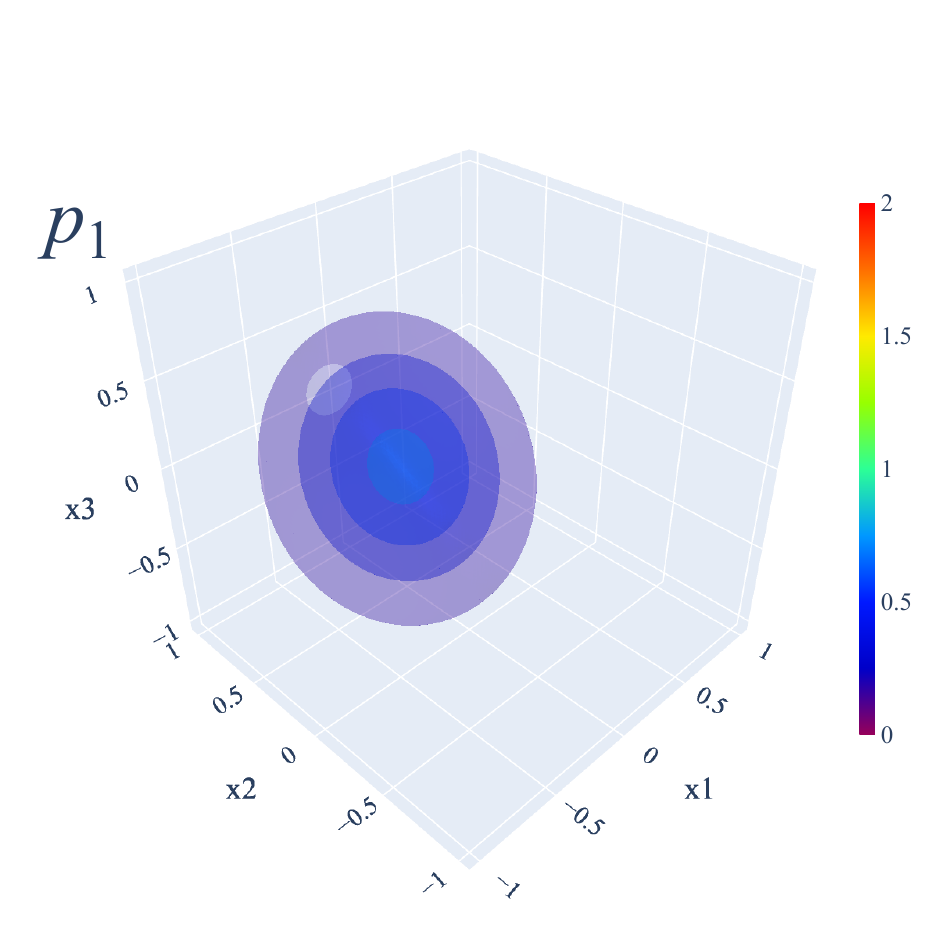}
        \caption{$p(x,t=1.0)$}
    \end{subfigure}
    \begin{subfigure}[b]{0.40\textwidth} %
        \includegraphics[width=\textwidth]{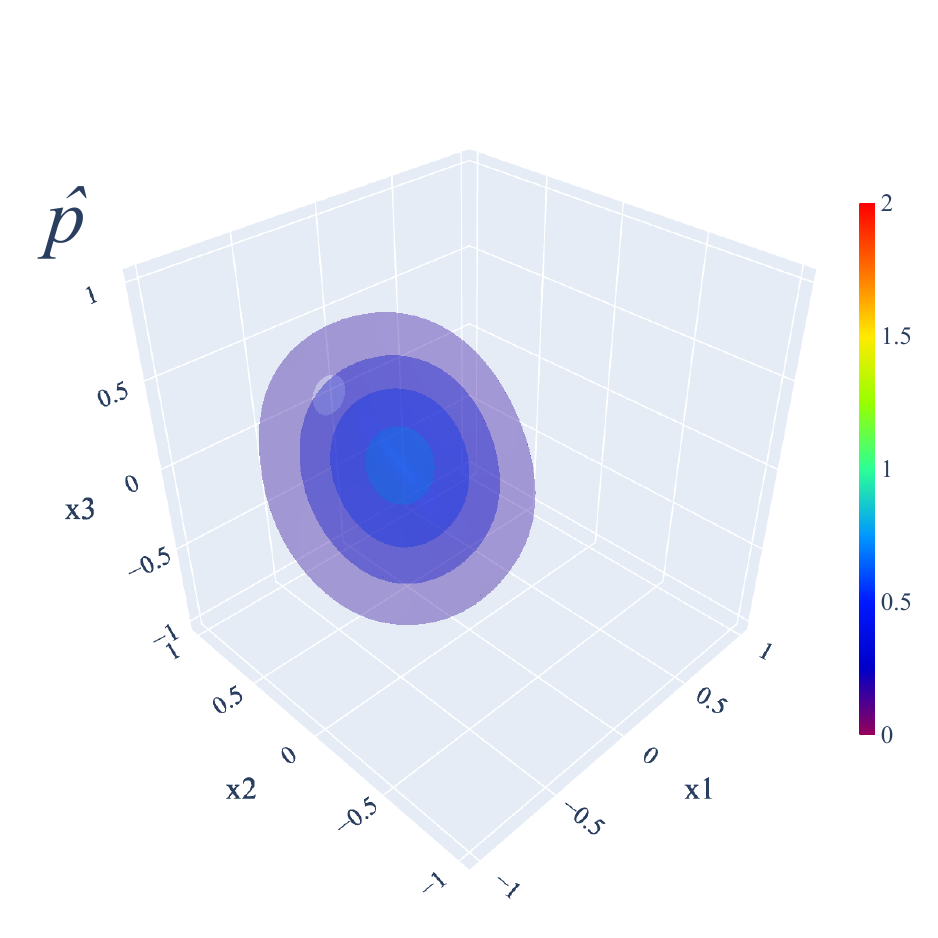}
        \caption{$\hat{p}(x,t=1.0)$}
    \end{subfigure}
    \caption{Trained PINNs $\hat{p}(x,t)$ v.s. true PDF $p(x,t)$ for all $x$ at some $t$.}
    \label{fig:3dtv_phatresult}
\end{figure}

\begin{figure}[htbp!]
    \centering
    \begin{subfigure}[b]{0.40\textwidth} %
        \includegraphics[width=\textwidth]{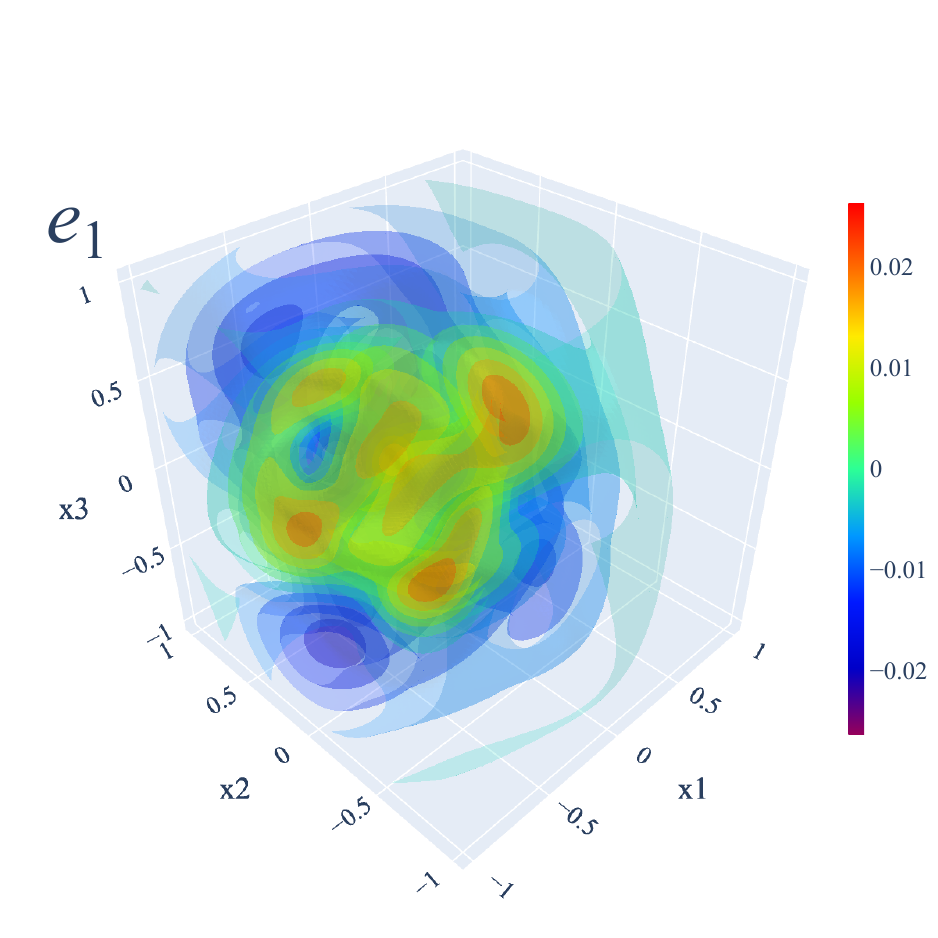}
        \caption{$e_1(x,t=0.2)$}
    \end{subfigure}
    \begin{subfigure}[b]{0.40\textwidth} %
        \includegraphics[width=\textwidth]{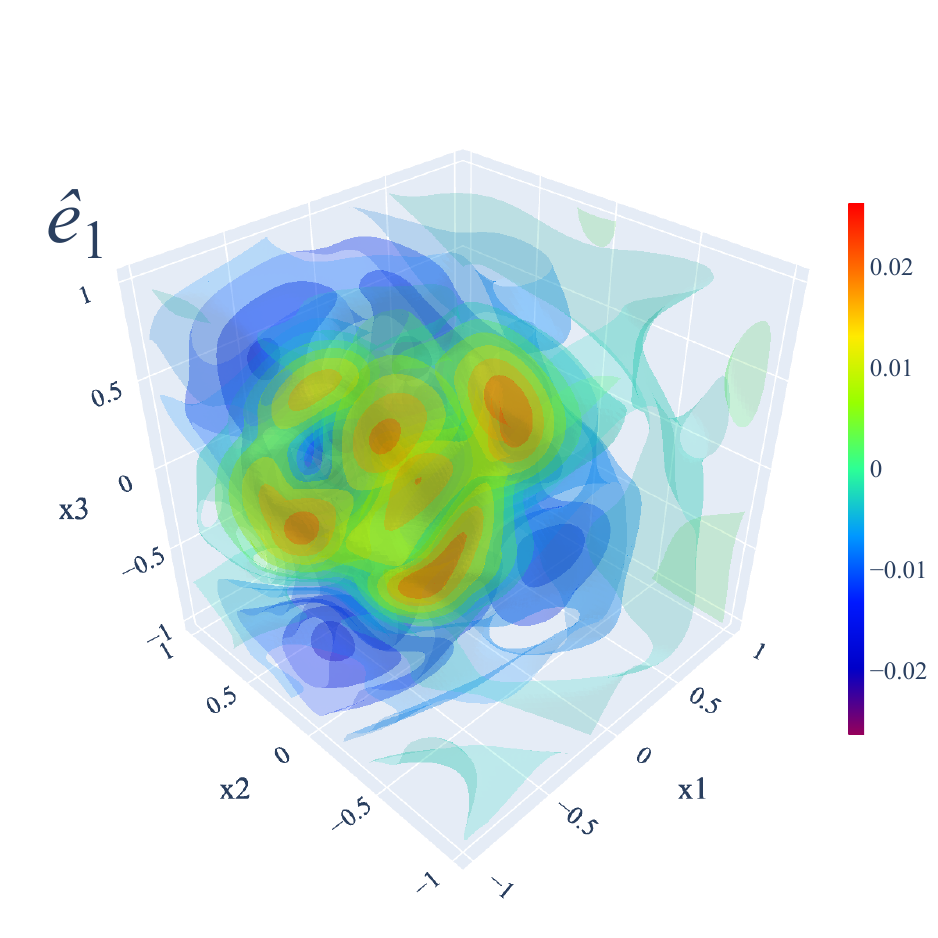}
        \caption{$\hat{e}_1(x,t=0.2)$}
    \end{subfigure}
    \begin{subfigure}[b]{0.40\textwidth} %
        \includegraphics[width=\textwidth]{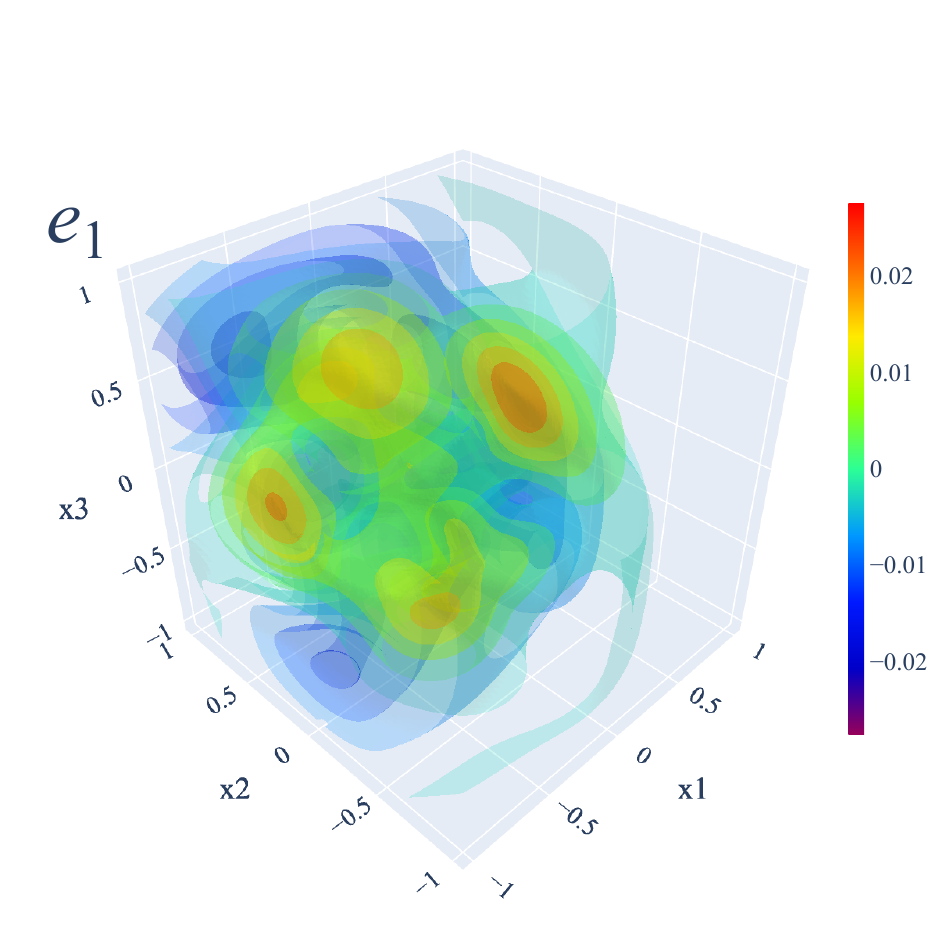}
        \caption{$e_1(x,t=0.6)$}
    \end{subfigure}
    \begin{subfigure}[b]{0.40\textwidth} %
        \includegraphics[width=\textwidth]{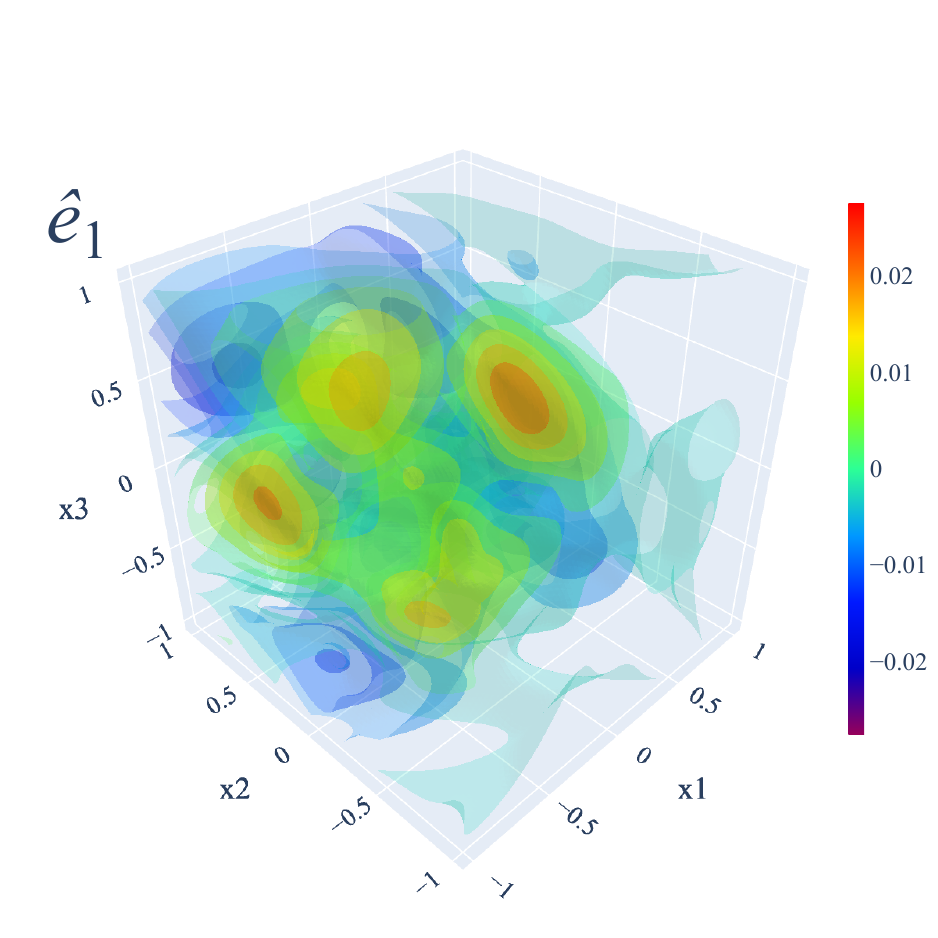}
        \caption{$\hat{e}_1(x,t=0.6)$}
    \end{subfigure}
    \begin{subfigure}[b]{0.40\textwidth} %
    \includegraphics[width=\textwidth]{figs/3dtv/e_t1.0.pdf}
        \caption{$e_1(x,t=1.0)$}
    \end{subfigure}
    \begin{subfigure}[b]{0.40\textwidth} %
        \includegraphics[width=\textwidth]{figs/3dtv/ehat_t1.0.pdf}
        \caption{$\hat{e}_1(x,t=1.0)$}
    \end{subfigure}
    \caption{Trained PINNs $\hat{e}_1(x,t)$ v.s. true error $e_1(x,t)$ for all $x$ at some $t$.}
    \label{fig:3dtv_e1hatresult}
\end{figure}

The neural networks of $\hat{p}$ and $\hat{e}_1$ are summarized in Table~\ref{tab:nnphat_3D-TVOU} and~\ref{tab:nne1hat_3D-TVOU}. 
For training $\hat{p}$, we begins with $N_0=N_r=2000$ samples. Half of these samples are drawn uniformly, and the other half follow the normal distribution specified by the initial condition. During training, we gradually add samples using adaptive sampling. 
For training $\hat{e}_1$, we begins with $N_0=N_r=300$ samples (with same distributions as training $\hat{p}$), and gradually add samples using adaptive sampling. 
The weights of training both $\hat{p}$ and $\hat{e}_1$ are $w_0=1,w_r=|T|=1$, and $w_{\nabla}=0$ without regularization.
The training losses of $\hat{p}$ and $\hat{e}_1$ are illustrated in Fig.~\ref{fig:3dtvou_trainloss}. 
The training results of $\hat{p}$ vs $p$ and $\hat{e}_1$ vs $e_t$ are shown in Fig.~\ref{fig:3dtv_phatresult} and~\ref{fig:3dtv_e1hatresult}. 

\begin{table}[htbp]
    \centering
    \begin{tabular}{lccc}
        \toprule
        \textbf{Layer Connection} & \textbf{Type} & \textbf{\# Neurons (Output)} & \textbf{Activation Function} \\
        \midrule
        Input Layer $\rightarrow$ Hidden Layer 1  & Fully Connected & 32  & GeLU \\
        Hidden Layer 1 $\rightarrow$ Hidden Layer 2  & Fully Connected & 32  & GeLU \\
        Hidden Layer 2 $\rightarrow$ Hidden Layer 3  & Fully Connected & 32  & GeLU \\
        Hidden Layer 3 $\rightarrow$ Hidden Layer 4  & Fully Connected & 32  & GeLU \\
        Hidden Layer 4 $\rightarrow$ Hidden Layer 5  & Fully Connected & 32  & GeLU \\
        Hidden Layer 5 $\rightarrow$ Output Layer  & Fully Connected & 1  & Softplus \\
        \bottomrule
    \end{tabular}
    \caption{Neural Network Architecture and Hyperparameters of $\hat{p}$}
    \label{tab:nnphat_3D-TVOU}
\end{table}

\begin{table}[htbp]
    \centering
    \begin{tabular}{lccc}
        \toprule
        \textbf{Layer Connection} & \textbf{Type} & \textbf{\# Neurons (Output)} & \textbf{Activation Function} \\
        \midrule
        Input Layer $\rightarrow$ Hidden Layer 1  & Fully Connected & 32  & GeLu \\
        Hidden Layer 1 $\rightarrow$ Hidden Layer 2  & Fully Connected & 32  & GeLu \\
        Hidden Layer 2 $\rightarrow$ Hidden Layer 3  & Fully Connected & 32  & GeLu \\
        Hidden Layer 3 $\rightarrow$ Hidden Layer 4  & Fully Connected & 32  & GeLu \\
        Hidden Layer 4 $\rightarrow$ Hidden Layer 5  & Fully Connected & 32  & GeLu \\
        Hidden Layer 5 $\rightarrow$ Output Layer  & Fully Connected & 1  & N/A \\
        \bottomrule
    \end{tabular}
    \caption{Neural network architecture and hyper-parameters of $\hat{e}_1$}
    \label{tab:nne1hat_3D-TVOU}
\end{table}

\clearpage
\subsection{7D Time-Varying OU Experiment}\label{appendix:7D-TVOU-exp}
\begin{figure}[htbp!]
    \centering
    \includegraphics[width=0.7\linewidth]{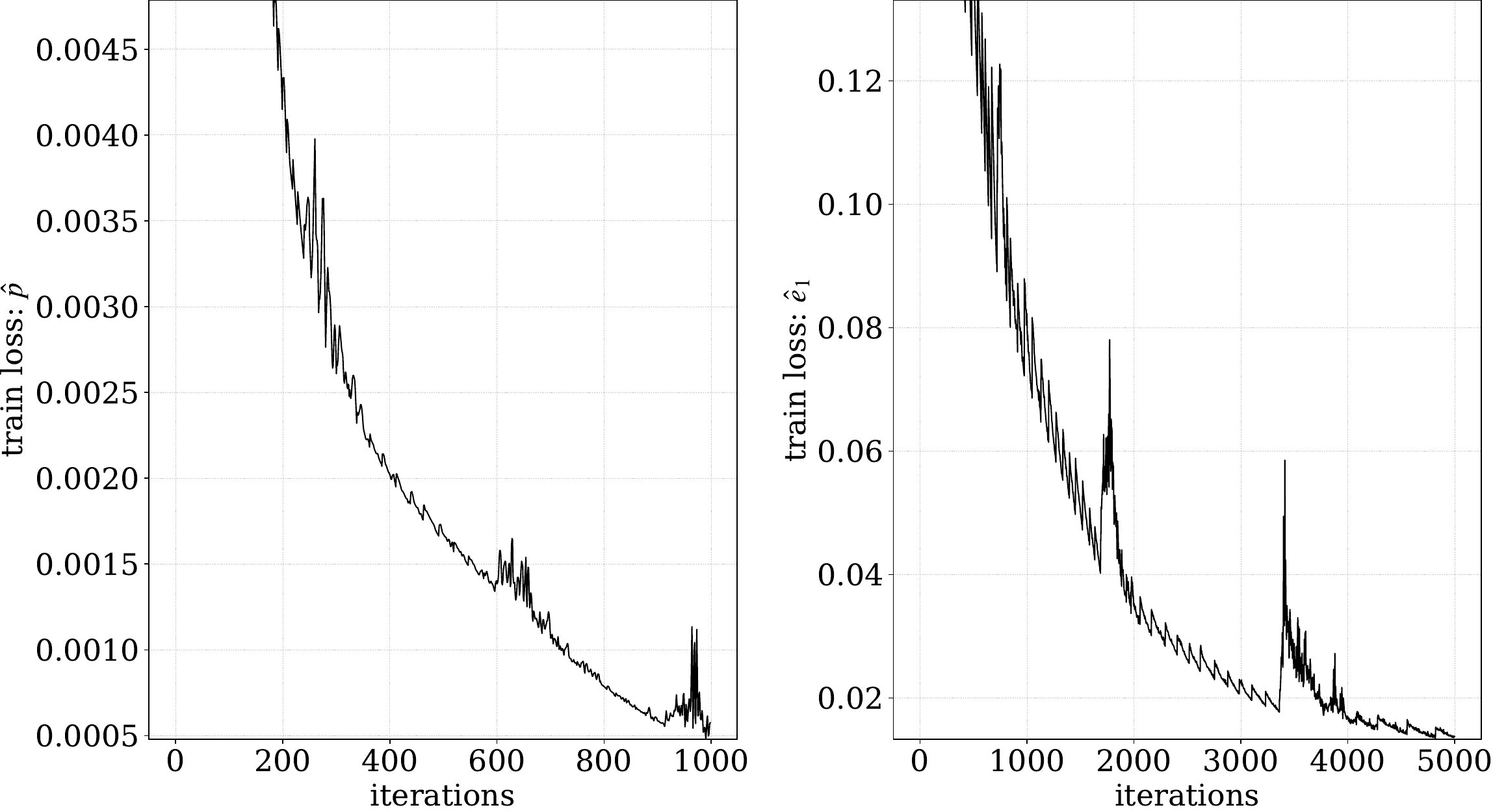}
    \caption{Training losses of $\hat{p}$ and $\hat{e}_1$}
    \label{fig:7dtvou_trainloss}
\end{figure}

The neural networks of $\hat{p}$ and $\hat{e}_1$ are summarized in Table~\ref{tab:nnphat_7D-TVOU} and~\ref{tab:nne1hat_7D-TVOU}. 
For training $\hat{p}$, we begins with $N_0=N_r=2000$ samples. Half of these samples are drawn uniformly, and the other half follow the normal distribution specified by the initial condition. During training, we gradually add samples using adaptive sampling. 
For training $\hat{e}_1$, we begins with $N_0=N_r=300$ samples (with same distributions as training $\hat{p}$), and gradually add samples using adaptive sampling. 
The weights of training both $\hat{p}$ and $\hat{e}_1$ are $w_0=1,w_r=|T|=1$, and $w_{\nabla}=0$ without regularization.
The training losses of $\hat{p}$ and $\hat{e}_1$ are illustrated in Fig.~\ref{fig:7dtvou_trainloss}. 

\begin{table}[htbp]
    \centering
    \begin{tabular}{lccc}
        \toprule
        \textbf{Layer Connection} & \textbf{Type} & \textbf{\# Neurons (Output)} & \textbf{Activation Function} \\
        \midrule
        Input Layer $\rightarrow$ Hidden Layer 1  & Fully Connected & 32  & GeLU \\
        Hidden Layer 1 $\rightarrow$ Hidden Layer 2  & Fully Connected & 32  & GeLU \\
        Hidden Layer 2 $\rightarrow$ Hidden Layer 3  & Fully Connected & 32  & GeLU \\
        Hidden Layer 3 $\rightarrow$ Hidden Layer 4  & Fully Connected & 32  & GeLU \\
        Hidden Layer 4 $\rightarrow$ Hidden Layer 5  & Fully Connected & 32  & GeLU \\
        Hidden Layer 5 $\rightarrow$ Output Layer  & Fully Connected & 1  & Softplus \\
        \bottomrule
    \end{tabular}
    \caption{Neural Network Architecture and Hyperparameters of $\hat{p}$}
    \label{tab:nnphat_7D-TVOU}
\end{table}

\begin{table}[htbp]
    \centering
    \begin{tabular}{lccc}
        \toprule
        \textbf{Layer Connection} & \textbf{Type} & \textbf{\# Neurons (Output)} & \textbf{Activation Function} \\
        \midrule
        Input Layer $\rightarrow$ Hidden Layer 1  & Fully Connected & 32  & GeLu \\
        Hidden Layer 1 $\rightarrow$ Hidden Layer 2  & Fully Connected & 32  & GeLu \\
        Hidden Layer 2 $\rightarrow$ Hidden Layer 3  & Fully Connected & 32  & GeLu \\
        Hidden Layer 3 $\rightarrow$ Hidden Layer 4  & Fully Connected & 32  & GeLu \\
        Hidden Layer 4 $\rightarrow$ Hidden Layer 5  & Fully Connected & 32  & GeLu \\
        Hidden Layer 5 $\rightarrow$ Output Layer  & Fully Connected & 1  & N/A \\
        \bottomrule
    \end{tabular}
    \caption{Neural network architecture and hyper-parameters of $\hat{e}_1$}
    \label{tab:nne1hat_7D-TVOU}
\end{table}

\clearpage
\subsection{10D Time-Varying OU Experiment}\label{appendix:10D-TVOU-exp}
\begin{figure}[htbp!]
    \centering
    \includegraphics[width=0.7\linewidth]{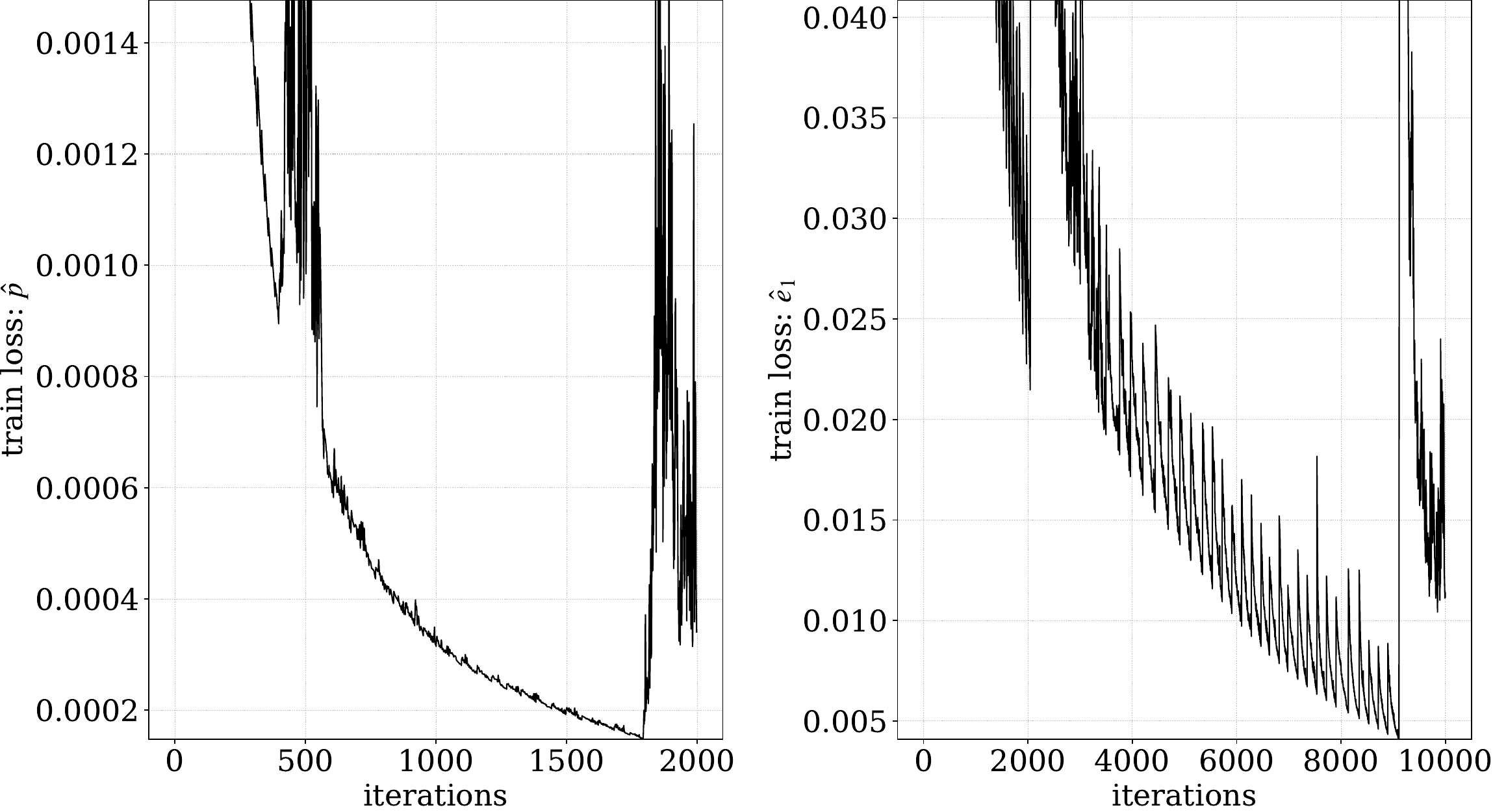}
    \caption{Training losses of $\hat{p}$ and $\hat{e}_1$}
    \label{fig:10dtvou_trainloss}
\end{figure}

The neural networks of $\hat{p}$ and $\hat{e}_1$ are summarized in Table~\ref{tab:nnphat_10D-TVOU} and~\ref{tab:nne1hat_10D-TVOU}. 
For training $\hat{p}$, we begins with $N_0=N_r=600$ samples. Half of these samples are drawn uniformly, and the other half follow the normal distribution specified by the initial condition. During training, we gradually add samples using adaptive sampling. 
For training $\hat{e}_1$, we begins with $N_0=N_r=600$ samples (with same distributions as training $\hat{p}$), and gradually add samples using adaptive sampling. 
The weights of training both $\hat{p}$ and $\hat{e}_1$ are $w_0=1,w_r=|T|=1$, and $w_{\nabla}=0$ without regularization.
The training losses of $\hat{p}$ and $\hat{e}_1$ are illustrated in Fig.~\ref{fig:10dtvou_trainloss}. 

\begin{table}[htbp]
    \centering
    \begin{tabular}{lccc}
        \toprule
        \textbf{Layer Connection} & \textbf{Type} & \textbf{\# Neurons (Output)} & \textbf{Activation Function} \\
        \midrule
        Input Layer $\rightarrow$ Hidden Layer 1  & Fully Connected & 50  & GeLU \\
        Hidden Layer 1 $\rightarrow$ Hidden Layer 2  & Fully Connected & 50  & GeLU \\
        Hidden Layer 2 $\rightarrow$ Hidden Layer 3  & Fully Connected & 50  & GeLU \\
        Hidden Layer 3 $\rightarrow$ Hidden Layer 4  & Fully Connected & 50  & GeLU \\
        Hidden Layer 4 $\rightarrow$ Hidden Layer 5  & Fully Connected & 50  & GeLU \\
        Hidden Layer 5 $\rightarrow$ Hidden Layer 6  & Fully Connected & 50  & GeLU \\
        Hidden Layer 6 $\rightarrow$ Output Layer  & Fully Connected & 1  & Softplus \\
        \bottomrule
    \end{tabular}
    \caption{Neural Network Architecture and Hyperparameters of $\hat{p}$}
    \label{tab:nnphat_10D-TVOU}
\end{table}

\begin{table}[htbp]
    \centering
    \begin{tabular}{lccc}
        \toprule
        \textbf{Layer Connection} & \textbf{Type} & \textbf{\# Neurons (Output)} & \textbf{Activation Function} \\
        \midrule
        Input Layer $\rightarrow$ Hidden Layer 1  & Fully Connected & 50  & GeLu \\
        Hidden Layer 1 $\rightarrow$ Hidden Layer 2  & Fully Connected & 50  & GeLu \\
        Hidden Layer 2 $\rightarrow$ Hidden Layer 3  & Fully Connected & 50  & GeLu \\
        Hidden Layer 3 $\rightarrow$ Hidden Layer 4  & Fully Connected & 50  & GeLu \\
        Hidden Layer 4 $\rightarrow$ Hidden Layer 5  & Fully Connected & 50  & GeLu \\
        Hidden Layer 5 $\rightarrow$ Hidden Layer 6  & Fully Connected & 50  & GeLu \\
        Hidden Layer 6 $\rightarrow$ Output Layer  & Fully Connected & 1  & N/A \\
        \bottomrule
    \end{tabular}
    \caption{Neural network architecture and hyper-parameters of $\hat{e}_1$}
    \label{tab:nne1hat_10D-TVOU}
\end{table}

\subsection{1D Heat PDE Experiment}\label{appendix:1D-HEAT-exp}
\begin{figure}[htbp!]
    \centering
    \includegraphics[width=0.7\linewidth]{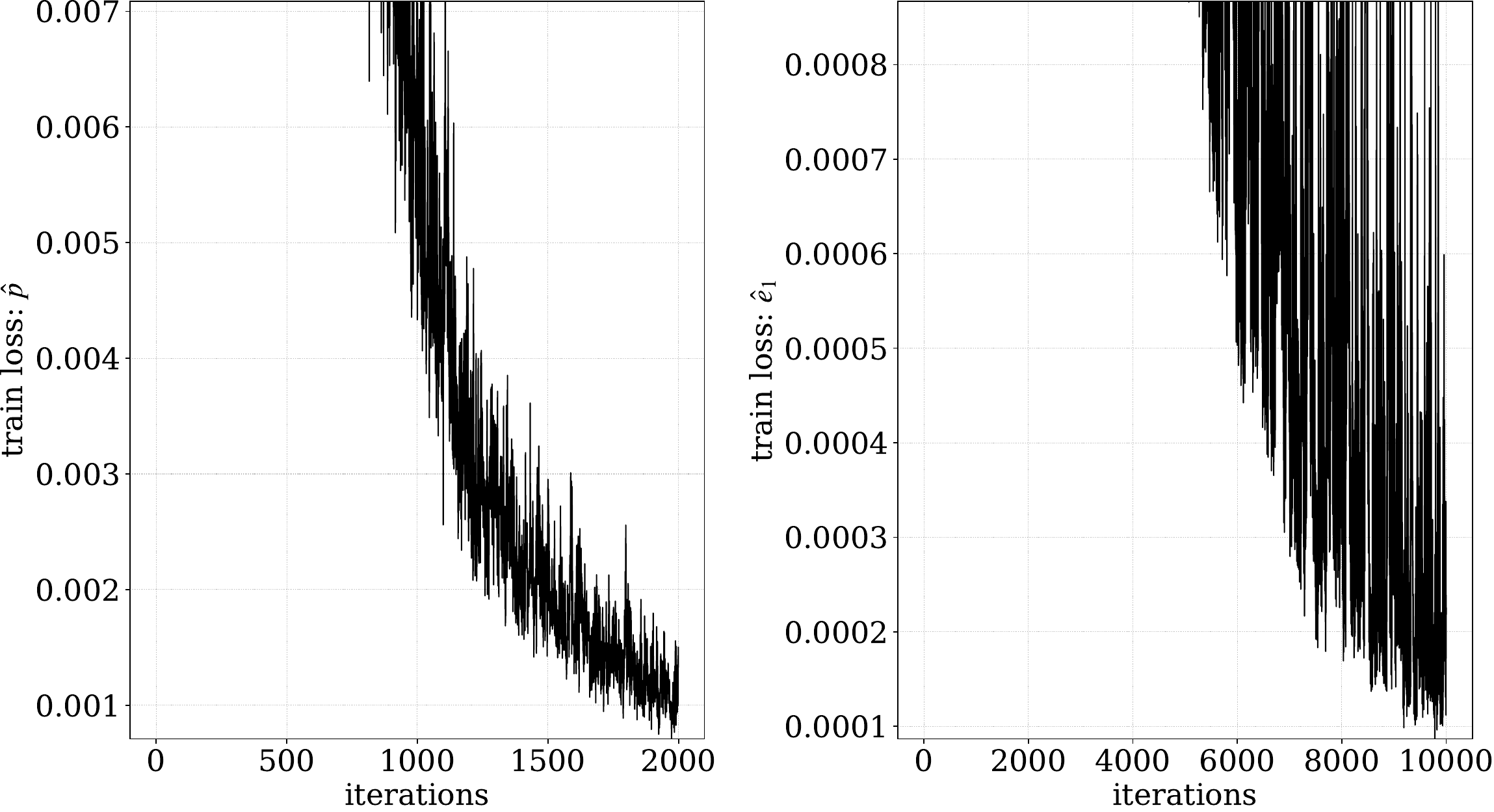}
    \caption{Training losses of $\hat{p}$ and $\hat{e}_1$}
    \label{fig:1dheat_trainloss}
\end{figure}

\begin{figure}[htbp!] 
    \centering
    \begin{subfigure}[b]{0.47\textwidth} 
        \includegraphics[width=\textwidth]{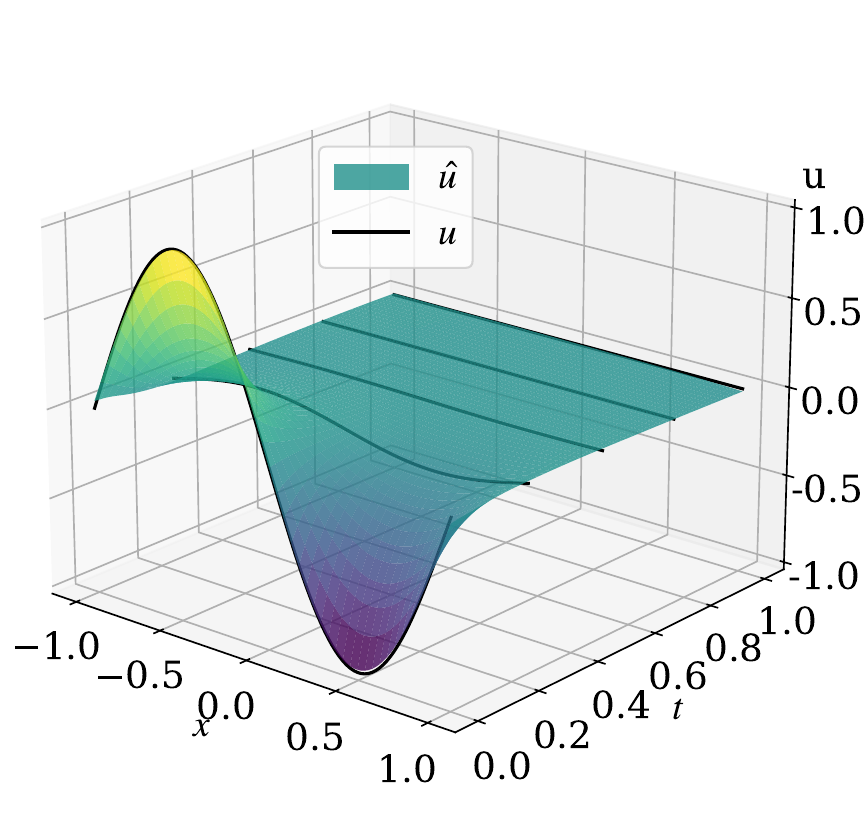}
        \caption{$\hat{u}$ vs $u$}
    \end{subfigure}
    \hfill
    \begin{subfigure}[b]{0.47\textwidth} 
        \includegraphics[width=\textwidth]{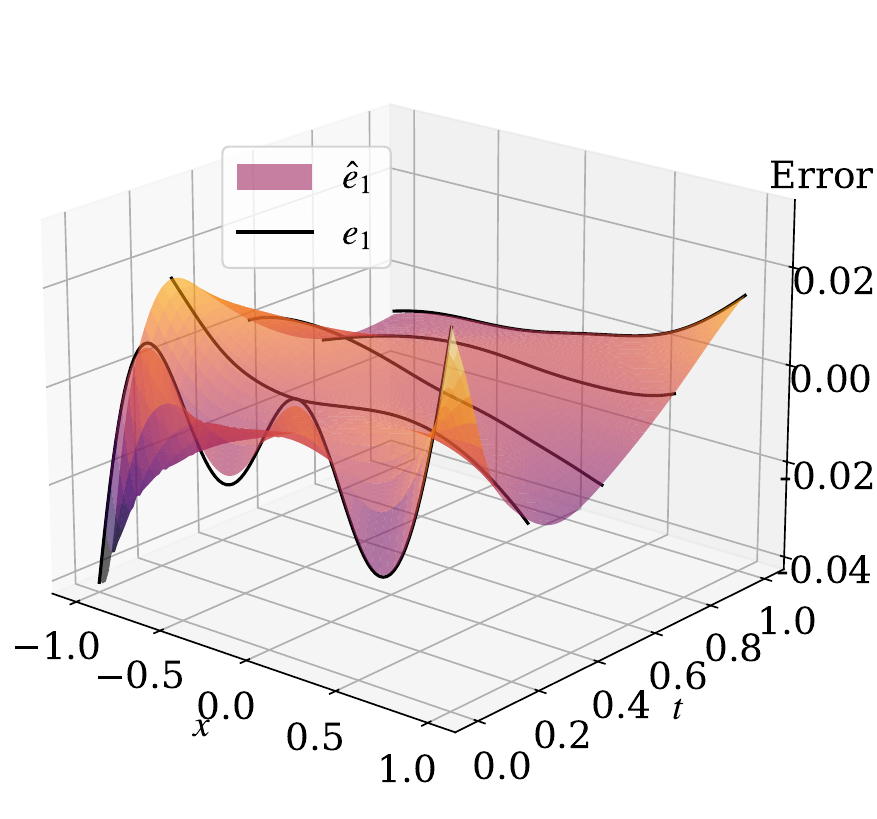}
        \caption{$\hat{e}_1$ vs $e_1$}
    \end{subfigure}
    \caption{Trained PINNs $\hat{u}(x,t)$ and $\hat{e}_1(x,t)$ v.s. true solution $u(x,t)$ and error $e_1(x,t)$ for all $x$ and $t$.}
    \label{fig:1dheat_surfaces}
\end{figure}

The neural networks of $\hat{p}$ and $\hat{e}_1$ are summarized in Table~\ref{tab:nnphat_1D-HEAT} and~\ref{tab:nne1hat_1D-HEAT}. 
For each training iteration, $N_0=N_r=500$ space-time points are uniformly sampled as in Eq.~\ref{eq:pinn_loss_general}, with weights $w_0=w_{bc}=1$ and $w_r=|T|=1$, where $w_{bc}$ is the weight of the Dirichlet boundary condition loss described in Appendix~\ref{proof:1D_Heat}.
The training losses of $\hat{p}$ and $\hat{e}_1$ are illustrated in Fig.~\ref{fig:1dheat_trainloss}. 
The training results of $\hat{p}$ vs $p$ and $\hat{e}_1$ vs $e_t$ are shown in Fig.~\ref{fig:1dheat_surfaces}, 

\begin{table}[htbp]
    \centering
    \begin{tabular}{lccc}
        \toprule
        \textbf{Layer Connection} & \textbf{Type} & \textbf{\# Neurons (Output)} & \textbf{Activation Function} \\
        \midrule
        Input Layer $\rightarrow$ Hidden Layer 1  & Fully Connected & 32  & Tanh \\
        Hidden Layer 1 $\rightarrow$ Hidden Layer 2  & Fully Connected & 32  & Tanh \\
        Hidden Layer 2 $\rightarrow$ Hidden Layer 3  & Fully Connected & 32  & Tanh \\
        Hidden Layer 3 $\rightarrow$ Output Layer  & Fully Connected & 1  & N/A \\
        \bottomrule
    \end{tabular}
    \caption{Neural Network Architecture and Hyperparameters of $\hat{p}$}
    \label{tab:nnphat_1D-HEAT}
\end{table}

\begin{table}[htbp]
    \centering
    \begin{tabular}{lccc}
        \toprule
        \textbf{Layer Connection} & \textbf{Type} & \textbf{\# Neurons (Output)} & \textbf{Activation Function} \\
        \midrule
        Input Layer $\rightarrow$ Hidden Layer 1  & Fully Connected & 50  & Tanh \\
        Hidden Layer 1 $\rightarrow$ Hidden Layer 2  & Fully Connected & 50  & Tanh \\
        Hidden Layer 2 $\rightarrow$ Hidden Layer 3  & Fully Connected & 50  & Tanh \\
        Hidden Layer 3 $\rightarrow$ Hidden Layer 4  & Fully Connected & 50  & Tanh \\
        Hidden Layer 4 $\rightarrow$ Hidden Layer 5  & Fully Connected & 50  & Tanh \\
        Hidden Layer 5 $\rightarrow$ Output Layer  & Fully Connected & 1  & N/A \\
        \bottomrule
    \end{tabular}
    \caption{Neural network architecture and hyper-parameters of $\hat{e}_1$}
    \label{tab:nne1hat_1D-HEAT}
\end{table}

\begin{figure}[htbp!]
    \centering
    \includegraphics[width=0.7\linewidth]{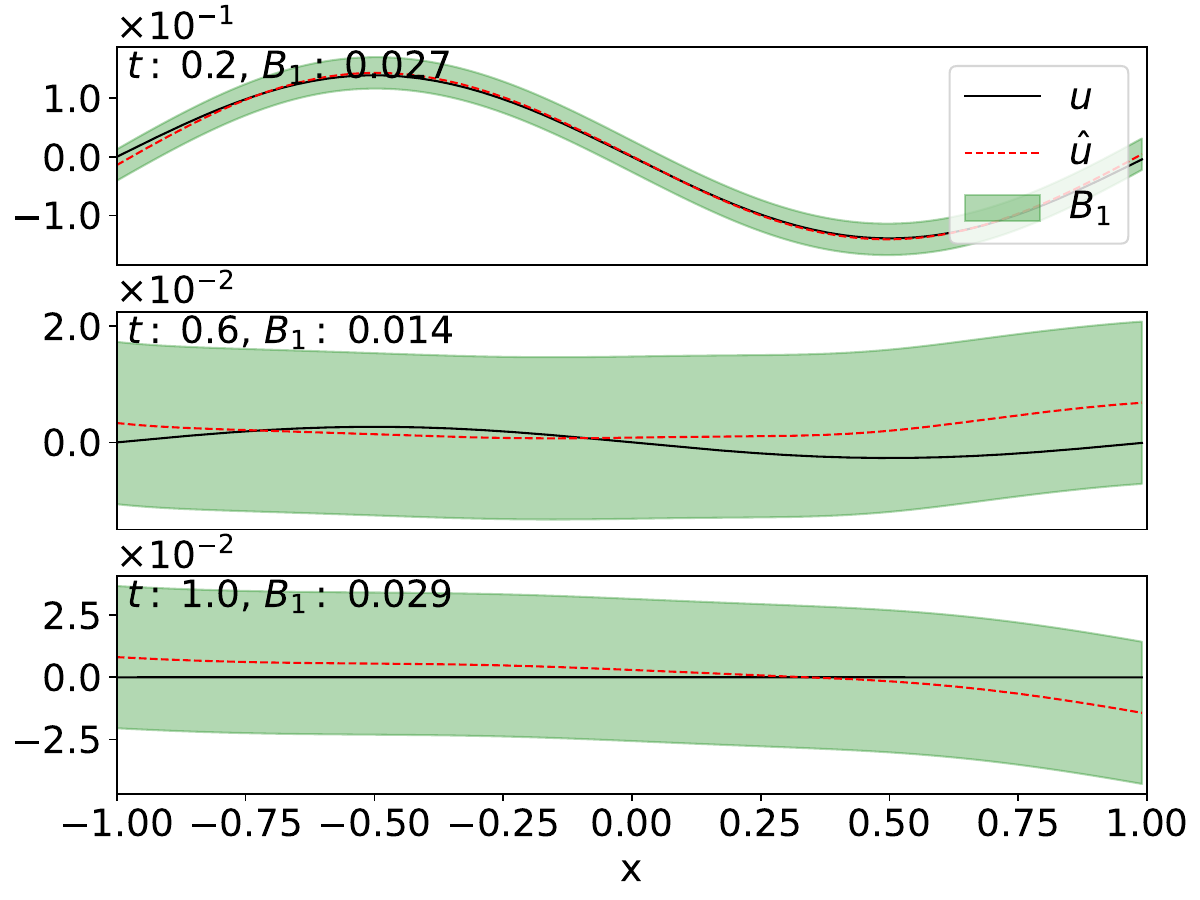}
    \caption{$u$, $\hat{u}$, and the error bound $B_1$ at some $t$.}
    \label{fig:1dheat_errorboundsresult}
\end{figure}

\end{document}